%% file: neurips_2020.tex
\pdfoutput=1

\documentclass{article}

\usepackage[preprint,nonatbib]{neurips_2020}

\usepackage[utf8]{inputenc} 
\usepackage[T1]{fontenc}    
\usepackage{hyperref}       
\usepackage{url}            
\usepackage{booktabs}       
\usepackage{amsfonts}       
\usepackage{nicefrac}       
\usepackage{microtype}      
\usepackage{graphicx}
\usepackage{soul}
\usepackage{booktabs}
\usepackage{multirow}
\usepackage{xcolor}
\usepackage{amsmath}
\usepackage[ruled,vlined]{algorithm2e}
\usepackage{amssymb}
\usepackage{arydshln}
\usepackage{siunitx}
\usepackage{subcaption}
\usepackage{tabularx}
\newcolumntype{C}{>{\centering\arraybackslash}X}

\DeclareMathOperator*{\argmax}{arg\,max}

\SetCommentSty{mycommfont}

\makeatletter
\def\adl@drawiv#1#2#3{%
        \hskip.5\tabcolsep
        \xleaders#3{#2.5\@tempdimb #1{1}#2.5\@tempdimb}%
                #2\z@ plus1fil minus1fil\relax
        \hskip.5\tabcolsep}
\newcommand{\cdashlinelr}[1]{%
  \noalign{\vskip\aboverulesep
           }
  \cdashline{#1}
  \noalign{
           \vskip\belowrulesep}}
\makeatother

\newcommand{\comment}[1]{}

\newcommand{\ours}{BRP-NAS}

\newcommand{\beginsupplement}{%
        \setcounter{table}{0}
        \renewcommand{\thetable}{S\arabic{table}}%
        \setcounter{figure}{0}
        \renewcommand{\thefigure}{S\arabic{figure}}%
        \setcounter{section}{0}
        \renewcommand{\thesection}{S\arabic{section}}%
     }
     
\title{Latency-Accuracy Prediction for Neural Architecture Search}
\title{Multiobjective Prediction-based Neural Architecture Search}
\title{LatBench: Advancing latency and accuracy prediction in NAS}
\title{BRP-NAS: Multiobjective Prediction-based NAS using GCNs}
\title{BRP-NAS: Prediction-based NAS using GCNs}

\author{%
  Łukasz Dudziak\textsuperscript{1*}, Thomas Chau\textsuperscript{1*}, Mohamed S. Abdelfattah\textsuperscript{1} \\
  \textbf{Royson Lee\textsuperscript{1}, Hyeji Kim\textsuperscript{3$\dagger$}, Nicholas D. Lane\textsuperscript{1,2}} \\
  \\
  \textsuperscript{1} Samsung AI Center, Cambridge, UK\\
  \textsuperscript{2} University of Cambridge, UK \hspace*{10pt}
  \textsuperscript{3} University of Texas at Austin, US\\
  \footnotesize{\textsuperscript{*} \textit{Indicates equal contributions}, 
  \textsuperscript{$\dagger$} \textit{Work performed while at Samsung AI Center}} \\
  \\
  \texttt{\{l.dudziak, thomas.chau, mohamed1.a, royson.lee, nic.lane\}@samsung.com} \\
  \texttt{hyeji.kim@austin.utexas.edu}
}

\begin{document}

\maketitle

\begin{abstract}
\vspace{-0.3cm}
Neural architecture search (NAS) enables researchers to automatically explore broad design spaces in order to improve efficiency of neural networks. This efficiency is especially important in the case of on-device deployment, where improvements in accuracy should be balanced out with computational demands of a model. In practice, performance metrics of model are computationally expensive to obtain. Previous work uses a proxy (e.g., number of operations) or a layer-wise measurement of neural network layers to estimate end-to-end hardware performance but the imprecise prediction diminishes the quality of NAS. To address this problem, we propose BRP-NAS, an efficient hardware-aware NAS enabled by an accurate performance predictor-based on graph convolutional network (GCN). What is more, we investigate prediction quality on different metrics and show that sample efficiency of the predictor-based NAS can be improved by considering binary relations of models and an iterative data selection strategy. We show that our proposed method outperforms all prior methods on NAS-Bench-101 and NAS-Bench-201, and that our predictor can consistently learn to extract useful features from the DARTS search space, improving upon the second-order baseline. Finally, to raise awareness of the fact that accurate latency estimation is not a trivial task, we release LatBench -- a latency dataset of NAS-Bench-201 models running on a broad range of devices.
\end{abstract}

\vspace{-0.3cm}
\section{Introduction}\label{sec:intro}
\vspace{-0.3cm}
\input{10-intro}

\section{Related work}\label{sec:related}
\vspace{-0.3cm}
\input{20-related}

\section{Latency prediction in NAS}\label{sec:latency}
\vspace{-0.3cm}
\input{30-latency}

\section{Accuracy prediction in NAS}\label{sec:accuracy}
\vspace{-0.3cm}
\input{31-accuracy}

\section{End-to-end results}\label{sec:results}
\vspace{-0.3cm}
\input{32-results}

\section{Latency prediction benchmark}\label{sec:benchmark}
\vspace{-0.4cm}
\input{40-benchmark}

\vspace{-0.25cm}

\section{Discussion and limitations}\label{sec:limitations}
\vspace{-0.4cm}
\input{50-limitations}
\vspace{-0.25cm}

\section{Conclusion}\label{sec:conclusion}
\vspace{-0.4cm}
\input{60-conclusion}


\section*{Broader Impact}


This research can democratize on-device deployment with cost-efficient NAS methodology for model optimization within device latency constraints. Additionally, carbon footprint of traditionally expensive NAS methods is vastly reduced.
On the other hand, measurement and benchmarking data can be used both to create new NAS methodologies, and to gain further insights about the device performance. This can bridge the machine learning and device research communities together.

\section*{Funding Disclosure}
This work was done as a part of the authors’ jobs at Samsung AI Center. The authors have nothing to disclose.


\small
\bibliographystyle{unsrt}
\bibliography{neurips_2020}

\newpage
\section*{Supplementary Material}
\input{70-appendix}

\end{document}

%% file: 10-intro.tex
Neural architecture search (NAS) has demonstrated great success in automatically designing competitive neural networks compared with hand-crafted alternatives~\cite{zoph2018, cai2018a, cai2018b, ProxylessNAS_2019}.
However, NAS is computationally expensive requiring to train models~\cite{NASinitial_2017, baker2017} or introduce non-trivial complexity into the search process~\cite{pham2018,brock2018}.
Additionally, real-world deployment demands models meeting efficiency or hardware constraints (e.g., latency, memory and energy consumption) on top of being accurate, but acquiring various performance metrics of a model can also be time consuming, independently from the cost of training it.

In this paper, we ($a$) show the limitations of the layer-wise predictor both in terms of prediction accuracy and NAS performance and ($b$) propose a \emph{Graph convolutional networks} (GCN)-based predictor for the end-to-end latency which significantly outperforms the layer-wise approach on devices of various specifications.

One of the key challenges in obtaining a reliable accuracy predictor is that acquiring training samples (pairs of \textit{(model, accuracy)}) is computationally expensive.
\emph{Sample efficiency} denotes how many samples are required to find the best model during a search under a target hardware constraint, and significantly advancing this metric is a key contribution of our work.
We propose several methods to improve sample efficiency -- 
($a$) We observe that in the context of NAS, instead of getting precise estimates of accuracy, we want to produce a linear ordering of accuracy to search for the best model.
Therefore, we propose a binary relation predictor to decide the accuracy ranking of neural networks without requiring to estimate absolute accuracy values.
($b$) To help the predictor focus on predicting the rankings of top candidates, which is the most important to yield the best results in NAS, we propose an iterative data selection scheme  which vastly improves the sample efficiency of NAS.

The contributions of this paper are summarized as follows: 
\begin{itemize}
    \item \textbf{Latency prediction.}  
    We empirically show that an accurate latency predictor plays an important role in NAS where latency on the target hardware is of interest, and existing latency predictors are overly error-prone.
    We propose an end-to-end NAS latency predictor-based on a GCN and show that it outperforms previous approaches (proxy, layer-wise) on various devices. To the best of our knowledge, this is the first end-to-end latency predictor. We illustrate its behaviour on various devices and show that this predictor works well across all of them (Section~\ref{sec:latency}). 

    \item \textbf{Accuracy prediction.} 
    We introduce a novel training methodology for a NAS-specific accuracy predictor by turning an accuracy prediction problem into a binary prediction problem, where we predict which one of two neural architectures performs better, resulting in improved overall ranking correlation between predicted and ground-truth rankings. 
    Furthermore, we propose a new prediction-based NAS framework called \ours. It combines a binary relation accuracy predictor architecture and an iterative data selection strategy to improve the top-K ranking correlation. \ours{} outperforms previous NAS methods by being more sample efficient (Section~\ref{sec:accuracy} and~\ref{sec:results}).

    \item \textbf{Towards reproducible research: latency benchmark.} We introduce \textit{LatBench}, the first large-scale latency measurement dataset for multi-objective NAS.
    Unlike existing datasets which either approximate the latency or focus on a single device, LatBench provides measurement dataset on a broad range of systems covering desktop CPU/GPU, embedded GPU/TPU and mobile GPU/DSP. We also release \textit{Eagle} which is a tool to measure and predict performance of models on various systems. We make LatBench and the source code of Eagle available publicly\footnote{https://github.com/thomasccp/eagle} (Section~\ref{sec:benchmark}).
\end{itemize}

%% file: 20-related.tex
\textbf{NAS and performance estimation.}
Various end-to-end predictors have been proposed and studied for accuracy.
Initially, accuracy predictors were shown to be helpful in guiding a NAS~\cite{Liu_2018_ECCV}.
More recently, it has been shown that accuracy predictor alone can be used to perform a search~\cite{NPENAS2020, 1912.00848}.

Performance of a neural network is captured by several metrics, such as accuracy, latency, and energy consumption.
Because measuring performance metrics is expensive both in terms of time and computation, interest in predicting them has surged since neural architecture search was introduced in~\cite{NASinitial_2017}.
Among several metrics, accuracy prediction is arguably the most actively studied in the context of NAS. 
Earlier performance prediction works focused on extrapolating the learning curve to reduce the training time~\cite{NAS_Predictor_2018,Klein2017LearningCP,Domhan2015SpeedingUA}.
Recent works, on the other hand, explored performance prediction based on architectural properties.
For instance, it was demonstrated that an accuracy predictor trained during the search could be successfully used to guide it and, in turn, accelerate it~\cite{Liu_2018_ECCV}. 
Taking advantage of the differentiability of accuracy predictors, NAO~\cite{NAO_2018_NIPS} introduced a gradient-based optimization of neural architectures. 
Graph neural networks, including GIN~\cite{GNN_2019}, D-VAE~\cite{DVAE_2019} and GATE~\cite{GATE_2020} have been used to learn representations of neural architectures.
In this paper, we use GCN~\cite{GCN_2017} to handle neural architectures described as directed graphs.
Closely related to our work, NPENAS~\cite{NPENAS2020} used graph neural network-based accuracy predictors and an iterative approach to estimate the accuracy of models. 
However, instead of training the predictor with the top-k mutated models based on previously found models, we trained the predictor by picking both the top-k and random models from the entire search space, an technique that balances exploration and exploitation.
Additionally, we incorporate transfer learning to further boost the performance of the predictor as shown in Section~\ref{sec:acc-lat}.

Recent focus has been on improving sample efficiency.
In that regard, ~\cite{LaNAS_2019} proposed adapting the action space during NAS, where a Monte-Carlo tree search was used to split the action space into good and bad regions.
Towards the same goal, ~\cite{1912.00848} introduced a simple NAS based on accuracy predictor, where models with the top-K best predicted accuracy were fully trained, after which the best one was chosen. 
Similarly, BONAS~\cite{BONAS_2020} incorporated a GCN accuracy predictor as a surrogate function of Bayesian Optimization to perform NAS and used exponential weighted loss to improve the prediction of models with high accuracy. 
Our work extends GCN to binary relation learning to focus on the prediction of ranking rather than absolute values and uses iterative data selection to explicitly train the predictor with high performing models.

\textbf{Focused latency prediction.} Several works have been proposed to predict performance metrics such as accuracy and latency, the two most popular metrics of interest, instead of measuring them~\cite{Liu_2018_ECCV, NPENAS2020, 1912.00848}. 
Recent work either measures the latency directly from devices during the search process~\cite{MnasNet_2019, FBNet_2019}, which is accurate but slow and expensive, or rely on a proxy metric (e.g., FLOPS or model size)~\cite{DPPNet_2018}, which is fast but inaccurate.
More recently, layer-wise predictors have been proposed~\cite{ProxylessNAS_2019} which effectively sum up the latencies of individual neural network layers.
While being more accurate than the proxy, layer-wise predictors have a significant drawback: they do not capture the complexities of multiple layer executing on real hardware.


\textbf{Multiobjective NAS.} A few works have explored NAS with multiple objectives and hardware constraints. 
Among those,~\cite{ChamNet_2019} proposed a hardware-aware adaptation of neural architectures via evolutionary search, where the performance metrics of each architecture were estimated by the predictors. Latency was estimated via a lookup table while accuracy and energy consumption predictors were modeled as Gaussian process regression.

%% file: 30-latency.tex
In this section, we demonstrate the limitations of existing latency predictors and introduce a GCN-based latency predictor which ($a$) significantly outperforms the existing predictors on a wide range of devices in absolute accuracy and ($b$) contributes to a significant improvement in NAS for latency constrained deployment.
Throughout the paper, we focus on NAS-Bench-201 dataset which includes 15,625 models.
We use desktop CPU, desktop GPU and embedded GPU to refer to the devices used in our analysis, with device details described in Section~\ref{sec:benchmark}. 

\subsection{Existing latency predictors and their limitations in NAS}
\label{sec:existing_latency_predictors}
\vspace{-0.25cm}

\begin{figure}[!ht]
    \centering
    \includegraphics[width=0.32\textwidth]{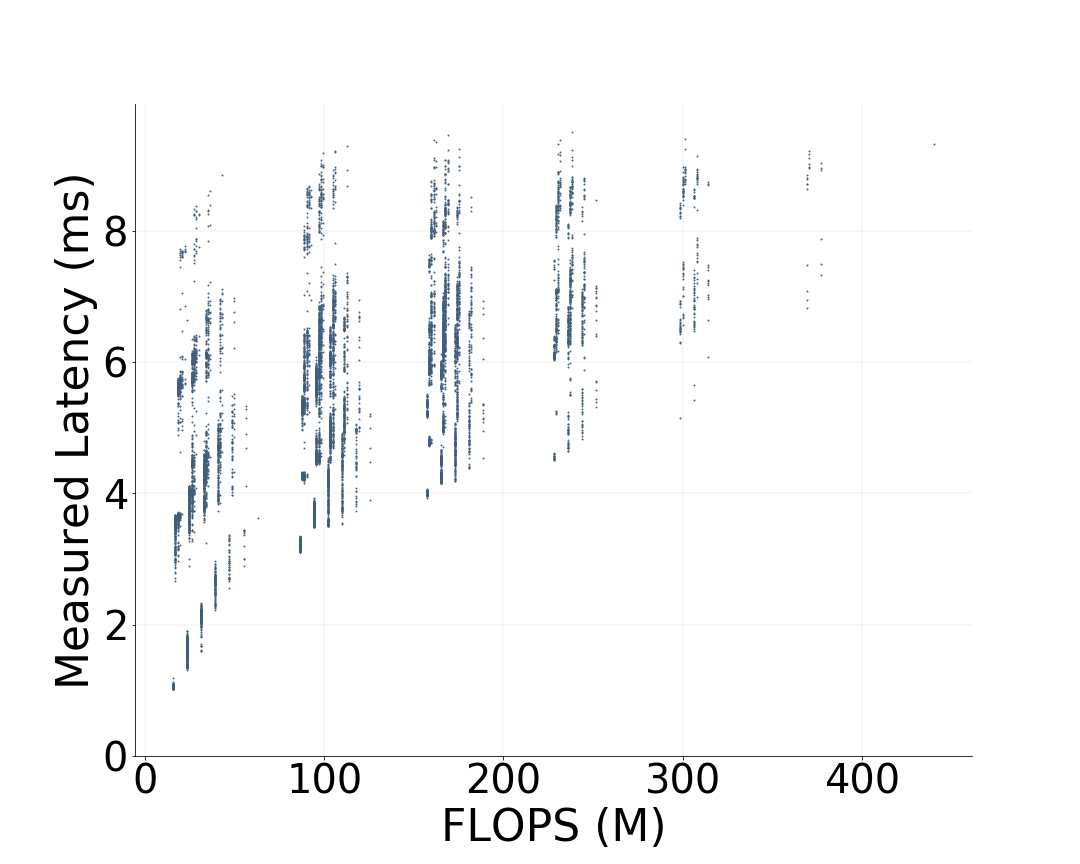} 
    \includegraphics[width=0.32\textwidth]{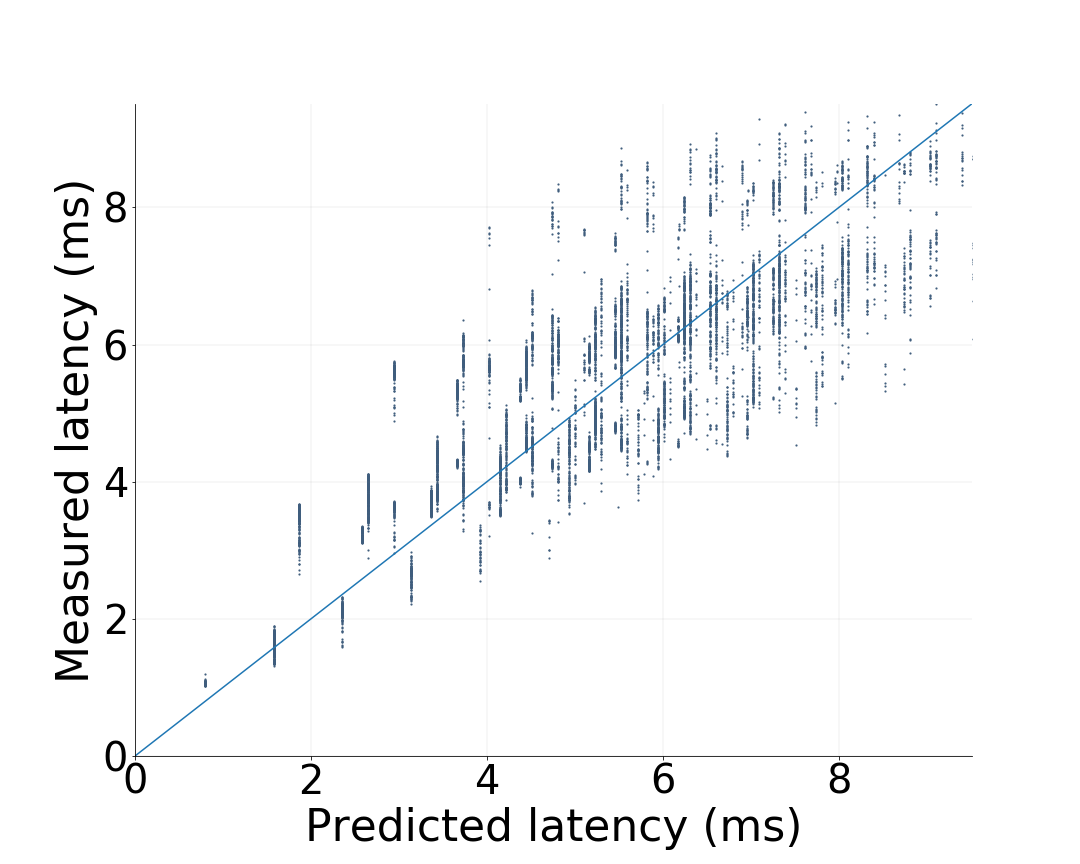}
    \includegraphics[width=0.32\textwidth]{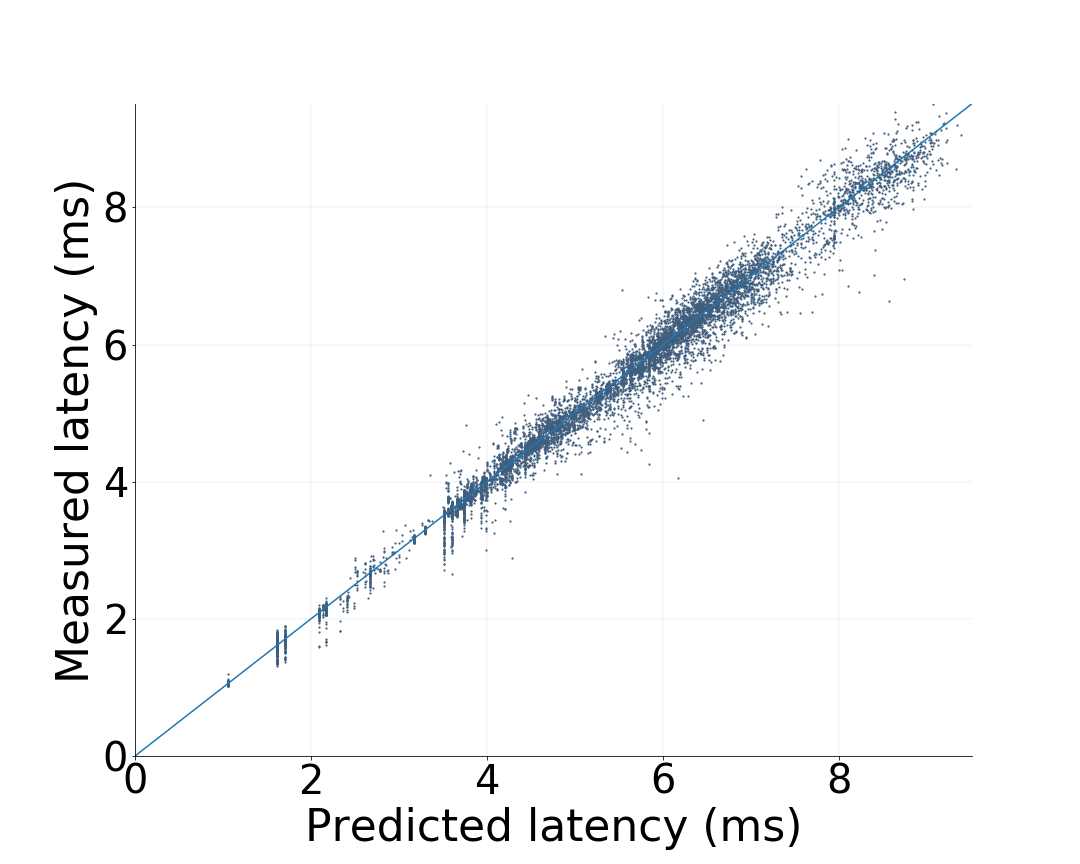}
    \caption{ 
    (Left) Number of FLOPS is not a good proxy for the latency estimation.
    Our GCN-based end-to-end latency predictor (right) is more accurate than layer-wise predictor (middle).
    Results shown here are based on desktop CPU.
    }
    \label{fig:benchmark-analysis}
\end{figure}

Number of FLOPS and parameters are often used as proxies for latency estimation for their simplicity but have been shown to be inaccurate in many cases~\cite{FBNet_2019, Ning_2018_ECCV}.
In Figure~\ref{fig:benchmark-analysis} (left), we show the scatter plot of models taken from NAS-Bench-201 dataset that illustrates the connection between the latency and FLOPS.
Each point in the plots represents the average latency of running a model on the stated device. 
We can see that latency is not strongly correlated with FLOPS. 
Recent works~\cite{ProxylessNAS_2019, FBNet_2019} use a layer-wise predictor which derives the latency by summing latency measured for each operation in the model individually. 
However, as shown in Figure~\ref{fig:benchmark-analysis} (middle), the layer-wise predictor\footnote{The layer-wise predictor is calibrated by a scaling factor to fit the end-to-end latency in the training set.} also leads to inaccurate predictions of the end-to-end latency.
It assumes sequential processing of operations and cannot reflect the key model and hardware characteristics that affect the end-to-end latency, e.g., whether operations within a model can be executed in parallel on the target hardware.
In section~\ref{sec:gcn}, we introduce an end-to-end latency predictor that is trained with the end-to-end measured latency that significantly improves the prediction accuracy as shown in Figure~\ref{fig:benchmark-analysis} (right). 

\begin{figure}[t]
    \centering
    \includegraphics[width=0.44\textwidth]{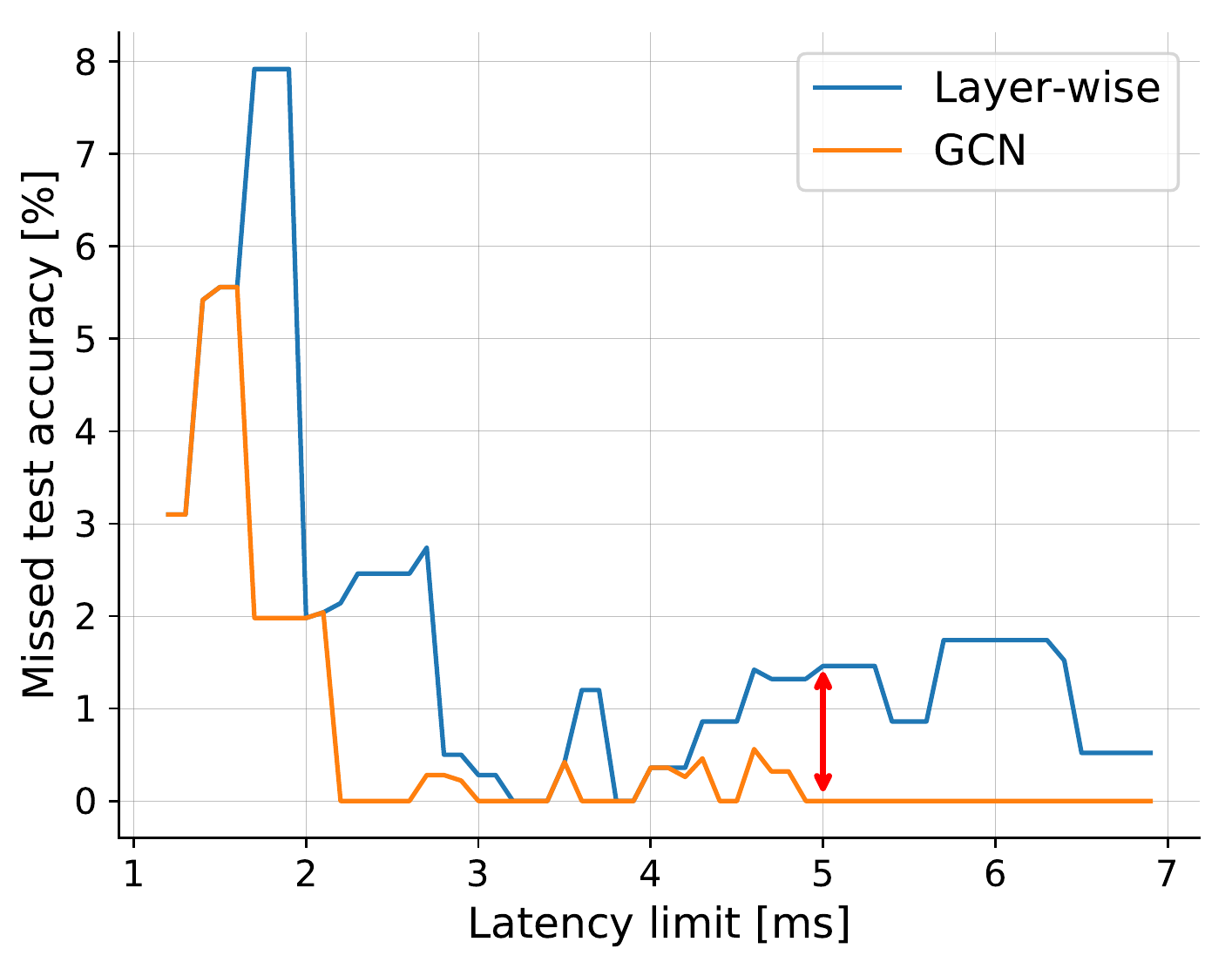}
    \includegraphics[width=0.44\textwidth]{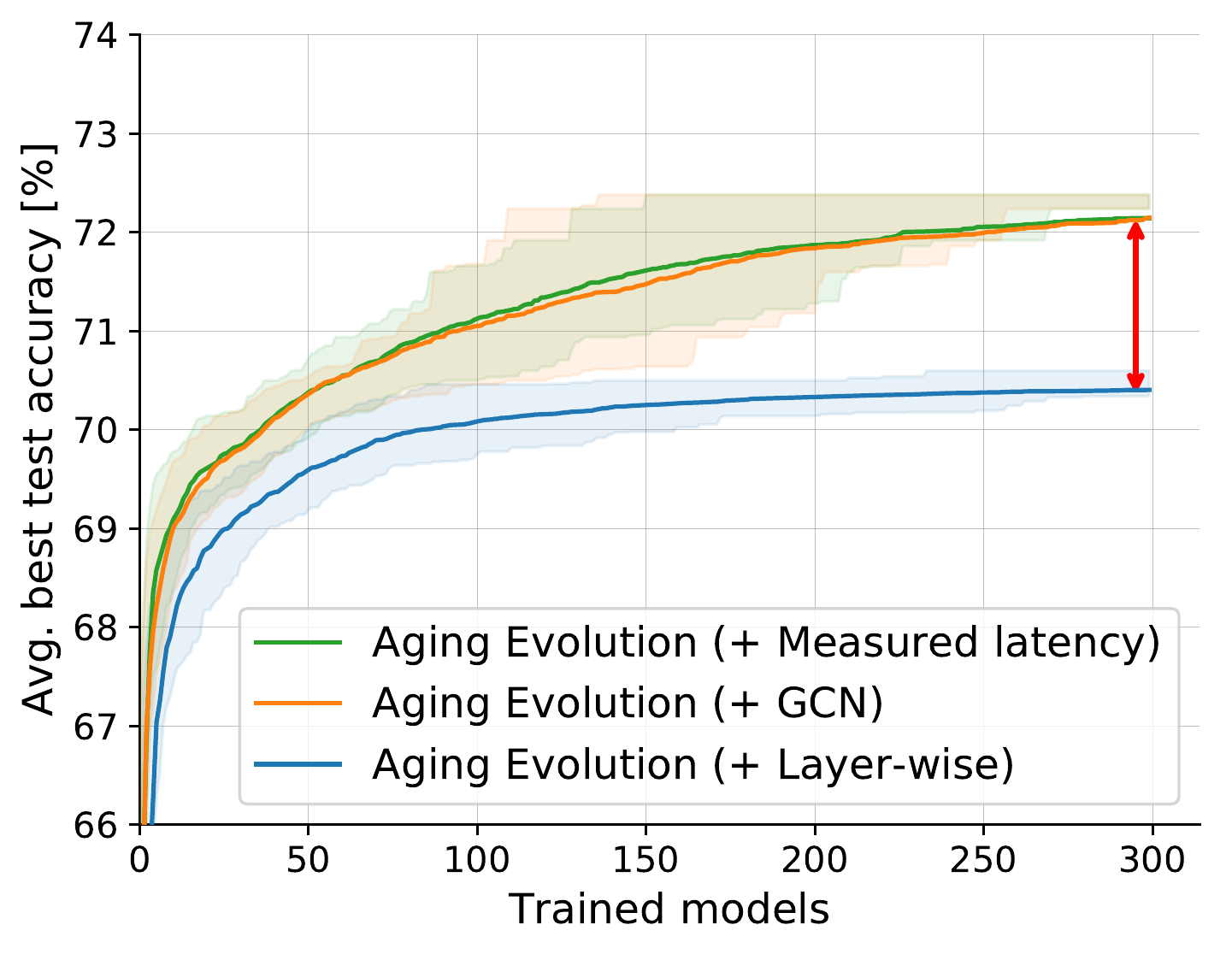}
    \caption{Importance of an accurate latency predictor in NAS for latency-constrained deployment -- (Left) Best achievable models missed when using an oracle NAS that relies on predicted latency on desktop GPU. The gap between the layer-wise and GCN curves shows the impact of poor latency predictor. (Right) Best accuracy of models found by the aging-evolution search with predicted latency and measured latency on desktop GPU with 5ms latency limit. Shaded regions mark interquartile range and the same applies to subsequent figures.} 
    \label{fig:terrible-NAS} 
    \vspace{-4mm}
\end{figure}

\textbf{NAS with latency predictors.} We analyze the impact of layer-wise latency predictor on NAS for latency-constrained deployment, where the objective is to find the most accurate model that satisfies a strict latency constraint (e.g., for real-time applications~\cite{venieris2017fpl, approxlstm2020cemag}). 
We consider two NAS algorithms, oracle NAS and Aging Evolution~\cite{Real2019RegularizedEF} (AE).
Oracle NAS can find the best performing model (in terms of accuracy) among those with latency satisfying the target constraint -- therefore its performance is limited entirely by the quality of latency estimation (more details are provided in the S.M.).
In Figure~\ref{fig:terrible-NAS} (left), we plot the difference between the best achievable accuracy and the best accuracy on CIFAR-100 dataset obtained by an oracle NAS that relies on the predicted latency - as a function of the target latency constraint.
We can see that the accuracy loss due to the inaccurate predictor is non-negligible and sometimes very large (up to 8$\%$). 
This loss is also visible when other NAS algorithms are used. In Figure~\ref{fig:terrible-NAS} (right), 
we plot the best accuracy of models found by the aging-evolution search with predicted latency and measured latency as a function of search step. 
These experimental results highlight the importance of an accurate latency predictor in NAS. 

\begin{figure}[t]
    \centering
    \includegraphics[width=0.99\textwidth]{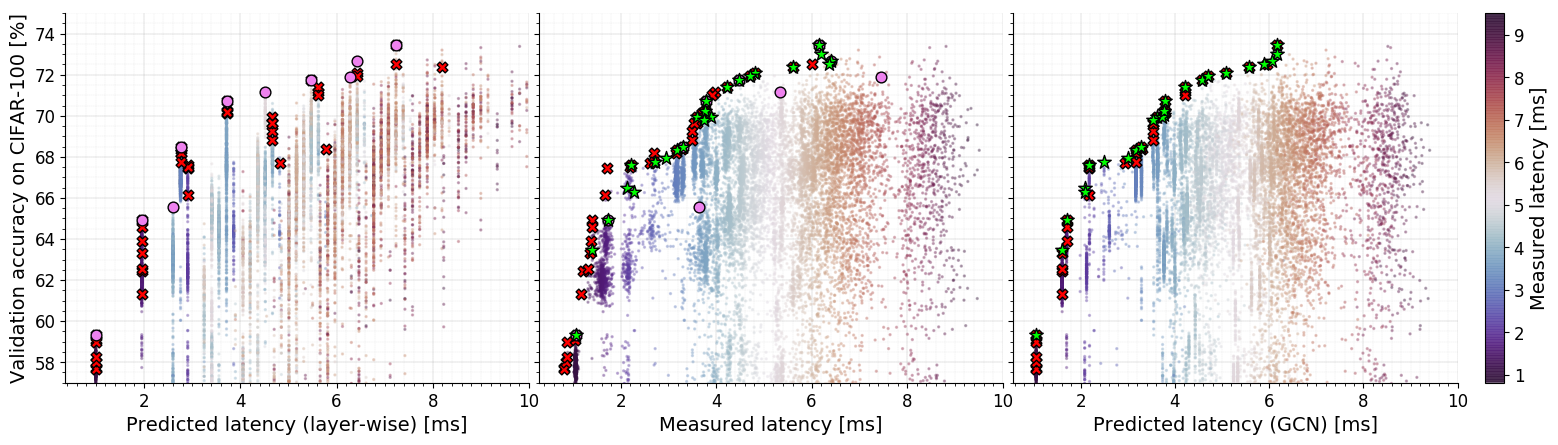}
    \caption{Red (x), Pink (o), Green marks (*) represent Pareto-optimal models on accuracy vs measured latency, layer-wise predicted latency, and GCN predicted latency, respectively, on desktop GPU.
    (Left) Many Pareto-optimal models (red, x) are not located at the Pareto-frontier implying that an oracle NAS cannot discover Pareto-optimal models with the layer-wise predicted latency.
    (Right) Most Pareto-optimal points (red, x) are located at the Pareto-frontier implying that an oracle NAS is able to discover Pareto-optimal models with our GCN predicted latency.
    }
    \label{fig:gcn_vs_layerwise_acc_and_pareto}
    \vspace{-4mm}
\end{figure}

\textbf{Analysis of Pareto-optimal models.}
In order to systematically study the impact of inaccurate predictions on latency-aware NAS, we run an analysis of the Pareto-optimal models.
Pareto-optimal models are solutions to NAS given a strict latency constraint or if the objective is a weighted combination of the accuracy and the latency~\cite{moo_book}.
We ask the following: can the Pareto-optimal models be discovered when a latency predictor is used?
Suppose we are given an oracle NAS algorithm that returns a Pareto-optimal model based on the accuracy and \emph{predicted} latency.
How far off is that model from the desired Pareto-optimal solution in the accuracy and \emph{measured} latency plot? 
In Figure~\ref{fig:gcn_vs_layerwise_acc_and_pareto} (left and middle), we show scatter plots of NAS-Bench-201~\cite{nasbench2} models, where the y-axis represents the accuracy and the x-axis represents latency predicted via the layer-wise predictor (left), and measured latency (middle), respectively. Pareto-optimal models in the predicted space and measured space are marked with pink (o) and red (x), respectively, and shown in both figures. 
We can see that the Pareto-optimal models in one space do not always lie at the Pareto frontier of the other space.
This is problematic because it implies that even if we had a perfect NAS algorithm that discovers Pareto optimal points (based on the predicted latency), there is a high chance that the discovered models would not be Pareto optimal in practice.

\vspace{-0.25cm}
\subsection{End-to-end GCN-based latency predictor.}\label{sec:gcn}
\vspace{-0.25cm}
Our proposed end-to-end latency predictor consists of a GCN which learns models for graph-structure data~\cite{GCN_2017}. 
Given a graph $g = (V, E)$, where $V$ is a set of $N$ nodes with $D$ features and $E$ is a set of edges, 
a GCN takes as input a feature description $X \in \mathbb{R}^{N\times D}$ and a description of the graph structure as an adjacency matrix $A \in \mathbb{R}^{N\times N}$. 
For an $L$-layer GCN, the layer-wise propagation rule is the following:
\begin{equation*}
    H^{l+1} = f(H^{l},A) = \sigma\left(AH^{l}W^{l}\right),
\end{equation*}

where H$^{l}$ and $W^{l}$ are the feature map and weight matrix at the $l$-th layer respectively, and $\sigma(\bullet)$ is a non-linear activation function like ReLU.
$H^0 = X$ and $H^L$ is the output with node-level representations. 

\textbf{Architecture.}
Our GCN predictor has 4 layers of GCNs, with 600 hidden units in each layer, followed by a fully connected layer that generates a scalar prediction of the latency. The input neural network model to the GCN is encoded by an adjacency matrix A (asymmetric as the computation flow is represented as a directed graph) and a feature matrix X (one-hot encoding).
We also introduce a global node (the node that connects to all the other node) to capture the graph embedding of neural architecture by aggregating all node-level information.
GCN can handle any set of neural network models. The details of models used in this paper are in the S.M.

\textbf{Training.} All predictors are trained for 100 times, each time using a randomly sampled set of 900 models from the NAS-Bench-201 dataset. 100 random models are used for validation and the remaining 14k models are used for testing.

\textbf{Results.} Our GCN predictor outperforms existing predictors, establishing new state-of-the-art (Figure~\ref{fig:benchmark-analysis}), and demonstrates strong performance suitable for NAS (Figure~\ref{fig:gcn_vs_layerwise_acc_and_pareto}). 
In Table~\ref{tab:latency_prediction}, we show the performance of the proposed GCN latency predictor comparing to the layer-wise predictor on various devices. The values are the percentage of models with predicted latency within the corresponding error bound relative to the measured latency.
We can see that the strong performance generalizes across various devices, which have vastly different latency behaviors.
We provide an extensive study on the latency behavior on various devices in the S.M.

\begin{table}[!ht]
\small
\centering
\caption{Performance of latency predictors on NAS-Bench-201: Our GCN predictor demonstrates significant improvement over the layer-wise predictor across devices. More results in the S.M.} 
\label{tab:latency_prediction}
\begin{tabular}{@{}lcccccc@{}}
 \toprule
 \multirow{2}{*}{\parbox{1cm}{\centering Error\\bound}\hspace{-0.2cm}} & \multicolumn{3}{c}{Accuracy of GCN predictor [\%]} & \multicolumn{3}{c}{Accuracy of Layer-wise predictor [\%]} \\
 \cmidrule(rl){2-4} \cmidrule(l){5-7}
  &\!Desktop CPU\!&\hspace{-0.15cm}Desktop GPU\hspace{-0.15cm}&\hspace{-0.15cm} Embedded GPU\hspace{-0.15cm} \hspace{-0.15cm}& \hspace{-0.15cm}Desktop CPU\hspace{-0.15cm} & \hspace{-0.15cm}Desktop GPU\hspace{-0.15cm} & \hspace{-0.15cm}Embedded GPU\\
 \midrule
 $\pm$1\%  & 36.0$\pm$3.5 & 36.7$\pm$4.0 & 24.3$\pm$1.4 & 3.5$\pm$0.2 & 4.2$\pm$0.2 & 6.1$\pm$0.3 \\
 $\pm$5\%  & 85.2$\pm$1.8 & 85.9$\pm$1.9 & 82.5$\pm$1.5 & 18.2$\pm$0.4 & 17.1$\pm$0.3 & 29.7$\pm$0.8 \\
 $\pm$10\% & 96.4$\pm$0.7 & 96.9$\pm$0.8 & 96.3$\pm$0.5 & 29.6$\pm$1.1 & 32.6$\pm$1.2 & 54.0$\pm$0.8 \\
\bottomrule
\end{tabular}
\end{table}

%% file: 31-accuracy.tex
In the previous section, we assumed that the accuracy of the model is freely available during the search and focused on the latency prediction.
In practice, accuracy of the model is computationally expensive to obtain, sometimes more than latency, as it requires training. The cost of NAS critically depends on the sample efficiency, which reflects how many models need to be trained and evaluated during the search.

In this section, we ($a$) propose a new prediction-based NAS framework, called Binary Relation Predictor-based NAS (\textit{BRP-NAS} in short), that combines a GCN binary relation predictor and a novel iterative data selection strategy; ($b$) demonstrate that it vastly improves the sample efficiency of NAS for accuracy optimization.

\subsection{Transfer learning from latency predictors to improve accuracy predictors}\label{sec:acc-lat}

We demonstrate that latency predictors are surprisingly helpful in improving the training of an accuracy predictor -- both in terms of absolute accuracy and the resulting NAS performance.
The idea is that these GCN-based predictors for latency, accuracy and FLOPS have the same input representation.
A trained GCN (e.g., for latency prediction) captures features of a model which are also useful for a similar GCN trained to predict a different metric (e.g., accuracy).

To show that, we initialize the weights of an accuracy predictor with those from a latency/FLOPS predictor. We then train the predictor using the validation accuracy of CIFAR-100 dataset.
All predictors are trained for 100 times, each time using a randomly sampled set of 100 models from NAS-Bench-201. Another 100 random models are used for validation and the remaining 14k models are left for testing.
Table~\ref{tab:accuracy_prediction} shows that the quality of accuracy prediction improves in all cases, in particular, the accuracy predictors with transfer learning from FLOPS, which is freely available, can increase the sample efficiency by around 2 times. The proposed transfer learning method is applicable to any accuracy predictors with existing training techniques. We refer to the S.M. for more analysis.

\begin{table}[!ht]
\small
\centering
\caption{Performance of accuracy predictors on the test set. Standard refers to training with random initialization of the weights in GCN. Init-GPU and Init-FLOPS have the weights initialized with those of the desktop GPU latency predictor and FLOPS predictor, respectively. Transfer learning increases the sample efficiency by around 2 times.}
\label{tab:accuracy_prediction}
\begin{tabular}{@{}lccccccc@{}}
 \toprule
 \multirow{3}{*}{\parbox{1cm}{\centering Error\\bound}} & \multicolumn{7}{c}{GCN accuracy [\%]} \\
 \cmidrule(l){2-8}
 & \multicolumn{3}{c}{50 samples} & \multicolumn{3}{c}{100 samples} & 200 samples \\
 \cmidrule(rl){2-4} \cmidrule(rl){5-7} \cmidrule(l){8-8}
 & Standard & Init-GPU & Init-FLOPS & Standard & Init-GPU & Init-FLOPS & Standard \\
 \midrule
 $\pm$1\%  & 22.1$\pm$3.3 & 26.3$\pm$3.8 & 25.3$\pm$4.1 & 27.5$\pm$3.9 & 32.0$\pm$4.1 & 32.3$\pm$3.8 & 34.6$\pm$2.9  \\
 $\pm$5\%  & 72.7$\pm$3.0 & 74.8$\pm$3.4 & 73.7$\pm$3.6 & 76.9$\pm$2.4 & 80.5$\pm$2.2 & 80.5$\pm$2.7 &  81.7$\pm$1.8  \\
 $\pm$10\% & 85.4$\pm$2.4 & 87.0$\pm$2.7 & 87.2$\pm$2.4 & 88.2$\pm$1.7 & 90.4$\pm$1.6 & 91.0$\pm$2.0 & 90.8$\pm$1.3  \\
\bottomrule
\end{tabular}
\end{table}

\begin{figure}[t]
    \centering
    \includegraphics[width=0.95\textwidth]{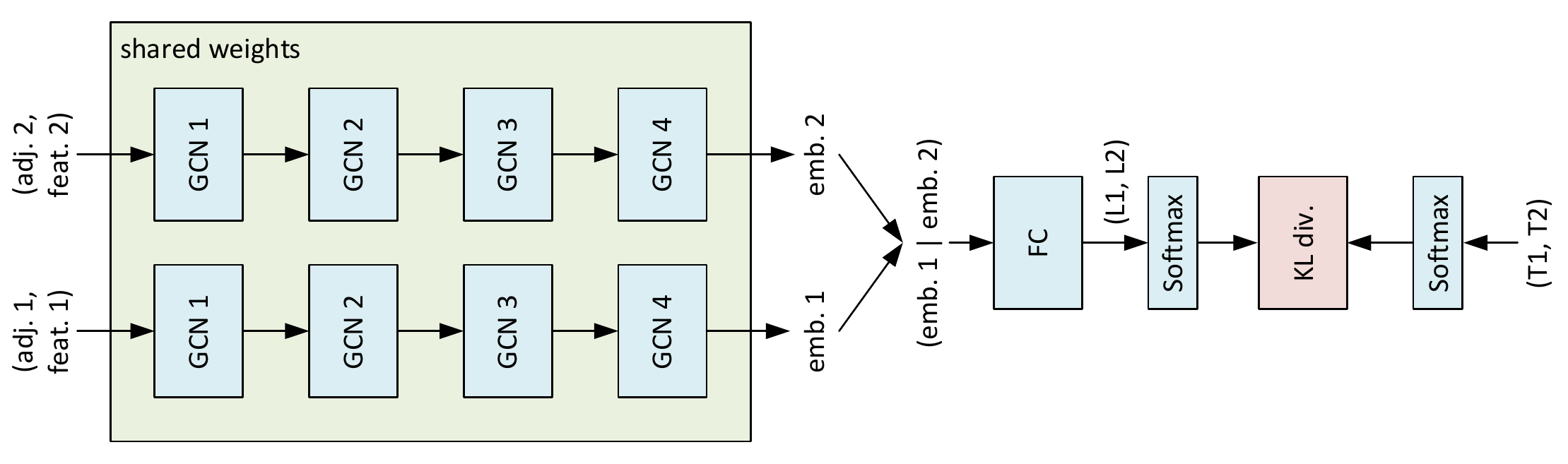}
    \caption{Overview of the proposed approach to train a binary relation predictor.} 
    \label{fig:bin_predictor}
\end{figure}

\subsection{Binary relation predictor-based NAS
} \label{sec:bin}
\vspace{-0.25cm}

We propose a new predictor-based NAS according to the following observations: 
($a$) accuracy prediction is not necessarily required to produce faithful estimates (in the absolute sense) as long as the predicted accuracy preserves the ranking of the models; 
($b$) any antisymmetric, transitive and connex \textit{binary relation} produces a linear ordering of its domain, implying that NAS could be solved by learning binary relations, where $O(n^2)$ training samples can be used from $n$ measurements; ($c$) accurately predicting the rankings of top candidates is the most important.
(We refer to the S.M. for a more formal discussion on these observations and intuition behind them.)

BRP-NAS consists of two phases. In the first phase, the ranking of all candidate models is predicted based on the outputs from a binary relation predictor, which is trained to predict the binary relation (accuracy comparisons between two models). In the second phase, based on the predicted rankings, models with high \emph{predicted} ranks are fully trained, after which, the model with the highest \emph{trained} accuracy is selected. 

\textbf{Binary relation predictor.} We propose a GCN based approach to learn a binary relation for NAS, as illustrated in Figure~\ref{fig:bin_predictor}.
The predictor reuses the GCN part of the latency predictor, without any changes, to generate graph embeddings for both input models.
The embeddings are then concatenated and passed to a fully connected layer which produces a 2-valued vector. The vector is then passed through a softmax function to construct a simple probability distribution $p = (p_1,p_2) \in \mathbb{R}^2$ with $p_1$ being the probability of the first model \textit{better} than the second, and $p_2$ being the probability of the opposite.
The produced probability distribution is then compared to the target probability distribution obtained by taking a softmax of the ground-truth accuracy of the two models $(T1, T2)$, and the objective is to minimize the KL divergence between the two distributions.
The overall network structure and the loss function are summarized in Figure~\ref{fig:bin_predictor}.

\textbf{Training via iterative data selection.} 
Given a budget of $T$ models and $I$ iterations, we start by randomly sampling and training $T/I$ models from the search space, and the results are used to train the initial version of the predictor. 
At the beginning of each subsequent iteration, we use the predictor to estimate the accuracy of all the models, denoted by $M$, in the search space.
We then select the top $\alpha * T/I$ unique models and randomly pick another $(1 - \alpha) * T/I$ models from the top $M/2^i$ models where $\alpha$ is a factor between $0$ and $1$ and $i$ is the iteration counter.
The selected $T/I$ models are trained and their resulting accuracies are used to further train the predictor for the next iteration.
Tuning $\alpha$ results in a trade-off between exploitation and exploration and we use $\alpha = 0.5$ for all our experiments. 

\begin{figure}[t]
    \centering
    \includegraphics[width=0.32\textwidth]{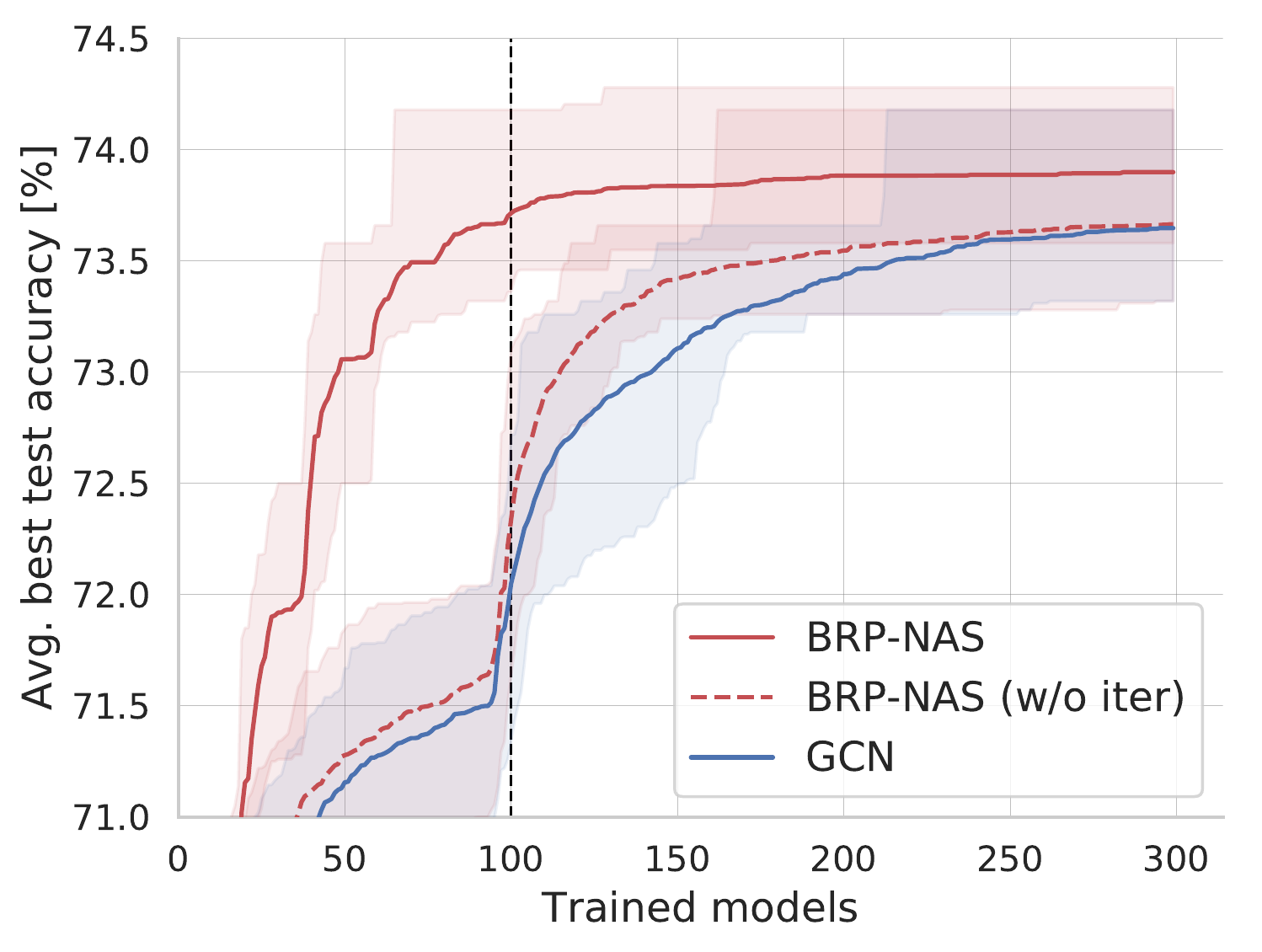} 
    \includegraphics[width=0.32\linewidth]{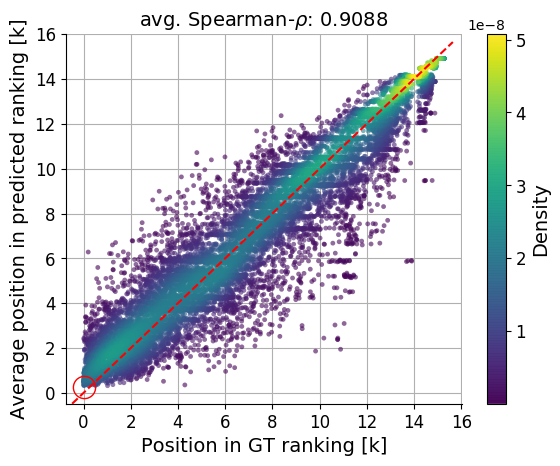}
    \includegraphics[width=0.32\linewidth]{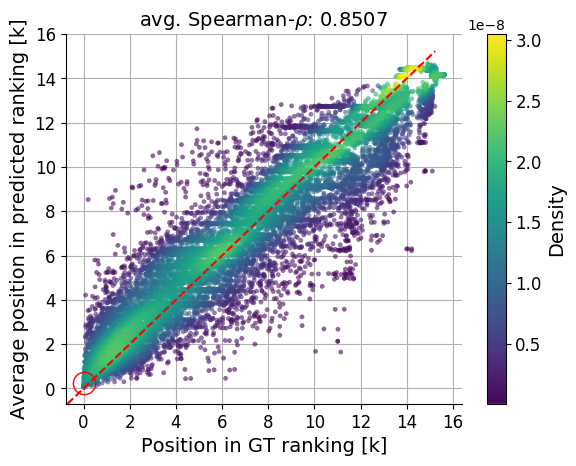}
    \caption{(Left) Binary relation prediction and iterative training significantly improve the performance of BRP-NAS over vanilla GCN approach. The dotted vertical line represents the point up to which the prediction was trained. (Middle) Ranking produced by the proposed binary relation predictor without iterative training data selection. (Right) Binary relation predictor trained via iterative data selection. 
    As indicated by the red circle in the lower-left corners, even though the plain binary predictor achieves higher overall ranking correlation with respect to the ground-truth ranking, its accuracy comes from the better fit in the lower ranking models. Iterative data selection mitigates this issue, at the expense of the global ranking quality, by focusing its training on high performing models.
    }\label{fig:ablation}
    \vspace{-0.5cm}
\end{figure}

\textbf{Results.}
Figure~\ref{fig:ablation} shows the advantages of the proposed binary relation and iterative data selection in BRP-NAS. Specifically, although the correlation between the measured ranking and the ground truth (GT) ranking decreases with iterative data selection (middle vs right), BRP-NAS is able to find better models because it focuses on high performing models only.
This greediness, together with the increased sample efficiency from using binary relations, makes our BRP-NAS significantly better than the other considered predictors.
More results (including ablation studies) can be found in the following section and in the S.M.

%% file: 32-results.tex
\begin{figure}[!ht]
    \centering
    \includegraphics[width=0.4\textwidth]{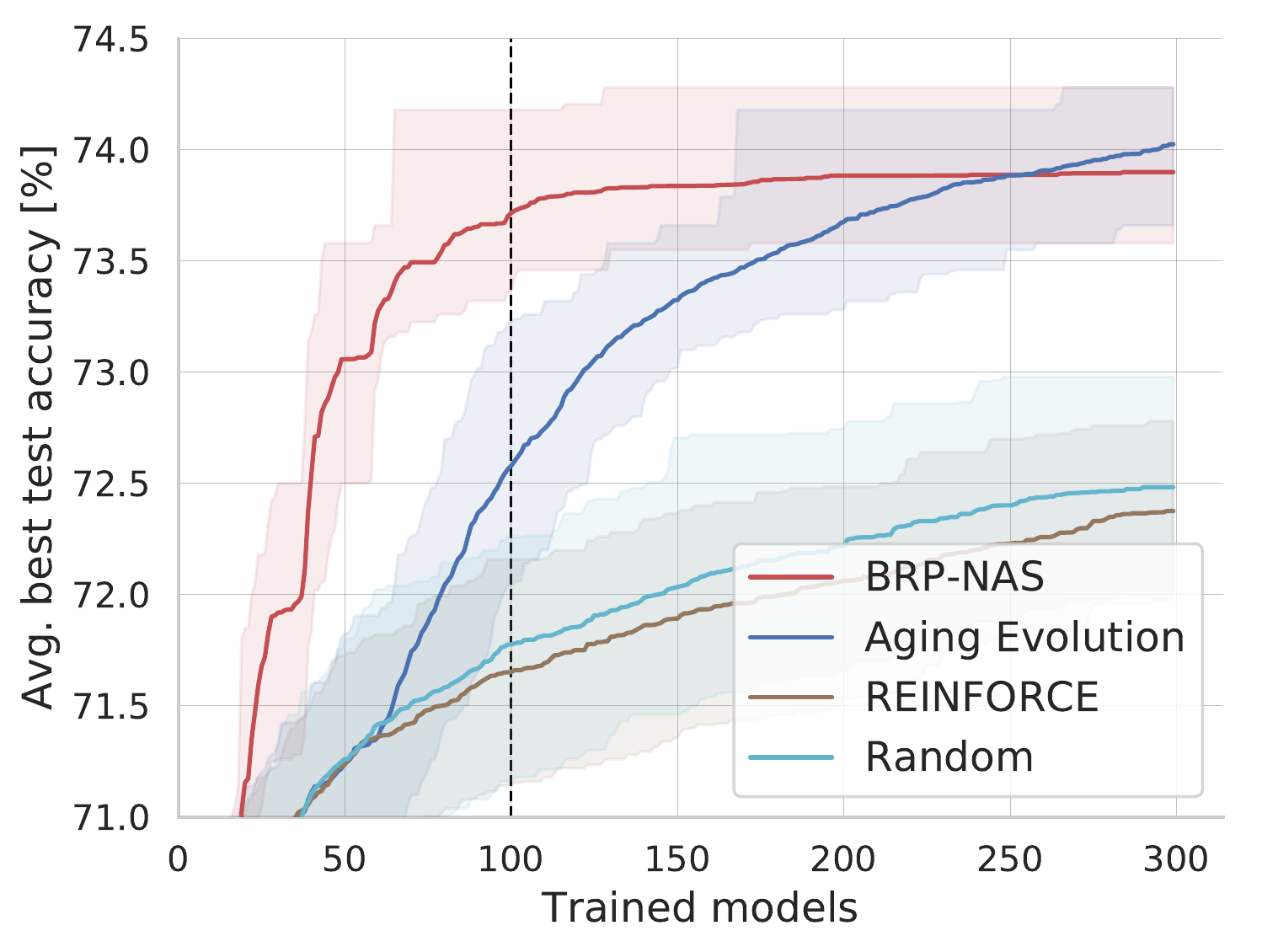}
    \includegraphics[width=0.4\textwidth]{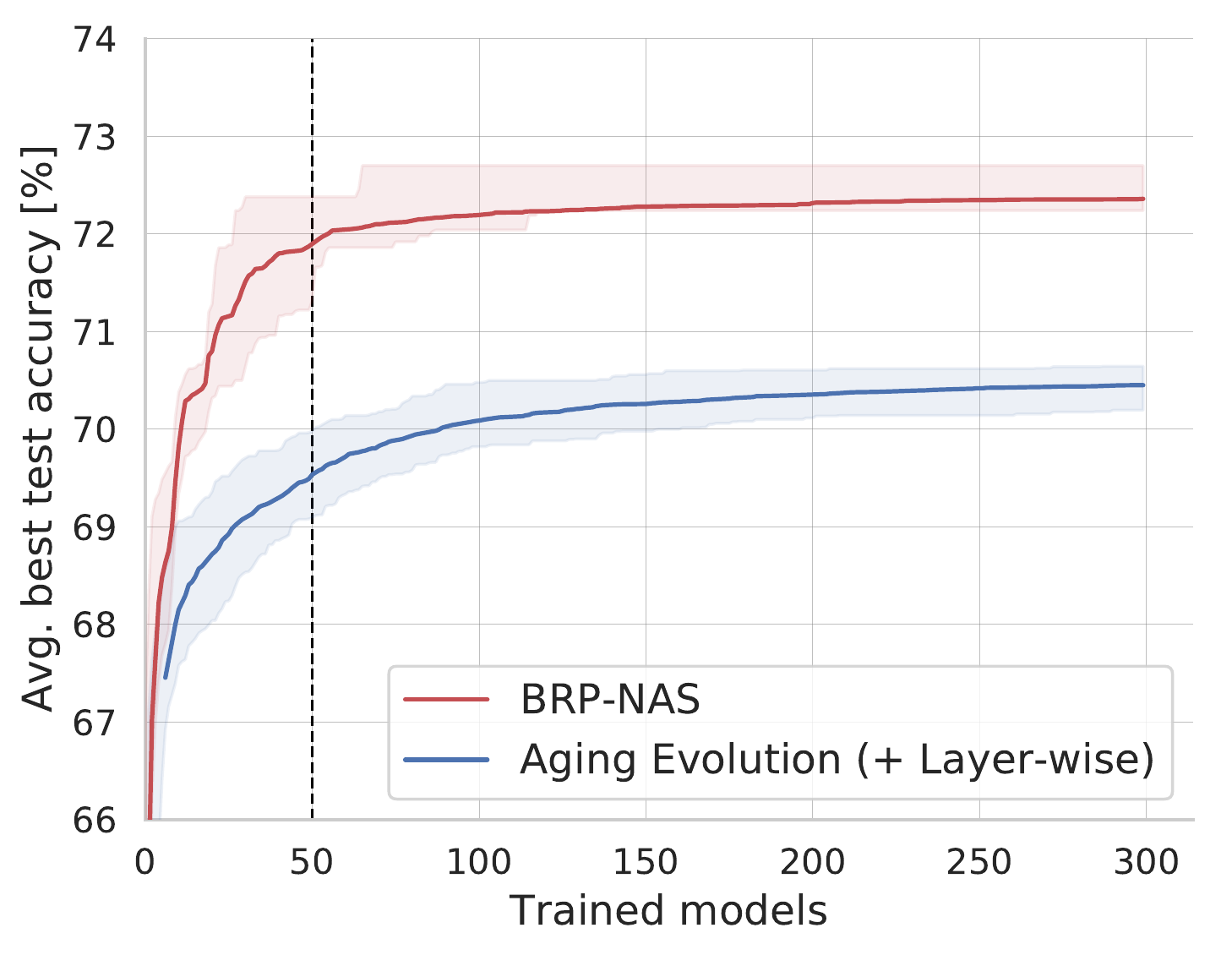}
    \caption{BRP-NAS outperforms Aging Evolution and other popular search methodologies on NAS-Bench-201, in a setting without latency constraints (left), and with a 5ms latency constraint on a desktop GPU (right).}
     \label{fig:comparison}
     \vspace{-0.5cm}
 \end{figure}

\textbf{Comparison to prior work on NAS-Bench-201.}
Figure~\ref{fig:comparison} (left) shows results in an unconstrained setting where we compare to AE~\cite{Real2019RegularizedEF}, REINFORCE~\cite{NASinitial_2017}, and random sampling.
\ours{} outperforms other methods by being more than $2x$ more sample efficient than the runner up (AE).
However, we also see that eventually the average performance of AE surpasses \ours{} -- we investigated that and observed that this difference comes from the SGD-induced randomness.
Specifically, unlike ours algorithm which trains each model only once, AE can (and often does) train the same model multiple times, which allows it to find good seeds more robustly.
Please note that this only happens after sufficiently many models have been trained and in the presence of high SGD noise, therefore constituting an edge case -- for example, our method still consistently outperforms AE on a larger NAS-Bench-101 and in a constrained setting.
More information about SGD-induced randomness can be found in the S.M. 

We then combined BRP-NAS together with our GCN-based latency predictor from Section~\ref{sec:latency} to further improve NAS in the constrained settings.
Similar to Figure~\ref{fig:terrible-NAS}, we compare to AE with a layer-wise predictor (current SOTA for latency prediction) -- the results, shown in Figure~\ref{fig:comparison} (right), demonstrate that the naive combination is far from optimal.

\textbf{Comparison to prior work on NAS-Bench-101.}
To gauge the efficiency and quality of our \ours,
we compare to previously published work (unconstrained case only).
Simultaneously, this comparison shows how our technique scales to larger benchmarks as NAS-Bench-101~\cite{nasbench1} contains over 423k points.
Table~\ref{tab:nas1_comp} shows the result of this comparison, both in terms of the total number of trained models (algorithm speed) and final test accuracy on CIFAR-10 dataset (algorithm quality).
We train the \ours{ }predictor using the validation accuracy from 100 models, then we train the top 40 models returned by the predictor.
With a total of 140 trained models, we are able to find a more accurate model faster than any previous work, as highlighted in Table~\ref{tab:nas1_comp} (detailed results in the S.M.).

\vspace{-0.25cm}
\begin{table}[!ht]
\small
\centering
\caption{Comparison to prior work on NAS-Bench-101 dataset. Our result is averaged from 32 runs.}
\label{tab:nas1_comp}
\begin{tabular}{@{}lccccccc@{}}
\toprule
  & NAO~\cite{NAO_2018_NIPS}& BONAS~\cite{BONAS_2020} & AE~\cite{Real2019RegularizedEF}  &  Wen et al. ~\cite{1912.00848} & NPENAS~\cite{NPENAS2020} & \ours{} \\
\midrule
  Trained models & 1000& 1000 & 418 & 256 & 150 & \textbf{140} \\
  Test Acc. [\%] & 93.81 & \textbf{94.22} & \textbf{94.22} & 94.17 & 94.14 & \textbf{94.22}  \\
\bottomrule
\end{tabular}
\end{table}

\textbf{Results on the DARTS search space.}
We further test our binary accuracy predictor on a much larger DARTS search space~\cite{darts}, performing classification on CIFAR-10.
The first challenge we had to face was related to its size -- with approximately $10^{18}$ models, it is infeasible to simply sort all of them.
Because of that, we decided to begin a search by randomly selecting a subset of 10 million architectures and then run everything else analogically to the previous scenarios (20 models trained in each iteration, up to 60 models trained in total).
We repeated the entire process 3 times, results are presented in Table~\ref{tab:darts}.
It is worth noticing that for a small search budget (6 GPU days), our BRP-NAS is in practice analogical to random search (since models trained in the first iteration are random).
This suggests that the difference observed between different approaches in the "small budget" group is actually noise introduced by the random nature of algorithms.
However, we can see that as we increase the number of trained models, our approach keeps improving results consistently, suggesting that the predictor is able to learn useful features and can robustly outperform DARTS on its search space.
Further details and comments are provided in the S.M.

\vspace{-0.25cm}
\begin{table}[!ht]
\small
\centering
\caption{Results on the DARTS search space for CIFAR-10 classification with different budgets.}
\label{tab:darts}
\begin{tabular}{@{}lcccccc}
\toprule
  & \multicolumn{3}{c}{Small budget} & \multicolumn{2}{c}{Medium budget} & High budget \\
  \cmidrule(r){2-4} \cmidrule(rl){5-6} \cmidrule(l){7-7}
                          & DARTS  & Random & BRP-NAS & Random & BRP-NAS & BRP-NAS \\
\midrule
  Cost (GPU days)  & 6                 & 6              & 6              & 30            & 30            & 60 \\
  Test error [\%]  & 2.76$\pm$0.09     & 2.87$\pm$0.06  & 2.71$\pm$0.07  & 2.75$\pm$0.18 & 2.66$\pm$0.09 & 2.59$\pm$0.11 \\
\bottomrule
\end{tabular}
\flushleft
\end{table}
\vspace{-0.25cm}

%% file: 40-benchmark.tex
In this section, we present LatBench -- a latency dataset of NAS-Bench-201 models on a wide range of devices. 
Similar to the motivations of NAS-Bench-101 and NAS-Bench-201, we aim towards ($a$) reproducibility and comparability in hardware-aware NAS and ($b$) ameliorating the need for researchers to have access to a broad range of devices. 
Although Nas-Bench-201 provides computational metrics such as number of parameters, FLOPS, and latency, these metrics are computed with operations and skip connections that do not contribute to the resulting output, leading to inaccurate measurements. 
Additionally, as latencies among devices often have weak correlations, more devices are required to facilitate the research of hardware-aware NAS.

We first remove any dangling nodes and edges and run each model in NAS-Bench-201 on the following devices:
\textit{(i)} Desktop CPU - Intel Core i7-7820X,
\textit{(ii)} Desktop GPU - NVIDIA GTX 1080 Ti,
\textit{(iii)} Embedded GPU - NVIDIA Jetson Nano,
\textit{(iv)} Embedded TPU - Google EdgeTPU,
\textit{(v)} Mobile GPU - Qualcomm Adreno 612 GPU,
\textit{(vi)} Mobile DSP - Qualcomm Hexagon 690 DSP.

Specifically, we run each model $1000$ times on each aforementioned non-mobile device using a patch size of $32\times 32$ and a batch size of $1$. For mobile devices, each model is run $10$ time with the same settings.
In order to lessen the impact of any startup/cool-down effects such as the creation and loading of inputs into buffer, we discard latencies that fall outside the lower and higher quartile values before taking the average of every $10$ runs.
These averages are discarded again with the aforementioned thresholds before a final average is taken.

For more details and further analysis of LatBench, please refer to S.M.
We also provide the FLOPS and the number of parameters for these models after removing unneeded nodes and edges.
We have plans of updating LatBench by adding more devices in the future.

%% file: 50-limitations.tex
\textbf{Generalization to other metrics and problems.}
Even though we focused our work on finding the most accurate architecture under a strict latency limit, we expect that our observations are still relevant under different settings.
Generally speaking, our searching procedure can be described as an efficient predictor-based approach to solving multi-objective optimization with $\epsilon$-constraint method, where the emphasis is put on: \textit{(a)} distinguishing between the constraining metric (latency in our case) and the optimised metric (accuracy), and \textit{(b)} using different variations of predictors for them (unary and binary, respectively).
Consequently, this approach can be used for other multi-objective optimisation problems.
Furthermore, without distancing from NAS, the latency prediction part can naturally be extended to other hardware-related metrics.

\textbf{Sensitivity to design choices.}
When designing our GCN predictor, we have tried different design choices -- forward propagation (with a global sink) vs backward propagation (with a global source), and softmax vs sigmoid activations. They all lead to similar performance.
We also notice that the performance of predictor is better without normalizing the adjacency matrix.

\textbf{Symmetry and transitivity of the learned relation.}
Even though we motivate usage of the binary predictor by the fact that antisymmetric and transitive binary relations produce linearly ordered sets, we do not directly enforce any of these properties.
Antisymmetry is enforced indirectly by including both orderings of pairs -- $(m_1,m_2)$ and $(m_2,m_1)$ -- in the predictor's training set.
We empirically checked that on a randomly sampled set of 1000 models from the NAS-Bench-201 space (i.e., $\sim$500k pairs) 98\% of cases remain antisymmetric.
To check transitivity, we investigated the number of simple cycles in a relation matrix for the same 1000 random models and found out that this number is exceptionally large (we stopped the program after reaching 10 million cycles).
This suggests that, in general, our predictor does not guarantee transitivity, despite producing strong empirical results.
We consider investigating the impact of cycles on NAS to be an interesting direction for future work.

%% file: 60-conclusion.tex
We introduced \ours, a new prediction-based NAS framework that combines a binary relation accuracy predictor architecture and an iterative data selection strategy to improve the performance of NAS. BRP-NAS outperforms previous NAS methods in both sample efficiency and accuracy for NAS-Bench-101 and NAS-Bench-201, and also surpasses DARTS in its search space.
We release LatBench -- a latency dataset for models in the NAS-Bench-201, and Eagle -- a tool to measure and predict performance of models on different devices.

%% file: 70-appendix.tex
\beginsupplement

\input{appendix/A1-latency}
\input{appendix/A2-accuracy}
\input{appendix/A3-results}
\input{appendix/A4-benchmark}

%% file: appendix/A1-latency.tex
\section{Supplementary Material for Section~\ref{sec:latency}: Latency prediction in NAS}\label{sec:app:latency}

\subsection{Neural network models supported by GCN predictor} 

GCN can handle any set of neural network models. In this paper, we apply GCN to NAS-Bench-201 and NAS-Bench-101. Their network structures are described below.

In NAS-Bench-201, the skeleton of any model consists of 3 stacks of 5 cells with a fixed structure and placeholders for 6 operation nodes.
In the original paper, this cell structure is described with the help of a directed acyclic graph whose nodes and edges represent tensors and data dependencies between them, respectively.
Additionally, each edge is also assigned a label which defines the operation to apply to the source tensor and whose result is used to define the content of the destination tensor.
Since the structure of the cell is fixed, the only "searchable" part of the cell are labels to be assigned to the edges -- the authors considered 5 different options for each label: "zero" operation, "identity" operation (a.k.a. skip-connection), convolution $3\times 3$, convolution $1\times 1$, and $3\times 3$ average pooling.
Therefore, each architecture in NAS-Bench-201 can be defined by selecting 6 elements (with repetitions) from the aforementioned set of operations ( $O_i$ for $i=1...6$) and represented with an architecture string: \textsc{|$O_1$\textasciitilde 0|+|$O_2$\textasciitilde 0|$O_3$\textasciitilde 1|+|$O_4$\textasciitilde 0|$O_5$\textasciitilde 1|$O_6$\textasciitilde 2|} (as defined by the NAS-Bench-201 authors).

\begin{figure}[!ht]
    \centering
    \includegraphics[width=0.9\textwidth]{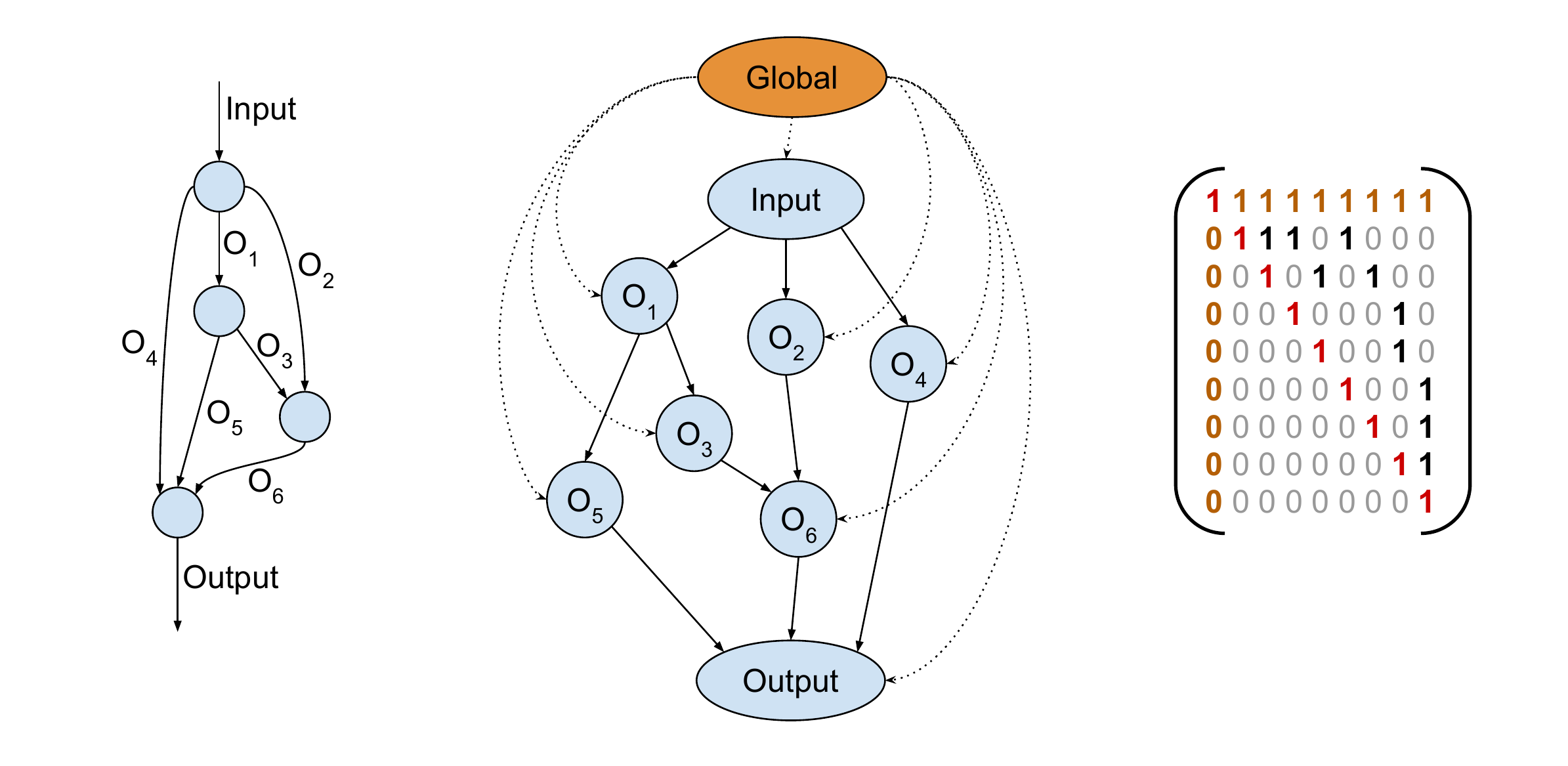}
    \caption{(Left) Graph representation of a cell used in NAS-Bench-201 models, as defined by the authors, (Middle, Right) equivalent representation (with additional global node) and its adjacency matrix (without considering optimizations to remove "zero" and "skip-connect" operations) which are used in this paper.}
    \label{fig:nasbench201_graphs}
\end{figure}

For the purpose of this work, we have modified the representation of the models in NAS-Bench-201 dataset in the following way:
\begin{itemize}
    \item When constructing the graph representation of a network to use it with our GCN predictors, we begin by converting the NAS-Bench-201 cell graph (Figure~\ref{fig:nasbench201_graphs} left) into its equivalent form using more traditional convention where nodes represent operations\footnote{One of the reasons behind this decision was to have a unified network representation consistent between NAS-Bench-101 and 201.} (Figure~\ref{fig:nasbench201_graphs} middle);
    \item We optimize the graph by completely detaching "zero" and "skip-connect"\footnote{For skip-connections, before detaching we make sure that all direct predecessors of the node are instead connected directly to all direct successors of the node.} operations, and then removing all other nodes which became dangling (i.e., they do not lay on the path from input to output) because of the previous step;
    \item As mentioned in Section~\ref{sec:gcn}, we add a global node which is connected to all other nodes (including the nodes which were detached due to optimizations) and also add self-connections for all nodes -- this results in an adjacency matrix (Figure~\ref{fig:nasbench201_graphs} right) with dimensions $9\times 9$ (6 operation nodes, "input", "output" and "global" node) which is one of the inputs to the GCN;
    \item Finally, for each node we construct its feature vector by encoding the node's type using a one-hot vector -- because "zero" and "skip-connect" operations were optimized out\footnote{Since the optimized nodes are actually still present in the graph (but detached from anything else) we simply considered them "typeless" and assign zeros-only vectors to them.}, the possible choices are: the three remaining operations plus "input", "output" and "global" node types -- thus the feature matrix is $9\times 6$.
\end{itemize}

In NAS-Bench-101, modules are represented by directed acyclic graphs with up to 7 nodes. The valid operations at each node are convolution $3\times 3$, convolution $1\times 1$, and $3\times 3$ max pooling. This results in an adjacency matrix with dimensions $8\times 8$ (5 operation nodes, "input", "output" and "global" node) and a feature matrix with dimensions $8\times 6$ (3 operations plus "input", "output" and "global" node types).

\subsection{Latency predictor for various devices}

\begin{figure}[!ht]
    \centering
    \begin{tabularx}{\linewidth}{CCC}
        \begin{subfigure}[b]{.3\textwidth}
        \includegraphics[width=\linewidth]{images/30-latency/10-latency-proxies/desktop_cpu_gcn.png}
        \caption{GCN - Desktop CPU}
        \end{subfigure}
        \begin{subfigure}[b]{.3\textwidth}
        \includegraphics[width=\linewidth]{images/30-latency/10-latency-proxies/desktop_cpu_layerwise.png}
        \caption{Layer-wise - Desktop CPU}
        \end{subfigure}
        &
        \begin{subfigure}[b]{.3\textwidth}
        \includegraphics[width=\linewidth]{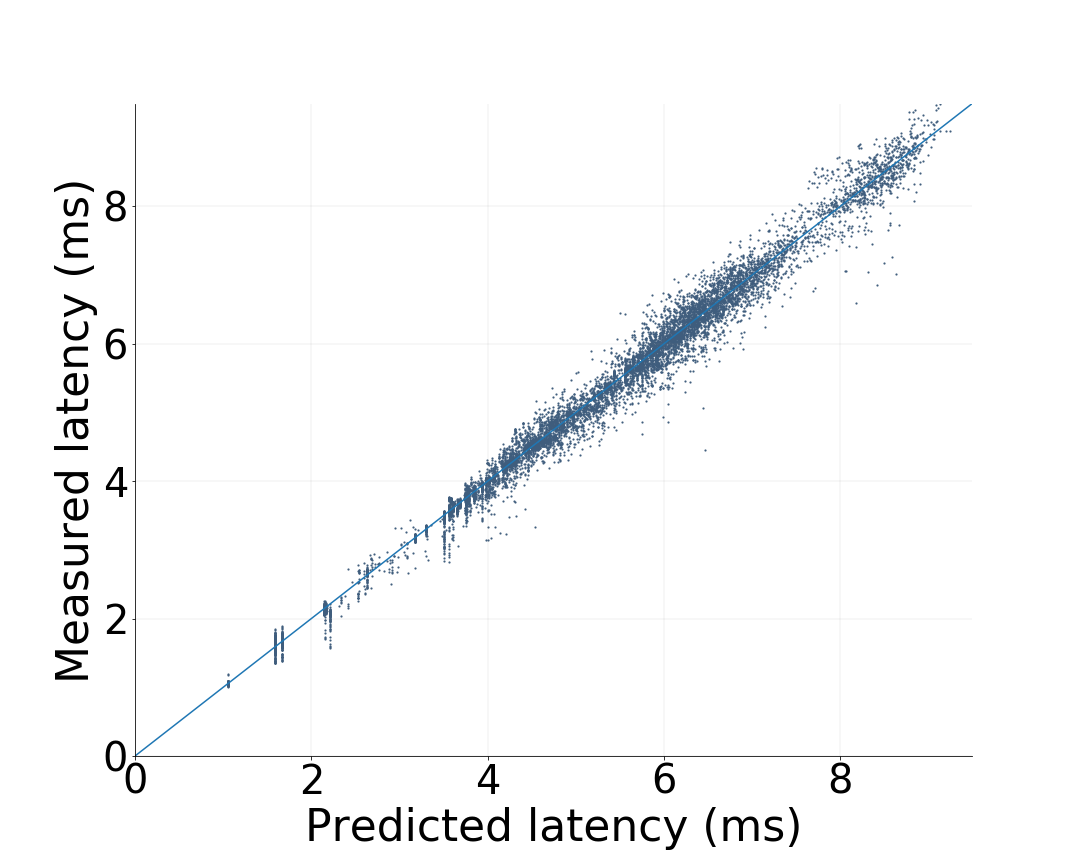}
        \caption{GCN - Desktop GPU}
        \end{subfigure}
        \begin{subfigure}[b]{.3\textwidth}
        \includegraphics[width=\linewidth]{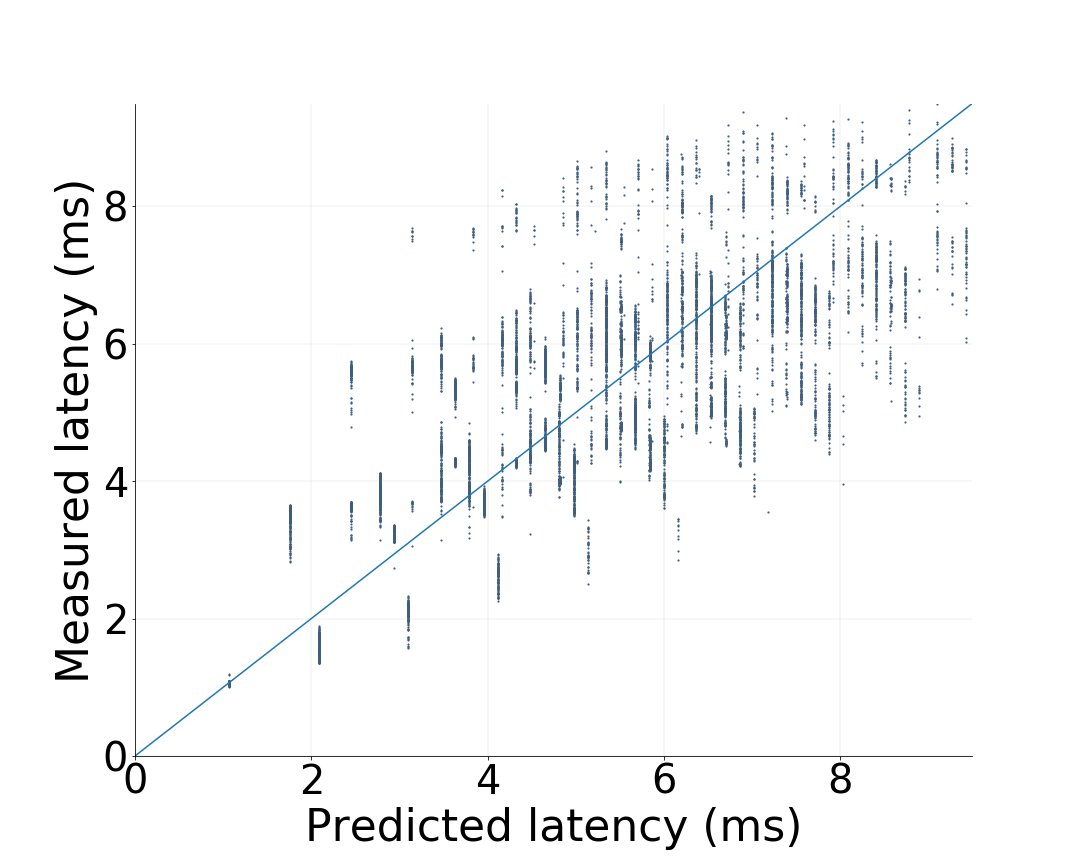}
        \caption{Layer-wise - Desktop GPU}
        \end{subfigure}
        &
        \begin{subfigure}[b]{.3\textwidth}
        \includegraphics[width=\linewidth]{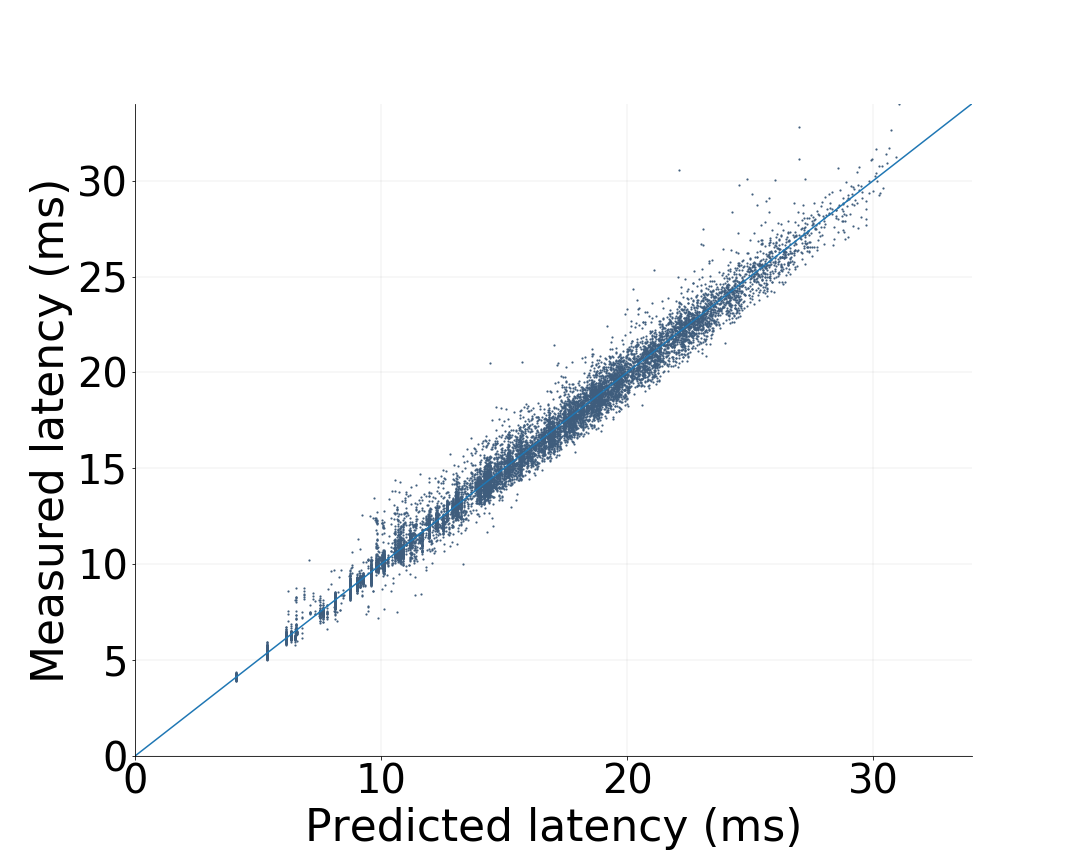}
        \caption{GCN - Embedded GPU}
        \end{subfigure}
        \begin{subfigure}[b]{.3\textwidth}
        \includegraphics[width=\linewidth]{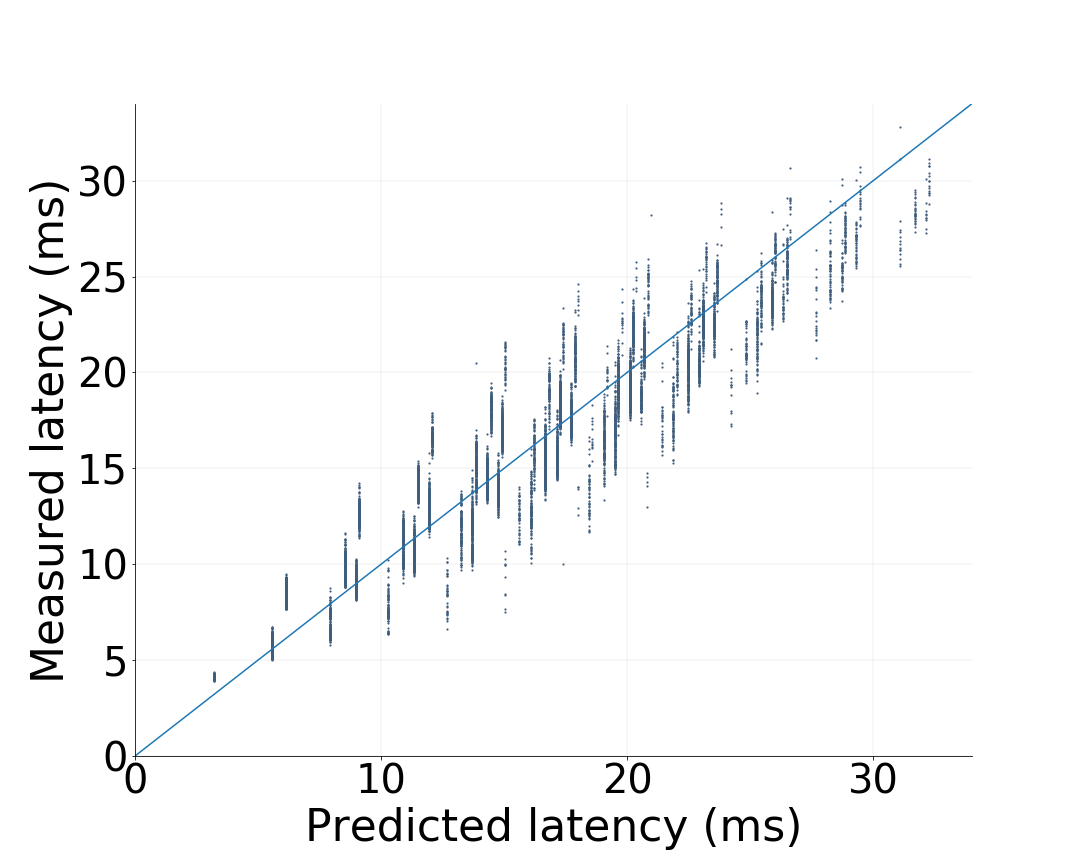}
        \caption{Layer-wise - Embedded GPU}
        \end{subfigure}
    \end{tabularx}
    \caption{Performance of latency predictors of various devices on NAS-Bench-201}
    \label{fig:layerwise_vs_gcn}
\end{figure}

Table~\ref{tab:latency_prediction_extended} and Figure~\ref{fig:layerwise_vs_gcn} show the performance of the proposed GCN predictor.
We first train each predictor for 100 times, each time using the hyperparameters summarized in Table~\ref{tab:latency_predictor_hyperparameters} and a randomly sampled set of 900 models in the NAS-Bench-201 dataset. 100 models are used for validation and the remaining 14k models are used for testing.
Values reported in Table~\ref{tab:latency_prediction_extended} are the percentage of models in the test set that the predicted latency is within the corresponding error bound of the measured latency.
The GCN predictors generalize well to unseen models in the NAS-Bench-201 dataset, and significantly outperform layer-wise predictors.
We also see that the strong performance generalizes across various devices, which have vastly different latency behaviors.
Then we experiment with training the GCN predictors using a smaller randomly sampled set of 100 models. The performance degrades but still outperform the layer-wise predictors.

\begin{table}[!ht]
    \caption{Training hyperparameters of the latency predictors.}
    \centering
    \begin{tabular}{l|l}
    \toprule
         Batch size & 10 \\
         Learning rate schedule & plateau \footnotesize{(reduce learning rate by half if no improvement is seen for 10 epochs)} \\
         Initial learning rate & 0.0008 \\
         Optimizer & AdamW \\
         L2 weight decay & 0.0005 \\
         Dropout ratio & 0.002 \\
         Training epochs & 250 \footnotesize{(early stopping patience of 35 epochs)} \\
    \bottomrule
    \end{tabular}
    \label{tab:latency_predictor_hyperparameters}
\end{table}

\begin{table}[!ht]
\caption{Performance of latency predictors of various devices on NAS-Bench-201: (i) D. CPU - Intel Core i7-7820X, (ii) D. GPU - NVIDIA GTX 1080 Ti, (iii) E. GPU - NVIDIA Jetson Nano, (iv) E. TPU - Google EdgeTPU, (v) M. GPU - Qualcomm Adreno 612 GPU, (vi) M. DSP - Qualcomm Hexagon 690 DSP.}
\label{tab:latency_prediction_extended}
\centering
\small
\begin{tabular}{@{}lcccccc@{}}
\toprule
\multirow{2}{*}{\parbox{1.5cm}{\centering \textbf{Error\\bounds}}} & \multicolumn{6}{c}{\textbf{Accuracy [\%]}} \\
\cmidrule(l){2-7}
& \textbf{D. CPU} & \textbf{D. GPU} & \textbf{E. GPU} & \textbf{E. TPU} & \textbf{M. GPU} & \textbf{M. DSP} \\
\midrule
\multicolumn{2}{@{}l}{\textbf{GCN} \textit{\scriptsize(900 pts.)}} & & & & \\
\cdashlinelr{1-7}
 $\pm$1\%   & 36.0$\pm$3.5 & 36.7$\pm$4.0 & 24.3$\pm$1.4 & 16.2$\pm$3.6 & 17.5$\pm$2.8 & 21.3$\pm$1.9 \\
 $\pm$5\%   & 85.2$\pm$1.8 & 85.9$\pm$1.9 & 82.5$\pm$1.5 & 64.0$\pm$5.7 & 67.5$\pm$7.4 & 77.5$\pm$2.6 \\
 $\pm$10\%  & 96.4$\pm$0.7 & 96.4$\pm$0.7 & 96.3$\pm$0.5 & 87.4$\pm$2.7 & 90.5$\pm$5.5 & 94.2$\pm$0.4 \\

\midrule
\multicolumn{2}{@{}l}{\textbf{GCN} \textit{\scriptsize(100 pts.)}} & & & & \\
\cdashlinelr{1-7}
 $\pm$1\%   & 6.1$\pm$1.7 & 5.9$\pm$1.3 & 9.9$\pm$1.3 & 6.2$\pm$1.0 & 5.2$\pm$0.9 & 10.3$\pm$1.1 \\
 $\pm$5\%   & 27.9$\pm$5.5 & 28.7$\pm$3.6 & 44.6$\pm$4.0 & 30.0$\pm$3.6 & 24.9$\pm$3.4 & 48.0$\pm$3.8 \\
 $\pm$10\%  & 51.5$\pm$8.3 & 52.9$\pm$5.0 & 71.8$\pm$3.5 & 54.6$\pm$5.7 & 46.3$\pm$4.1 & 78.4$\pm$3.6 \\

\midrule
\multicolumn{2}{@{}l}{\textbf{Layer-wise} \textit{\scriptsize(900 pts.)}} & & & & \\
\cdashlinelr{1-7}
 $\pm$1\%   &  3.5$\pm$0.2 &  4.2$\pm$0.2 &  6.1$\pm$0.3 & N/A & N/A & N/A \\
 $\pm$5\%   & 18.2$\pm$1.5 & 17.1$\pm$0.3 & 29.7$\pm$0.8 & N/A & N/A & N/A \\
 $\pm$10\%  & 29.6$\pm$1.1 & 32.6$\pm$1.2 & 54.0$\pm$0.8 & N/A & N/A & N/A \\
\bottomrule
\end{tabular}
\end{table}

\subsection{Oracle NAS} 

\begin{figure}
    \centering
    \includegraphics[width=0.9\textwidth]{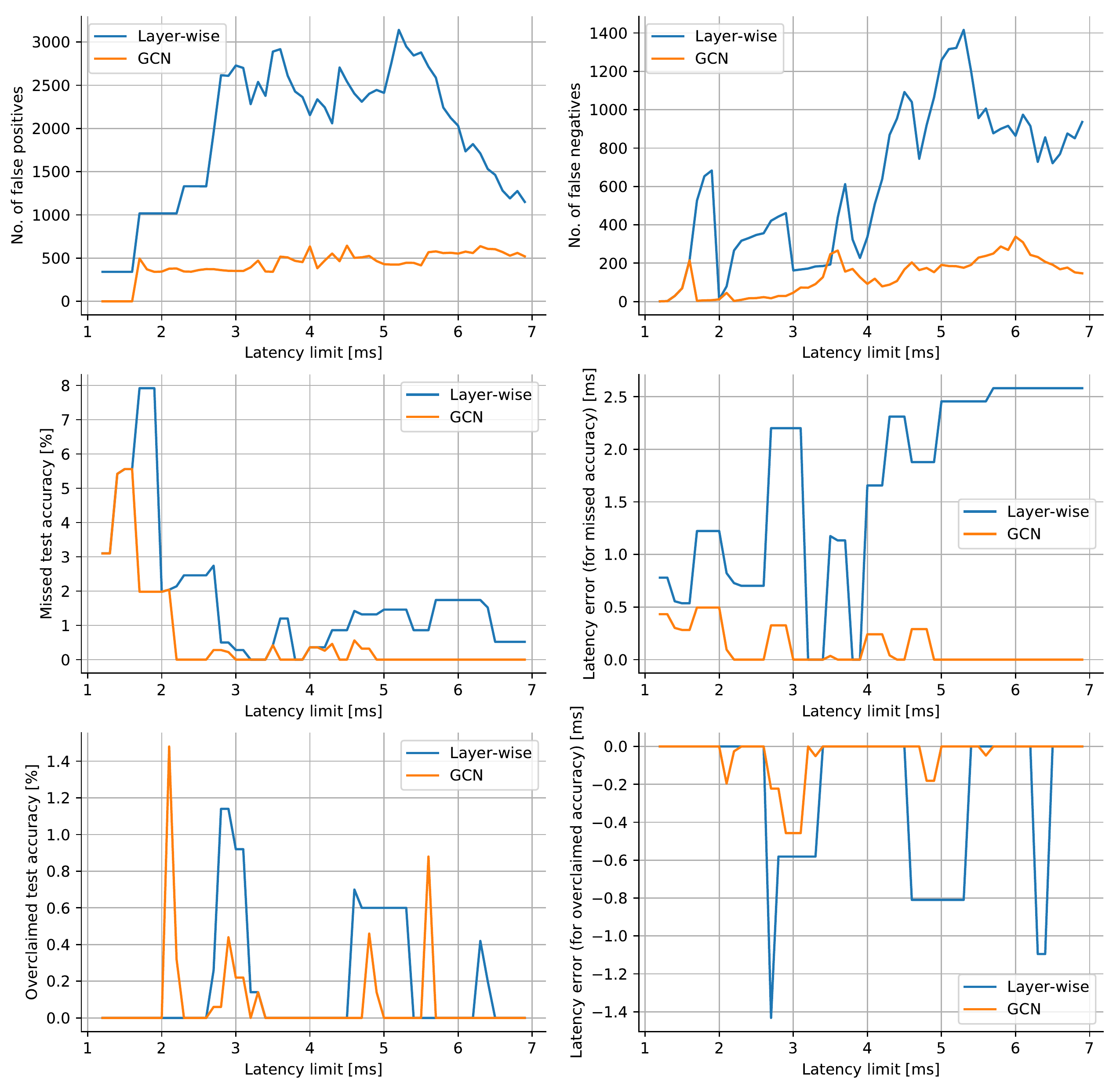}
    \caption{Oracle NAS results for desktop GPU, using GCN and Layer-wise latency predictors.
    Results are obtained on the NAS-Bench-201 dataset for desktop-GPU using both the layer-wise and GCN-based latency predictors.
    All latency thresholds are between 1-7ms with a step size of 0.1ms.}
    \label{fig:oracle-nas}
\end{figure}

In this section, we provide a detailed description of oracle NAS and the comparison between the layer-wise latency predictor-based oracle NAS and our GCN latency predictor-based oracle NAS (Figure~\ref{fig:oracle-nas}). As noted in Section~\ref{sec:existing_latency_predictors}, in order to analyze the error introduced by an inaccurate latency estimation on NAS, we consider a set of experiments where a perfect searching algorithm, denoted by \emph{oracle NAS}, is used to find the best possible model under a varying latency threshold.
Here "perfect" means that the algorithm has the full knowledge about accuracy of all points and is always able to find the most accurate one\footnote{To simplify this process, we did not consider the SGD noise and instead only used accuracy values from NAS-Bench-201 with seed 888.}, but its knowledge about latency of different models is potentially limited by how latency is estimated.
For each latency threshold $l_{th}$, we begin by discarding all models which are believed to be too expensive according to the latency predictor in use.
We then obtain the best model out of those which are left using our oracle search -- this model is the \textit{assumed best} but we are still not sure because it might have been falsely accepted due to imperfect latency estimation.
Therefore we re-validate the assumed best model, this time using its measured latency, and only accept it if it truly falls below the latency limit.
Otherwise we call it a \textit{false positive} and discard it, repeating the aforementioned process with the second best point according to the initial search result.
The first model encountered during this re-validation phase whose latency falls below the threshold is called the \textit{effective best} for a given predictor with latency limit, and the effective best of a search when the ground-truth measured latency is always used is called \textit{ground-truth best}.

As shown in Figure~\ref{fig:oracle-nas}, we extensively study the difference between the assumed best and the effective best (introduced by false positives), as well as the difference between the accuracy of the ground-truth best and the effective best (introduced by false negatives) and some other accompanying metrics.
Formally, for a set of models $\mathbb{S}$, a latency threshold $l_{th}$, latency predictor $\text{pred}(\cdot)$ and measured latency $\text{lat}(\cdot)$, we can define:
(i) the set of false positives: $\{ s\; |\; s\in \mathbb{S} \land \text{pred}(s) < l_{th} \land \text{lat}(s) > l_{th} \}$; (ii) the set of false negatives: $\text{pred}(s) > l_{th} \land \text{lat}(s) < l_{th}$; (iii) analogically true positives/negatives if comparison with $l_{th}$ is consistent between $\text{pred}(\cdot)$ and $\text{lat}(\cdot)$; (iv) and finally the set of truly positive points when $\text{lat}(s) < l_{th}$.

Let us denote the set of: false/true negatives as $\mathbb{N}_f$ and $\mathbb{N}_t$ respectively, analogically false and true positives as $\mathbb{P}_f$ and $\mathbb{P}_t$, and the set of truly positives as $\mathbb{P}$.
The \textit{assumed best} is defined as: $s^{\star}_f = \argmax_{s\in \mathbb{P}_f\cup\mathbb{P}_t}\text{accuracy}(s)$;
\textit{effective best} as: $s^{\star}_p = \argmax_{s\in \mathbb{P}_t}\text{accuracy}(s)$;
and \textit{ground-truth best} as: $s^{\star} = \argmax_{s\in \mathbb{P}}\text{accuracy}(s)$.
Then, Figure~\ref{fig:oracle-nas} shows the following metrics as functions of latency threshold:

\begin{itemize}
    \item Top left: the number of \emph{false positives} denotes how many models were considered below the limit incorrectly, i.e., $|\mathbb{P}_f|$.
    \item Top right: the number of \emph{false negatives} denotes how many models were considered above the limit incorrectly, i.e., $|\mathbb{N}_f|$.
    \item Middle left: the \emph{missed accuracy} denotes the accuracy difference between the ground-truth best model and the effective best model, i.e., 
        $\text{accuracy}(s^{\star}) - \text{accuracy}(s^{\star}_p)$.
    \item Middle right: the related \emph{latency prediction error} if a model was missed, i.e.,
        $\begin{cases}
            \text{pred}(s^{\star}) - \text{lat}(s^{\star}) & \text{if } s^{\star} \ne s^{\star}_p \\
            0 & \text{otherwise}
        \end{cases}$
    \item Bottom left: the \emph{over-claimed accuracy} denotes the accuracy difference between the assumed best model (i.e., including false positives) and the effective best model after removing false positives, i.e., 
        $\text{accuracy}(s^{\star}_f) - \text{accuracy}(s^{\star}_p)$
    \item Bottom right: the related \emph{latency prediction error} if a model was over-claimed, i.e.,
        $\begin{cases}
            \text{pred}(s^{\star}_f) - \text{lat}(s^{\star}_f) & \text{if } s^{\star}_f \ne s^{\star}_p \\
            0 & \text{otherwise}
        \end{cases}$
\end{itemize}

%% file: appendix/A2-accuracy.tex
\section{Supplementary Material for Section~\ref{sec:accuracy}: Accuracy prediction in NAS}\label{sec:app:accuracy}

\subsection{Transfer learning from latency predictors to improve accuracy predictors}\label{sec:app:transfer_learning}

\begin{figure}[!ht]
    \centering
    \includegraphics[width=0.32\textwidth]{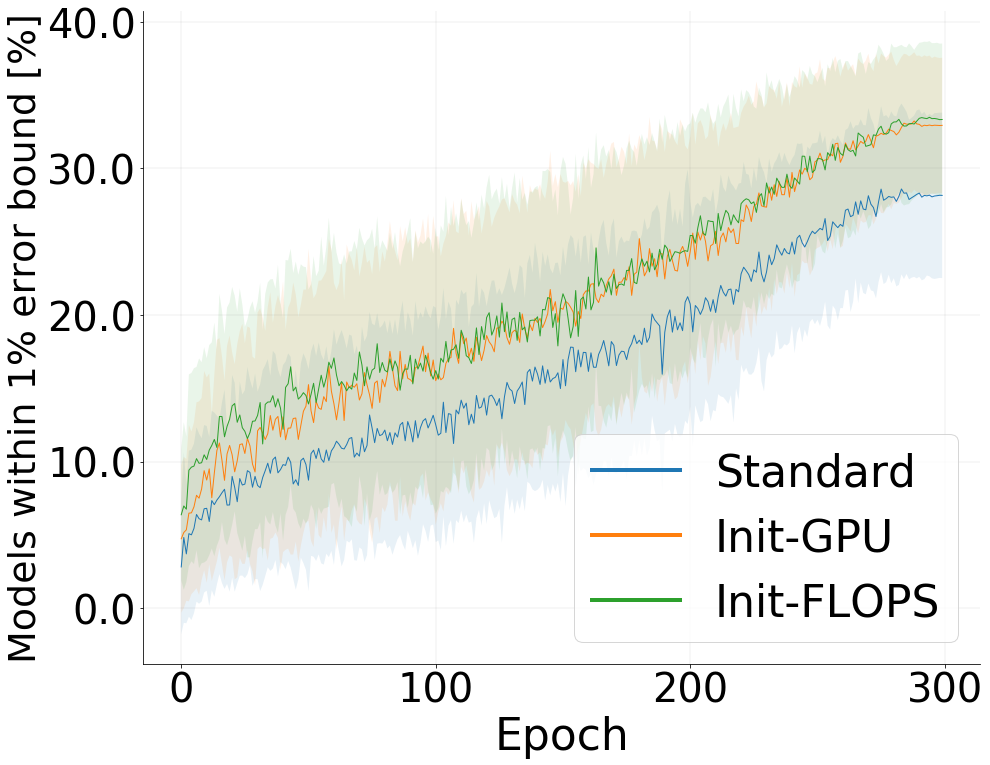}
    \includegraphics[width=0.32\linewidth]{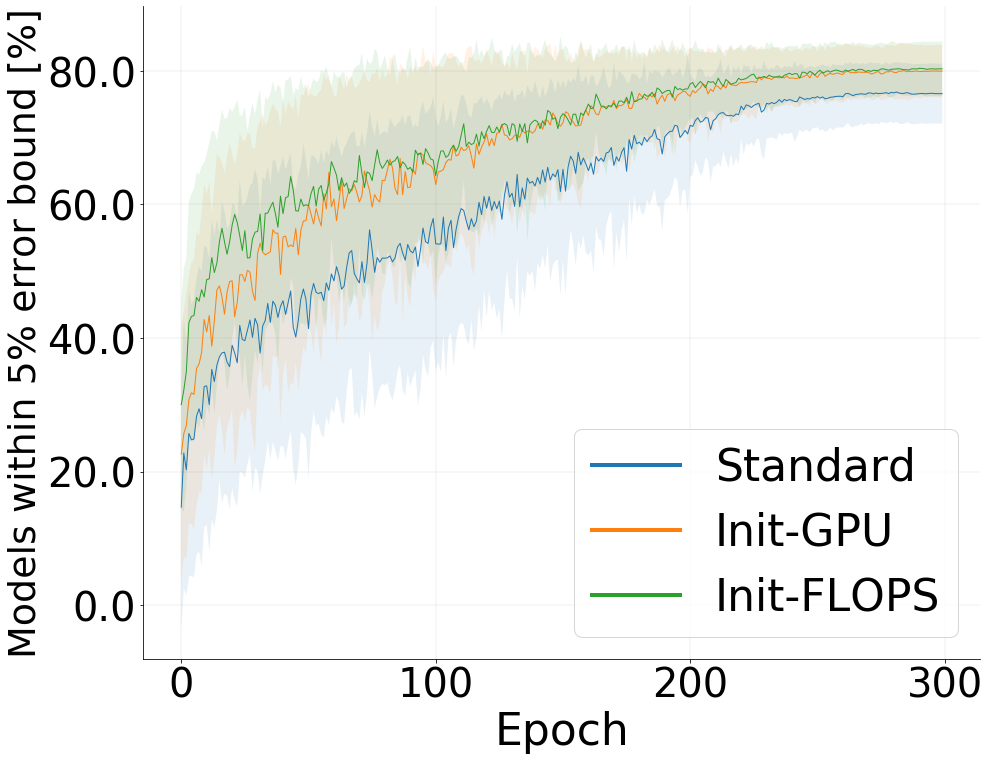}
    \includegraphics[width=0.32\linewidth]{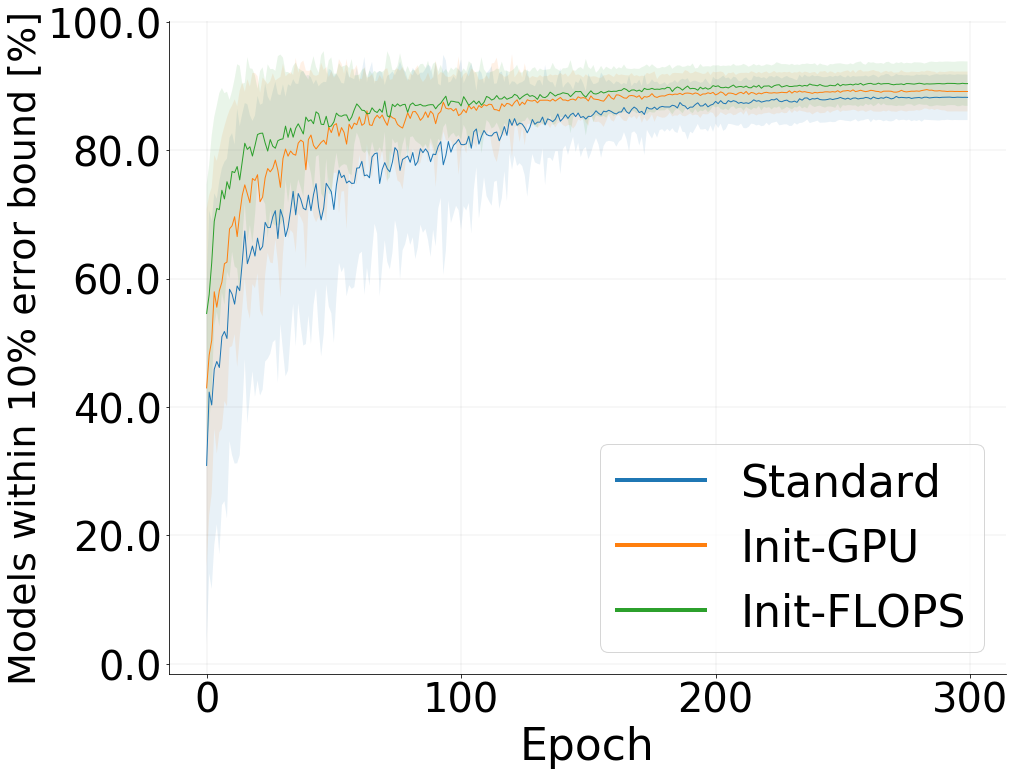}
    \caption{Training curves of the accuracy predictors without transfer learning (standard), with transfer learning from desktop GPU latency predictor (Init-GPU), and with transfer learning from FLOPS predictor (Init-FLOPS).}
    \label{fig:accuracy_training}
\end{figure}

Latency predictors can improve the performance of accuracy predictor as shown in Table~\ref{tab:accuracy_prediction} of Section~\ref{sec:latency}. The trained GCN captures the correlated features in the model which is useful to guide the training of a different GCN. Figure~\ref{fig:accuracy_training} further shows that the training process is improved. Y-axis is the percentage of models with predicted validation accuracy within the error bound relative to the actual validation accuracy. When initialized with the weight of latency/FLOPS predictor, the training process of accuracy predictor converges faster to better results.

\begin{figure}[!ht]
    \centering
    {\includegraphics[width=0.47\linewidth]{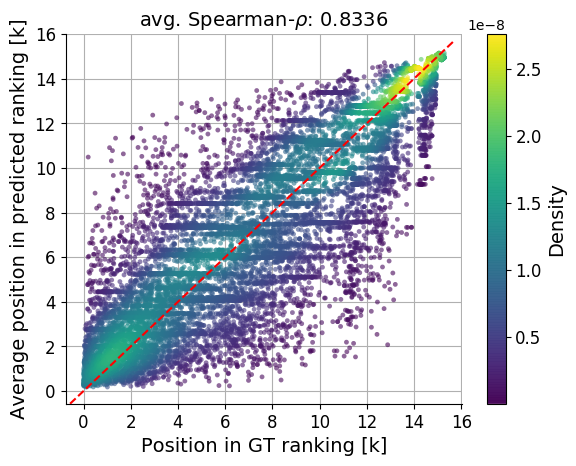}} 
    {\includegraphics[width=0.47\linewidth]{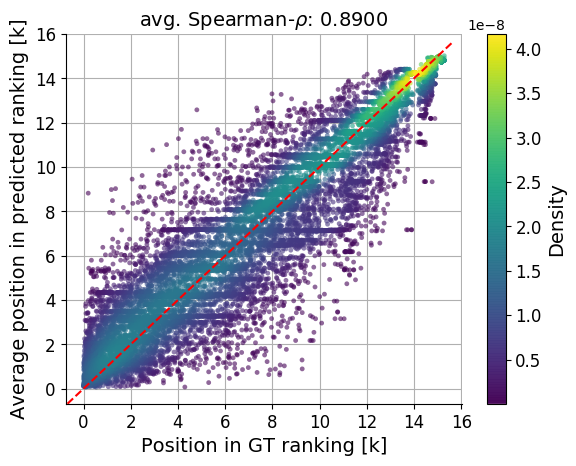}}
    \caption{
    (Left) Rankings produced by a standard predictor-based search. (Right) ranking produced by a predictor transferred from a latency predictor, x-axis is the position of a model according to the ground-truth ranking using validation accuracy, y-axis represents the average position of model in a ranking produced by a relevant method and the dashed red line marks the $x=y$ diagonal (i.e., perfect ranking).}
    \label{fig:acc-overview}
\end{figure}

In order to understand the underlying behavior and to improve the accuracy predictor proposed in Section~\ref{sec:acc-lat}, we plot the rankings produced by a standard predictor-based search (Figure~\ref{fig:acc-overview} left) and by a predictor transferred from latency predictor against the ground-truth ranking (Figure~\ref{fig:acc-overview} right).
Even though the accuracy predictor with transferred knowledge performs better in predicting the accuracy values of the models overall, the gain in accuracy ranking (which is important to NAS performance) is not as much.
This motivates \ours{} described in Section~\ref{sec:accuracy}.

\begin{algorithm}
\begin{small}
\LinesNumbered
\DontPrintSemicolon
\SetNoFillComment
\SetAlgoLined
\KwIn{\textit{(i)} Search space $\mathbb{S}$,
\textit{(ii)} budget for predictor training $K$ (number of models),
\textit{(iii)} number of iterations $I$,
\textit{(iv)} latency limit $L_{max}$ and latency predictor $P_L$,
\textit{(v)} trade-off factor $\alpha$ between $0$ and $1$,
\textit{(vi)} overall maximum number of models we can afford to train $M,M>K$}
\KwOut{The best found model $m^*$}
$m^* \leftarrow$ \textsc{none} \;
$\mathbb{C} \leftarrow \{\; s \; | \; s \in \mathbb{S} \land P_L(s) \le L_{max} \;\}$ \tcp*{candidates, remove models predicted to be too expensive} 
$\mathbb{T} \leftarrow \varnothing $ \tcp*{training set for the binary predictor}
$BP \leftarrow \text{initialize binary predictor with weights from } P_L$  \tcp*{binary predictor, optional transfer from $P_L$}
\For{$i \leftarrow 1$ \KwTo I}{
    \tcc{Update training set for the predictor\newline In each iteration we add $K/I$ models in total}
    $\mathbb{M} \leftarrow \{$ from $\mathbb{C}$, select the top $\alpha * K/I$ models and randomly select $(1-\alpha) * K/I$ models from the top $|\mathbb{C}|/2^i$ which are not already in $\mathbb{T} \;\}$ \tcp*{this is completely random in i=1}
    \ForEach(\tcp*[f]{models higher in $\mathbb{C}$ first}){$m \in \mathbb{M}$}{
        $a \leftarrow \text{train\_and\_validate}(m)$ \tcp*{Get accuracy $a$ of model $m$}
        $\mathbb{T} \leftarrow \mathbb{T} \cup \{(m,a)\}$ \tcp*{Add the model-accuracy pair to the training set}
        \tcc{keep track of the trained models \newline throughout the entire procedure}
        \If(\tcp*[f]{check if $m$ is truly below the lat. limit}){$\text{latency}(m) \le L_{max}$}{
            update $m^*$ if $m$ happens to be better
        }
    }
    \tcc{Update the predictor}
    \ForEach(\tcp*[f]{possibly shuffle and batch}){$\big((m_1,a_1),(m_2,a_2)\big) \in \mathbb{T}^2 \text{ s.t. } m_1 \ne m_2$}{
        $l \leftarrow \text{softmax}(BP(m_1, m_2))$ \;
        $t \leftarrow \text{softmax}([a_1, a_2])$ \;
        optimize $BP$ to minimize KL-divergence between $t$ and $l$
    }
    \tcc{Use the updated predictor to reevaluate the models in $\mathbb{C}$}
    $\mathbb{C} \leftarrow $ sort $\mathbb{C}$ using $BP$ to compare models
}
\tcc{At this point we stop training the predictor. \newline We have the final ordering of candidate models in $\mathbb{C}$ \newline and we can still train $M-K$ models to potentially find better $m^*$}
$\mathbb{M} \leftarrow \{$ top $M-K$ models from $\mathbb{C}$ which have not been trained yet $\}$ \;
\ForEach(\tcp*[f]{models higher in $\mathbb{C}$ first}){$m \in \mathbb{M}$}{
    \If(\tcp*[f]{check if $m$ is truly below the lat. limit}){$\text{latency}(m) \le L_{max}$}{
        $a \leftarrow \text{train\_and\_validate}(m)$ \;
        update $m^*$ if $m$ happens to be better
    }
}
\caption{The proposed search method based on pairwise relation learning.}
\label{alg:iter}
\end{small}
\vspace*{-1mm}
\end{algorithm}

\subsection{Motivation behind binary relation predictors}
We propose the binary relation predictor in Section 4 based on the following observations and intuition. 

\textbf{Observation 1: 
Ranking candidate models correctly according to their accuracy is more important than improving the absolute average accuracy of the accuracy predictor.
}

When the number of training samples is to be minimized, like in NAS -- the prediction quality of a GCN accuracy predictor can be improved by considering cheaper but roughly-correlated metrics, such as latency or FLOPS.
However, even when using those cheaper metrics, the achievable prediction accuracy degrades significantly as the number of training samples becomes heavily limited as shown in Table~\ref{tab:accuracy_prediction} of Section~\ref{sec:latency}. In order to maintain decent NAS quality even in those extreme cases, we propose to make more fundamental (compared to naively increasing a predictor's accuracy) changes in the predictor-based NAS by relaxing some of the current assumptions behind it.

More formally, we consider a predictor which gives each model a \textit{score} rather than predicts the absolute accuracy a model. Let $P$ be the predicted order of models obtained from the estimated scores, i.e., $P_n$ is the model in the search space which achieves the $n$-th largest score as estimated by a predictor.
Consequently, let $GT$ be the ground-truth order, i.e., $GT_n$ is the model which achieves the $n$-th highest validation accuracy.
Furthermore, let $P(GT_n)$ be the position in $P$ of the $GT_n$ model, i.e., $P(GT_n) = m \iff P_m = GT_n$. Similarly, $GT(P_n)$ is the analogical reverse.
It is easy to see that the performance of our predictor-based NAS should be maximized when $P=GT$, regardless of the values predicted by the scoring function.
Although a perfect accuracy predictor (in the absolute sense) would produce the perfect ordering of models, we argue that learning the perfect accuracy function is more challenging than learning a function which is only supposed to produce faithful ordering of models.

\textbf{Observation 2: Learning a binary relation rather than predicting absolute models.
} 

Taking a step further, the predicted order $P$ even need not be produced by a scoring function. Instead, we lean on the fact that any antisymmetric, transitive and connex \textit{binary relation} produces a linear ordering of its domain.
Thus, NAS could be solved by learning a binary relation rather than predicting absolute accuracy values.
This is a very important observation to maximize sample efficiency, since the reformulated binary relation changes the number of training samples for the predictor in a function of trained models to $O(n^2)$, rather than $O(n)$ in the standard approach.
This provides the predictor with more opportunities to learn efficiently when $n$ is limited.

We quantify the quality of different rankings $P$ produced by the proposed binary relation predictor together with different variations of the standard predictor.
All predictors were trained for 200 times, each time using a randomly sampled set of 100 models.
Then they are used to sort all 15k models in the NAS-Bench-201 dataset to produce a ranking.
We compared the predicted rankings by considering their correlation to the GT rankings in Figure~\ref{fig:ablation} (middle) of Section~\ref{sec:accuracy}.
The average Spearman-$\rho$ correlation coefficient between the position in prediction ranking and that in GT ranking shows that the proposed binary relation predictor achieves the best ranking correlation out of all our experiments.
However, despite producing the best results globally, the binary predictor did not yield the best NAS results.
This had led us to the next observation.

\textbf{Observation 3. Top-K rankings are important.} 

Even though $P=GT$ maximizes the performance of predictor-based NAS, achieving perfect correlation between the two rankings is very challenging in practice considering a limited number of training samples.
Although errors are expected to occur \textit{somewhere} in the predicted ranking, in the context of NAS it is especially important to make sure that those errors are minimized in the top of the rankings, otherwise even a very well correlated ranking might fall short to a less optimal alternative.

When closely examining the results obtained by running the binary relation predictor, we saw that even though the global correlation was very good, the best performing models happened to be burdened with a relatively higher error than the rest of the search space, as indicated by the red circle in Figure~\ref{fig:ablation}.
In the context of NAS, any ranking $P'$ which satisfies $P'_1 = GT_1$ is as good as the perfectly correlated ranking $P=GT$, and analogically, any ranking $P''$ for which $GT(P''_1)$ is very high is less likely to yield good results in practice, regardless of their global landscapes. 

Section~\ref{sec:accuracy} has introduced \ours{} - a NAS method based on binary relation predictor combined with an iterative data selection strategy. Algorithm~\ref{alg:iter} describes the steps to search for the best model based on pairwise relation learning and to find better models by focusing on high performing models.
Figure~\ref{fig:ablation_supp} shows that the proposed method achieves the best NAS performance.
The iterative training approach helps with the top model performance even though achieves worse results globally.

\subsection{BRP-NAS details} 

We train each predictor using the hyperparameters listed in Table~\ref{tab:accuracy_predictor_hyperparameters}.
A technical detail related to accuracy predictors is that when training them we used fixed accuracy values for each model by considering only the seed 888 from NAS-Bench-201, and by taking an average accuracy from NAS-Bench-101 (for DARTS we were taking accuracy after a single training, whatever that was).
Even though the predictors were trained using noise-free data, when performing a search and reporting results we were returning a random accuracy for each model, thus resulting in a potential distribution shift between training and testing sets for our predictors, induced by the stochastic nature of the training process.
More details about the impact of SGD, and related experiments, can be found in Section~\ref{sec:app:sgd}.

\begin{table}[!ht]
    \caption{Training hyperparameters of the accuracy predictors.}
    \centering
    \begin{tabular}{l|llll}
    \toprule
         \multirow{3}{*}{batch size} & & K=100 & K=50 & K=25 \\
                                     \cmidrule(lr){2-5}
                                     & normal & 50 & 32 & 16 \\
                                     & binary & 64 & 32 & 32 \\
                                     \cmidrule(lr){2-5}
         Learning rate schedule & \multicolumn{4}{l}{cosine annealing} \\
         Initial learning rate & \multicolumn{4}{l}{0.00035} \\
         Optimizer & \multicolumn{4}{l}{AdamW} \\
         L2 weight decay & \multicolumn{4}{l}{0.0005} \\
         Dropout ratio & \multicolumn{4}{l}{0.2} \\
         Training epochs & \multicolumn{4}{l}{250 \footnotesize{(early stopping patience of 35 epochs)}} \\
    \bottomrule
    \end{tabular}
    \label{tab:accuracy_predictor_hyperparameters}
\end{table}

For all NAS experiments, the training set for predictors was discovered online (i.e., no prior knowledge was assumed), therefore, we did not have access to a separate validation set -- whenever a validation set would have been used, we used training set instead.

To perform NAS with our predictor, in the first phase, we use a randomly selected small set of $K$ models from the search space to train the predictor (with or without iterative data selection).
Then in the second phase, we use the predictor to score all the models in the search space in order to find potentially the most efficient one.
The predictor could be trained in the second phase if required, but we did not do that in order to check the achievable performance when $K$ is upper-bounded.

When training the predictor iteratively, we always use 5 iterations.
It means that different number of models ($K/5$) were trained and added to the predictor's training set in each iteration.

\subsubsection{Ablation studies}

\begin{figure}[!ht]
    \centering
    \includegraphics[width=0.32\textwidth]{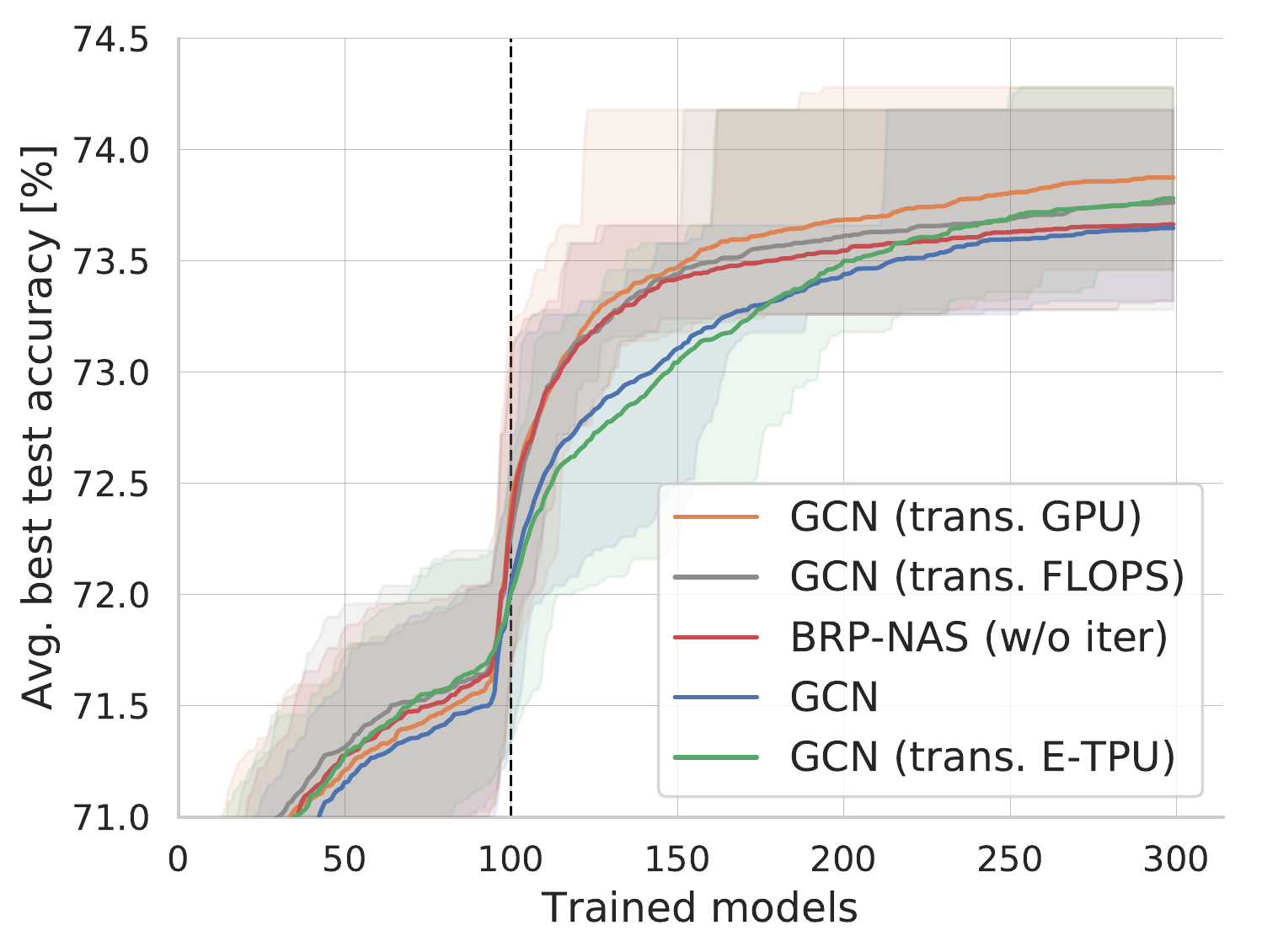}
    \includegraphics[width=0.32\linewidth]{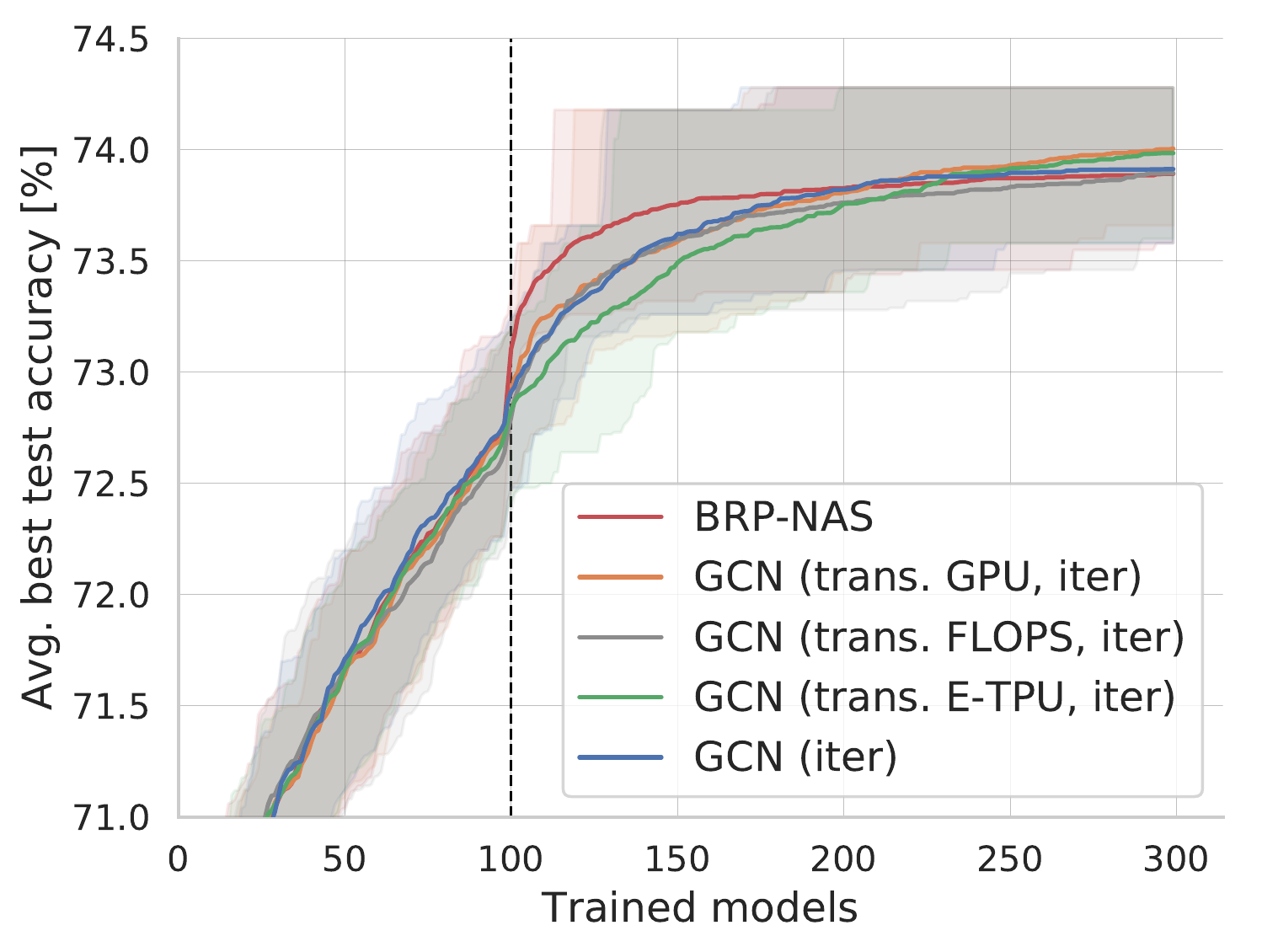}
    \includegraphics[width=0.32\linewidth]{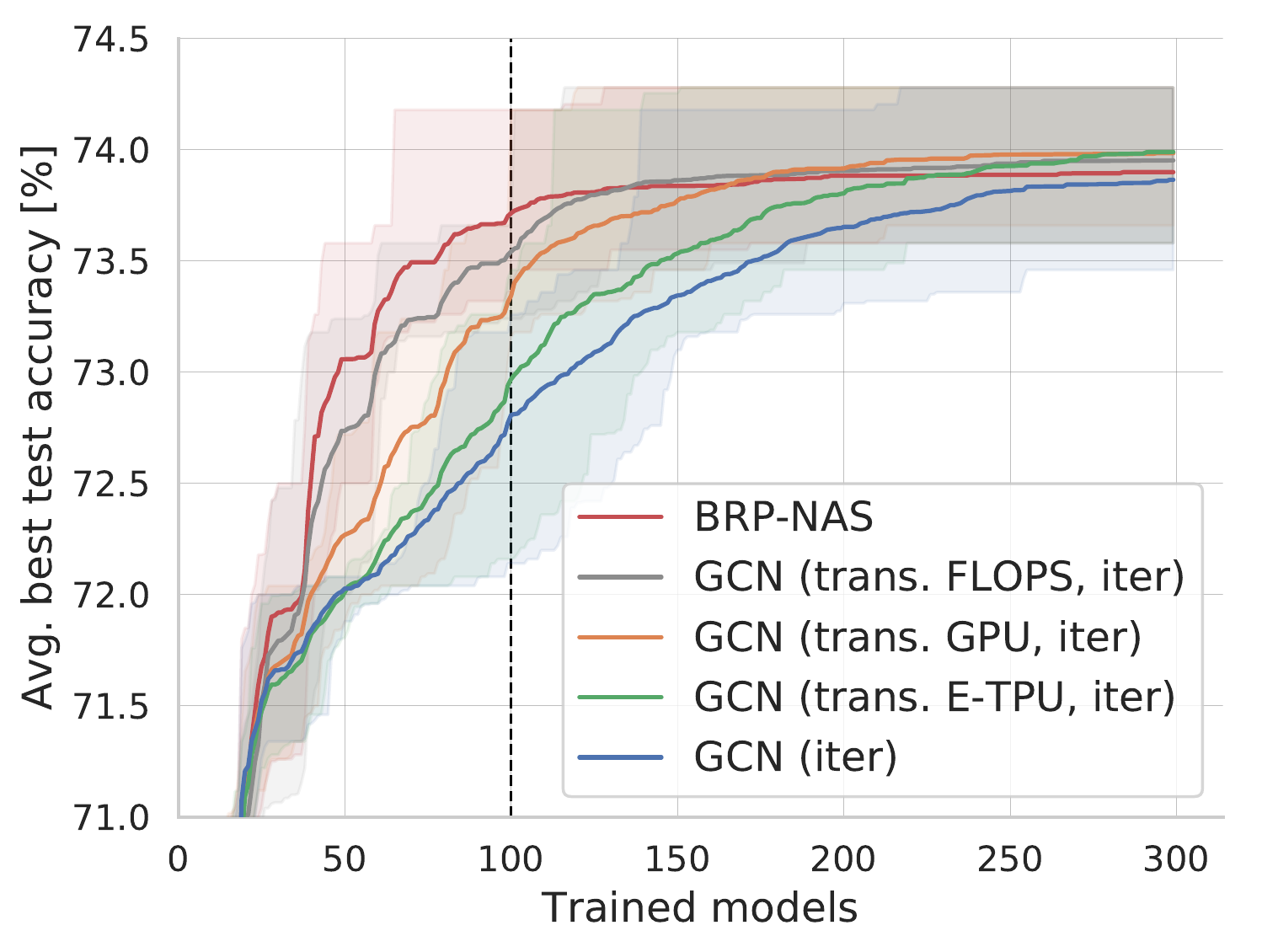}
    \caption{(Left) Comparison of NAS performance with the standard GCN predictor, GCN predictor with transfer learning (from desktop GPU, Embedded TPU and FLOPS predictors) and \ours, all trained non-iteratively.
    (Middle, Right) Comparison with predictors trained via iterative approach with $\alpha=0$ (middle) and $\alpha=0.5$ (right). }
    \label{fig:ablation_supp}
\end{figure}

\begin{figure}[!ht]
    \centering
    \includegraphics[width=0.32\textwidth]{appendix/images/A2-accuracy/20-brp-ablation/test_acc_ablation_iter_best_curves.pdf}
    \includegraphics[width=0.32\linewidth]{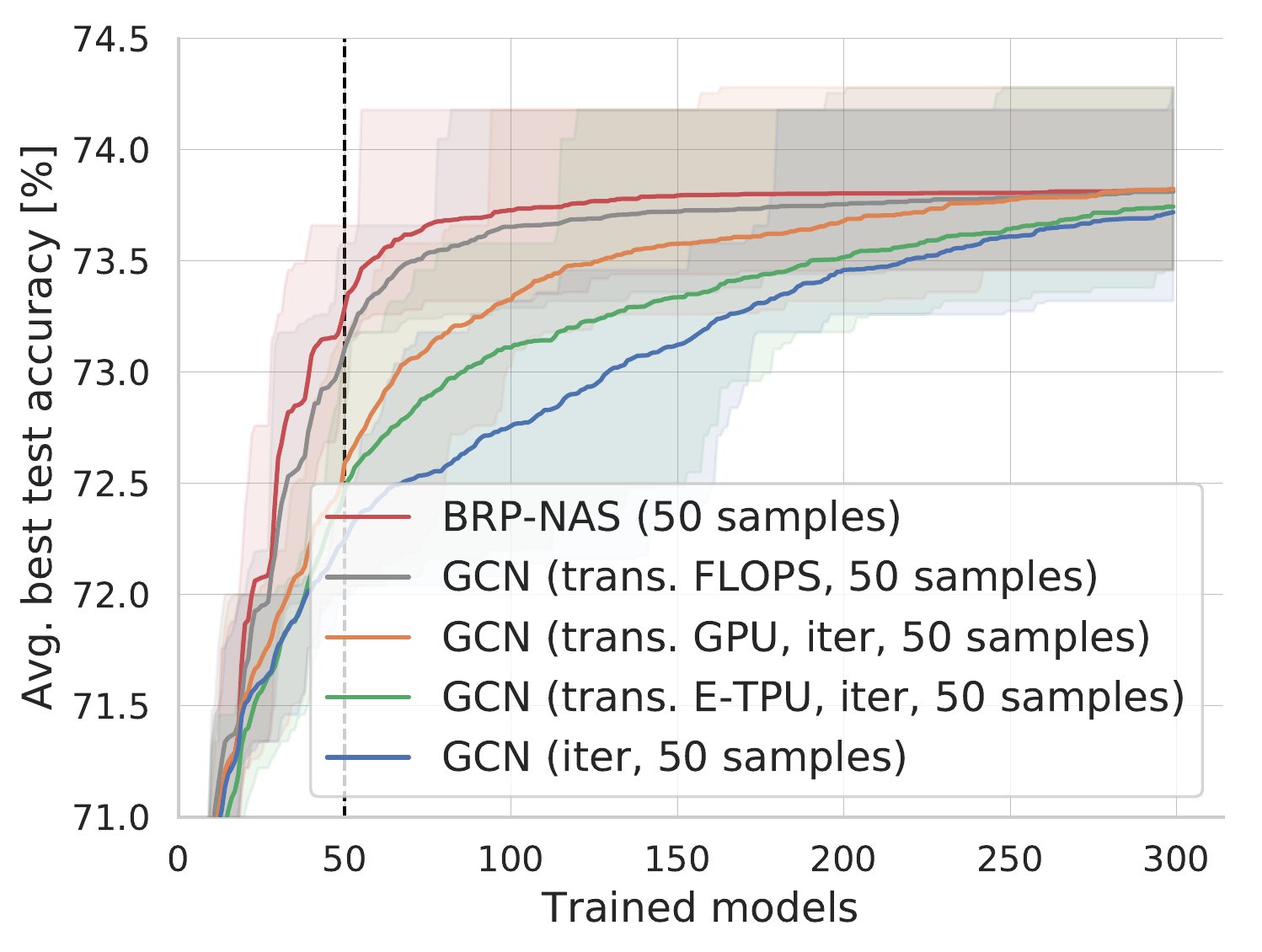}
    \includegraphics[width=0.32\linewidth]{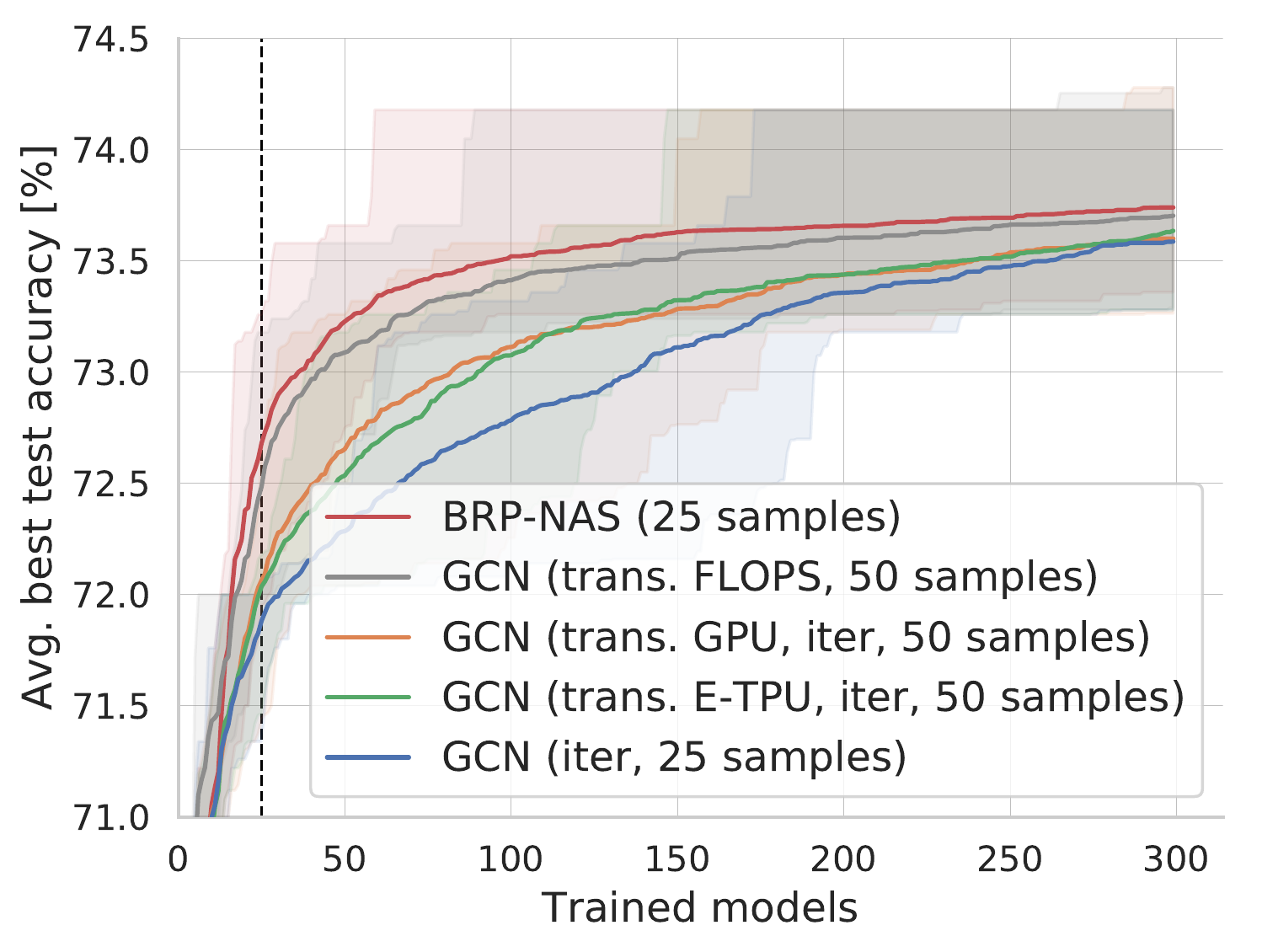}
    \caption{Comparison of GCN accuracy predictors and our BRP-NAS predictor (with $\alpha=0.5$) under different total number of models used to train the predictor: 100, 50 and 25, respectively.
    In all cases, 5 iterations were used.}
    \label{fig:ablation_supp_samples}
\end{figure}

Figure~\ref{fig:ablation_supp} summarizes the results of NAS using various GCN-based predictors, and Figure~\ref{fig:ablation_supp_samples} additionally shows results with different total number of models used to train the predictors.
\ours{}, which utilizes binary relation predictor trained via iterative data selection, has the best performance comparing to other approaches, and is able to achieve competitive results even when very few models are used to train it.

\begin{figure}[!ht]
    \centering
    \begin{minipage}{.45\textwidth}
        \centering
        \includegraphics[width=\linewidth]{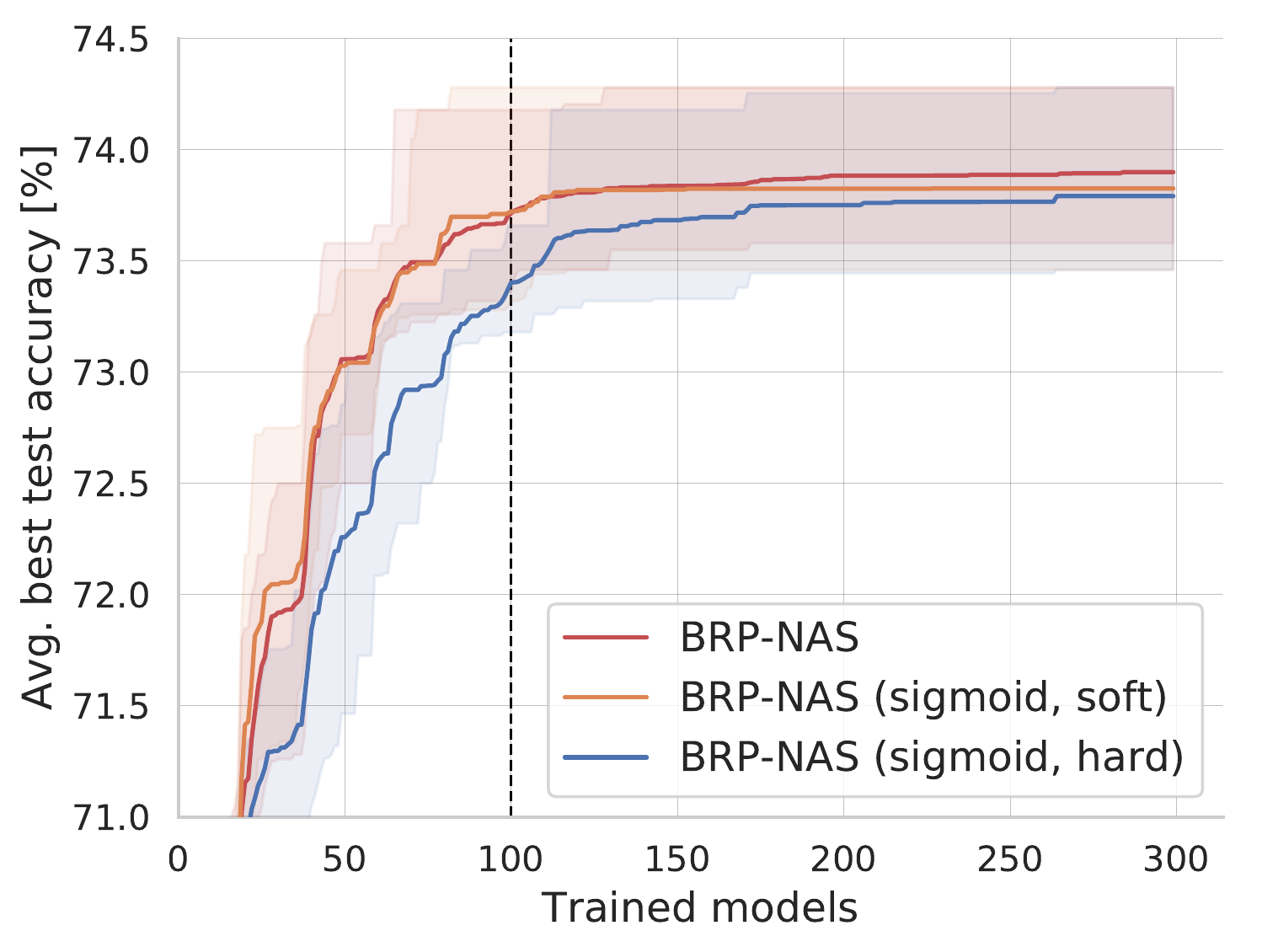}
        \captionof{figure}{Effects of using a different activation functions with our binary accuracy predictor.}
        \label{fig:ablation_sigmoid}
    \end{minipage}%
    \hspace{.05\textwidth}
    \begin{minipage}{.45\textwidth}
        \centering
        \includegraphics[width=\linewidth]{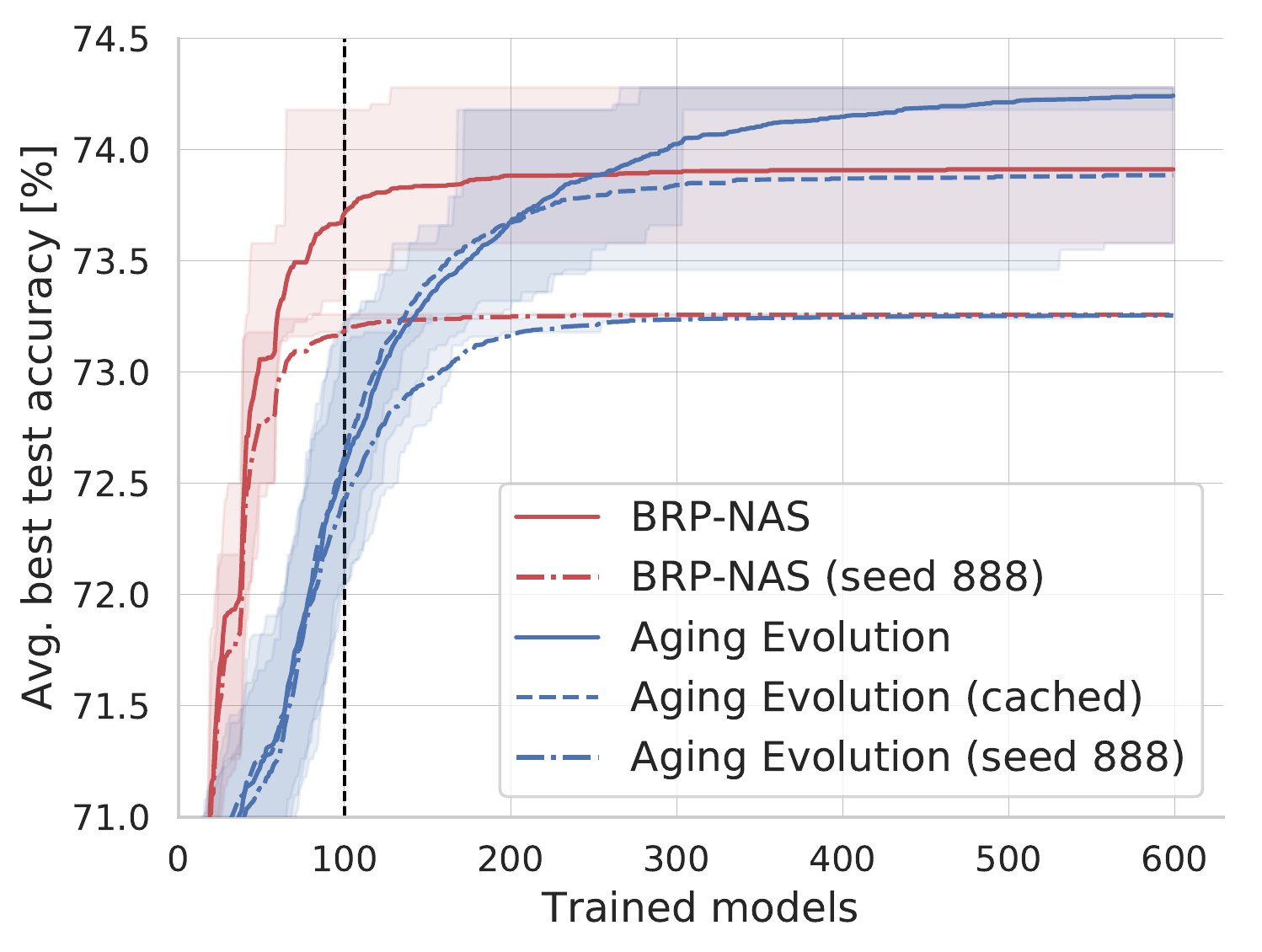}
        \captionof{figure}{Effects of the SGD-induced randomness on NAS performance. Runs labeled with "(seed 888)" were using only one seed from NAS-Bench-201. Aging Evolution (cached) was not allowed to train the same model multiple times.}
        \label{fig:ablation_sgd}
    \end{minipage}
\end{figure}

Figure~\ref{fig:ablation_sigmoid} additionally shows the effects of using different activation functions and related labels.
Specifically, we tried replacing our proposed 2-way softmax with analogical 1-way sigmoid activation (which some people can consider more canonical when dealing with probability of opposite events) with either soft or hard labels.
Soft labels were obtained by by linearly interpolating between accuracies of the related networks ($a,b$): $l_{a,b}=\frac{\text{acc}(a)-\text{acc}(b)+1}{2}$, whereas hard labels were simply $1$ if model $a$ is more accurate or $0$ otherwise.
In both cases the predictor was trained to minimize binary cross-entropy loss.
As can be seen, there is no visible difference between softmax and sigmoid activations when both use soft labels, however, switching to hard labels results in a reduced sample efficiency.
We decided to stick with the softmax activation 
as it can be easily extended to the $n$-ary case (although doing so is outside the scope of this paper) and thus we consider it more general.

%% file: appendix/A3-results.tex
\section{Supplementary Material for Section~\ref{sec:results}: End-to-end results}\label{sec:app:results}

\subsection{Details of the baseline NAS algorithms used in the paper}

\textbf{Aging Evolution. } We run aging evolution with pool size 64 and sample size 16, values which we found work well for NAS-Bench-201 models.

\textbf{REINFORCE. } Similar to \cite{zoph2018}, we use a single cell LSTM controller which is trained with REINFORCE (no PPO). 

\textbf{Random search. } We select models randomly by picking 6 numbers from the range of 1-5 uniformly.

\subsection{SGD-induced randomness in NAS-Bench-201}\label{sec:app:sgd}
As mentioned in the main paper, when comparing different algorithms on NAS-Bench-201 we observed that eventually AE surpasses our \ours{} due to its ability to train models multiple times.
To validate our reasoning and at the same time formally measure the effect of SGD randomness on NAS, we conducted a set of experiments in addition to the ones presented in the main text.
Specifically, we checked: \textit{(a)} performance of both \ours{} and AE when SGD noise is not present (using only seed 888), and \textit{(b)} running a modified AE where we remember trained models and avoid mutations that would result in the same model being trained more than once (cached mode).
The results are summarized in Figure~\ref{fig:ablation_sgd}.
We can see that indeed AE surpasses our predictor due to its ability to train models multiple times and that the gap is proportional to the level of the SGD-induced noise.
At the same time we can see that the difference in results obtainable with and without considering the SGD noise is surprisingly high -- the performance of the best model increases, in terms of the top-1 accuracy, by almost 2\% absolute.
Interestingly, we did not observe a similar gap when running analogical experiments in a constrained setting (5ms, D-GPU), suggesting that the huge gap observed for the top performing models is not necessarily common, even within the same search space.

\subsection{Comparison to Prior Work on NAS-Bench-101} 

\begin{figure}[!ht]
    \centering
    \includegraphics[width=0.8\textwidth]{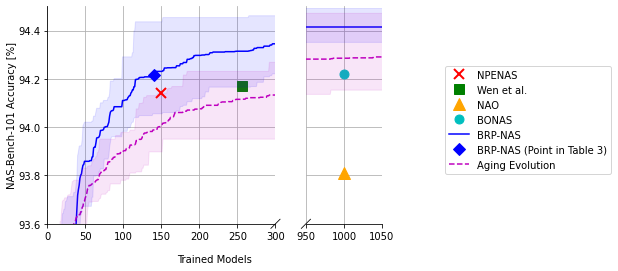}
    \caption{Comparison to prior work on NAS-Bench-101 dataset.}
    \label{fig:nas1_comp}
\end{figure}

We compare \ours{} with the previously published predictor-related NAS on NAS-Bench-101. 
We first train the predictor using the validation accuracy from 100 models, then we train the subsequent models which are picked from the top-ranked models returned by the predictor. 
As shown in Figure~\ref{fig:nas1_comp}, \ours{} finds a model with higher final test accuracy on CIFAR-10 dataset (94.22\%) using fewer steps (140 trained models) than the work under comparison.

\subsection{DARTS search space}

We run BRP-NAS on the DARTS search space, as described in Section~\ref{sec:results}.
All models were trained using the official DARTS implementation\footnote{https://github.com/quark0/darts} to ensure fair comparison.
When quantifying searching cost, we assumed one model takes 1 GPU day to train, which we empirically observed to be true on average when training our models using V100 GPUs.

\begin{figure}[!ht]
    \centering
    \includegraphics[width=0.5\textwidth]{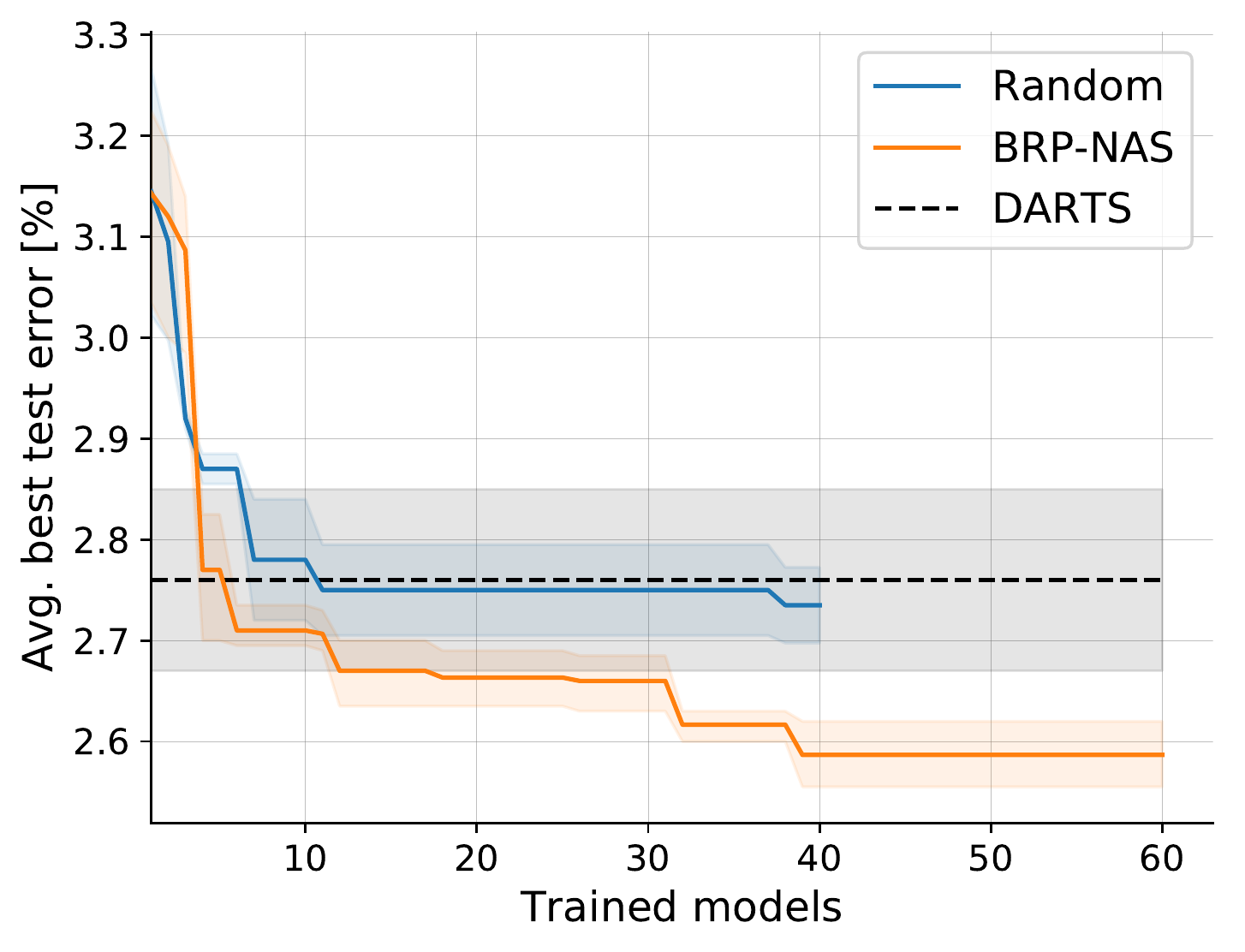}
    \caption{Performance on DARTS search space.}
    \label{fig:darts_comp}
\end{figure}

The results, additionally to Table~\ref{tab:darts} presented in the main paper, are shown in Figure~\ref{fig:darts_comp}.
Our first observation was that the search space is actually quite dense with good models and -- to our surprise -- our simple random search algorithm (take a random model and train) already was able to achieve very good results -- comparable with differentiable search for the same searching budget.
This is also shown visible in good performance of our BRP-NAS during the first iteration.
Because of that, we also include more detailed comparison against random search -- to validate if our approach can actually improve.
From our results, we can see that indeed as we move onto to the second and third iterations of our searching algorithm, the gap between random search and ours increases, suggesting that the predictor is able to extract useful features and use them to identify better models.
On the other hand, the gap between random search and ours in the first iteration (first 20 models, should be close to each other) suggests that the results can still be influenced by the random nature of the first few selected models -- unfortunately, due to limited computational resources we weren't able to run each algorithm for more than 3 times (each time training up to 60 models, as presented in the figure).
We suspect that the gap between random search and BRP-NAS should remain if we take average of more runs, but to test it robustly one would probably need to do more exhaustive evaluation.

%% file: appendix/A4-benchmark.tex
\section{Supplementary Material for Section~\ref{sec:benchmark}: Latency prediction benchmark}\label{sec:app:benchmark}

We run each model in NAS-Bench-201 on the follow devices and run-times.
(i) Desktop CPU - Intel Core i7-7820X - TensorFlow 1.15.0,
(ii) Desktop GPU - NVIDIA GTX 1080 Ti - TensorFlow 1.15.0,
(iii) Embedded GPU - NVIDIA Jetson Nano - TensorFlow 1.15.0,
(iv) Embedded TPU - Google EdgeTPU - TensorFlow Lite Runtime 2.1.0,
(v) Mobile GPU - Qualcomm Adreno 612 GPU - SNPE 1.36.0.746,
(vi) Mobile DSP - Qualcomm Hexagon 690 DSP - SNPE 1.36.0.746.

\begin{figure}[!ht]
    \centering
    \begin{tabularx}{\linewidth}{CCC}
        \begin{subfigure}[b]{.32\textwidth}
        \includegraphics[width=\linewidth]{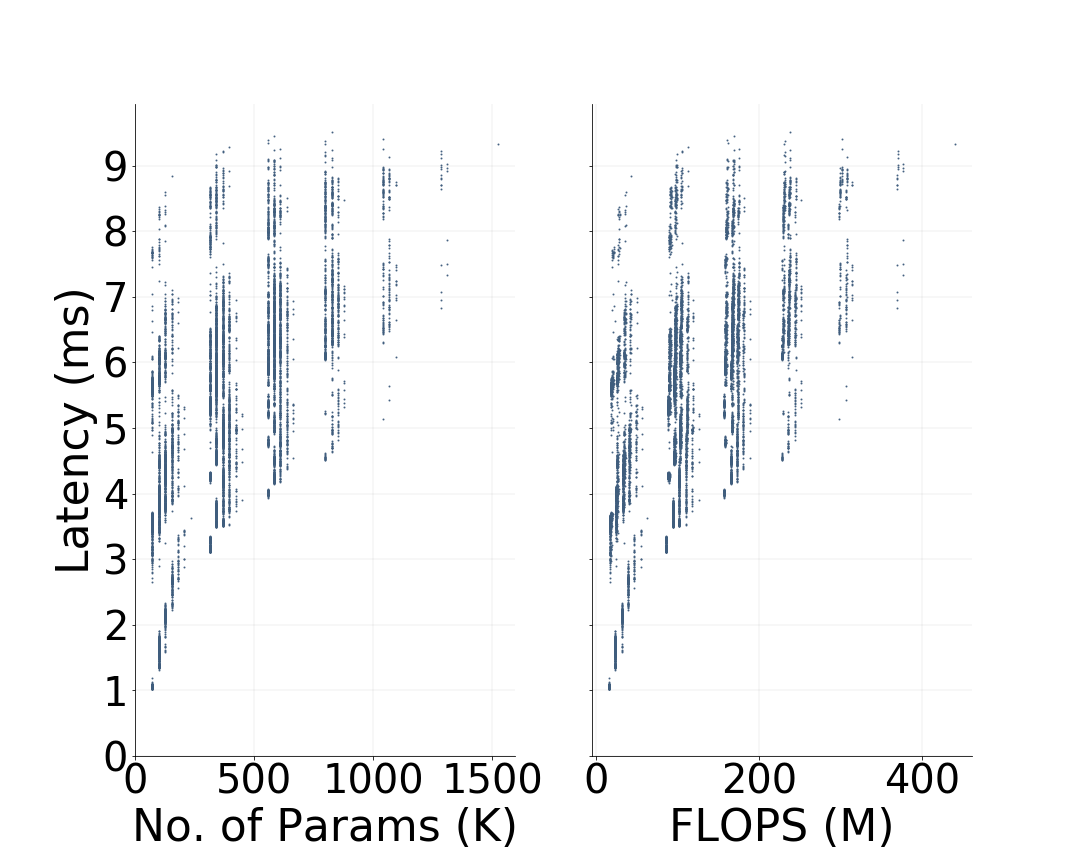}
        \caption{Desktop CPU}
        \end{subfigure}
        \begin{subfigure}[b]{.32\textwidth}
        \includegraphics[width=\linewidth]{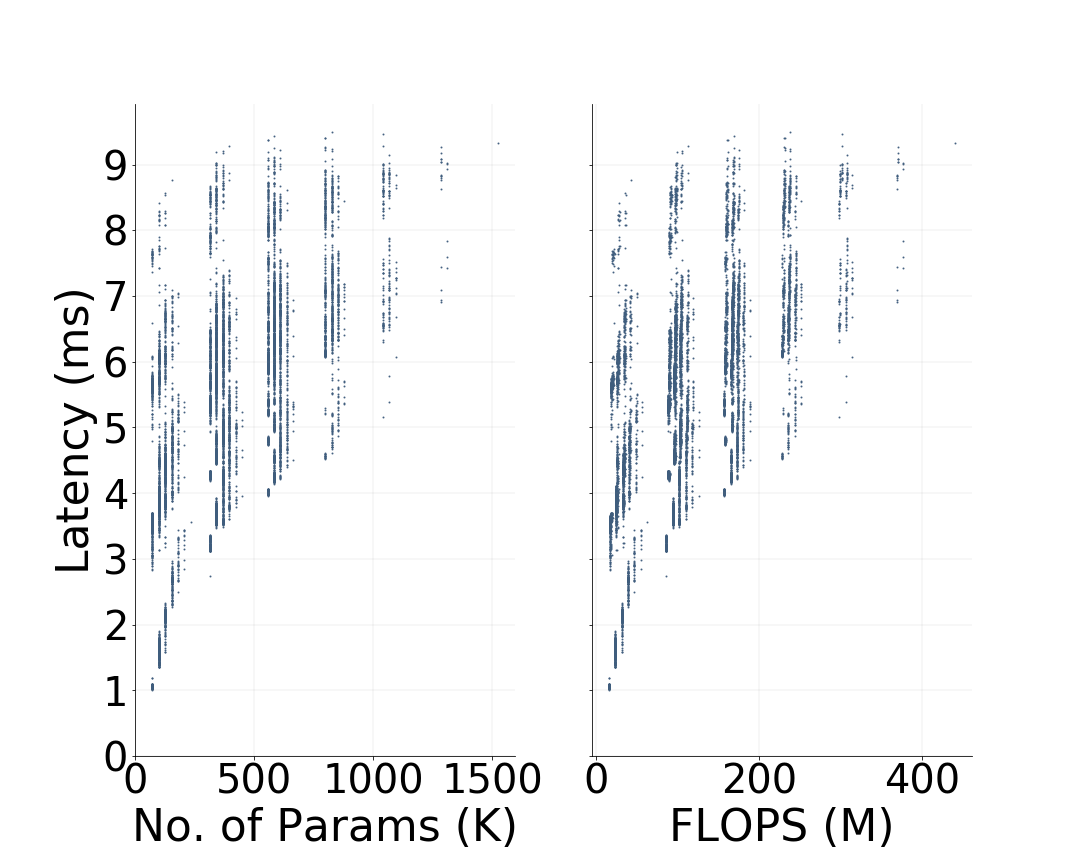}
        \caption{Desktop GPU}
        \end{subfigure}
        &
        \begin{subfigure}[b]{.32\textwidth}
        \includegraphics[width=\linewidth]{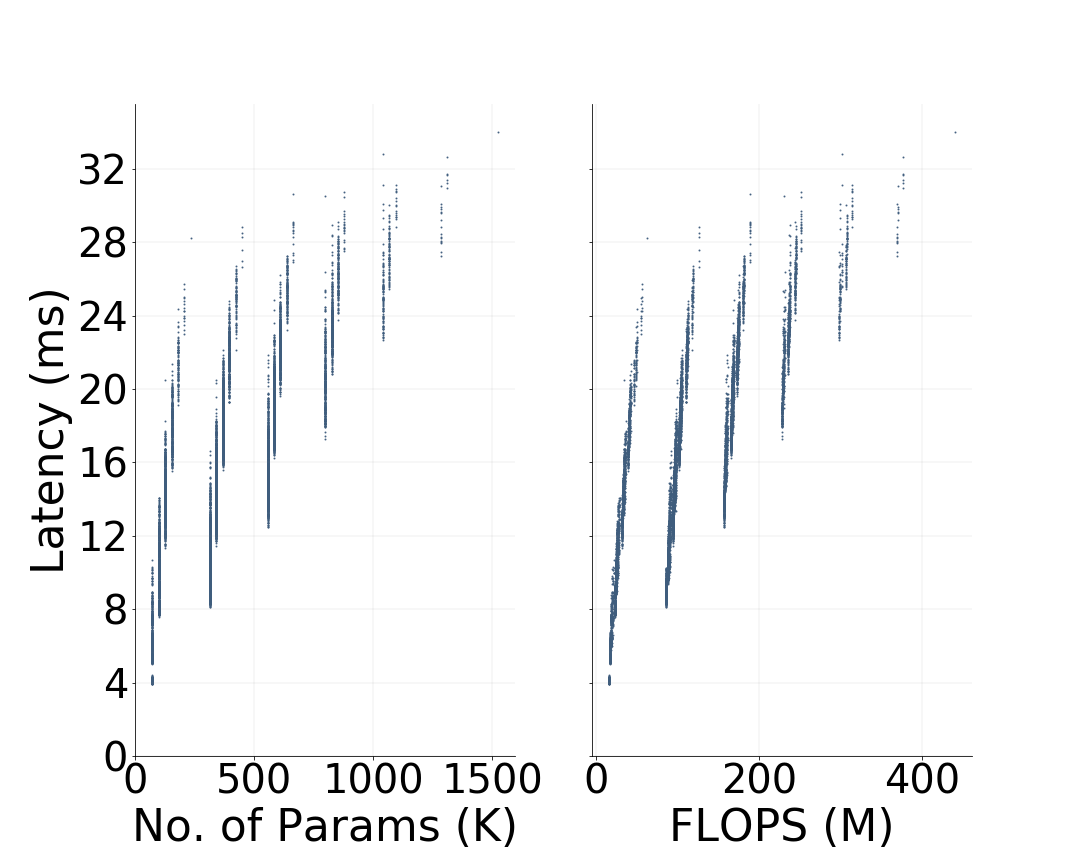}
        \caption{Embedded GPU}
        \end{subfigure}
        \begin{subfigure}[b]{.32\textwidth}
        \includegraphics[width=\linewidth]{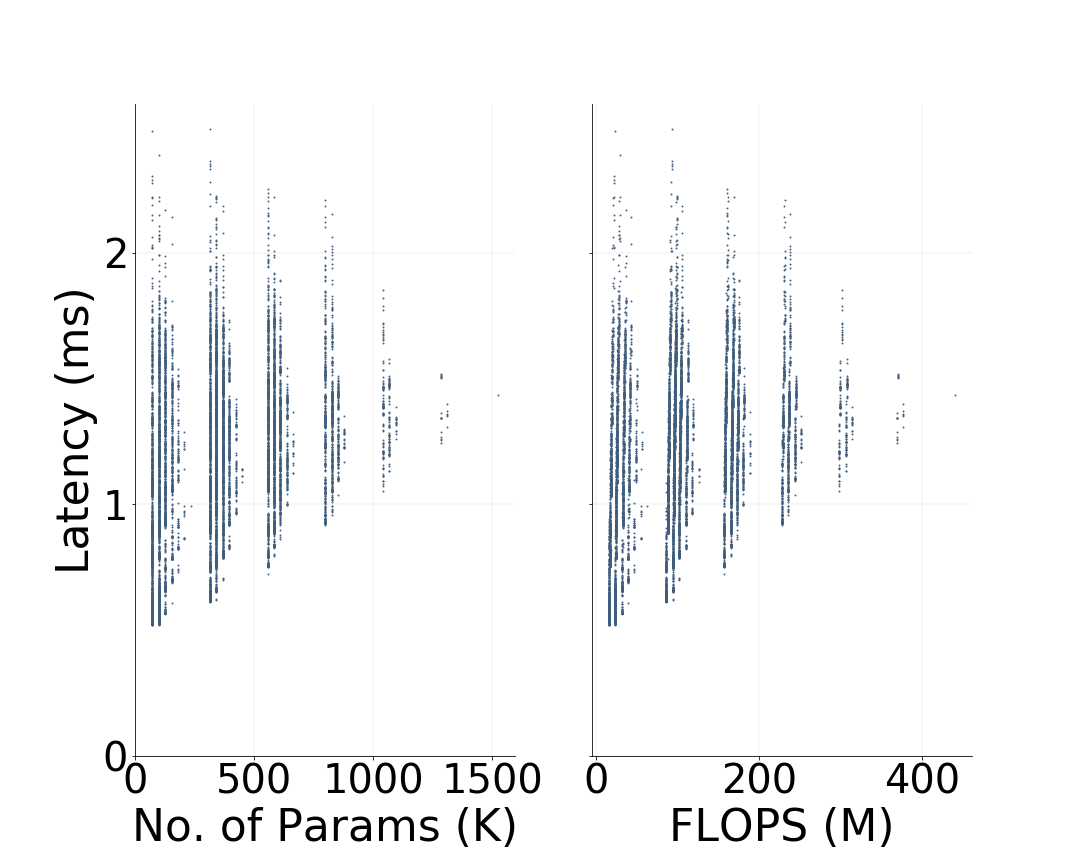}
        \caption{Embedded TPU}
        \end{subfigure}
        &
        \begin{subfigure}[b]{.32\textwidth}
        \includegraphics[width=\linewidth]{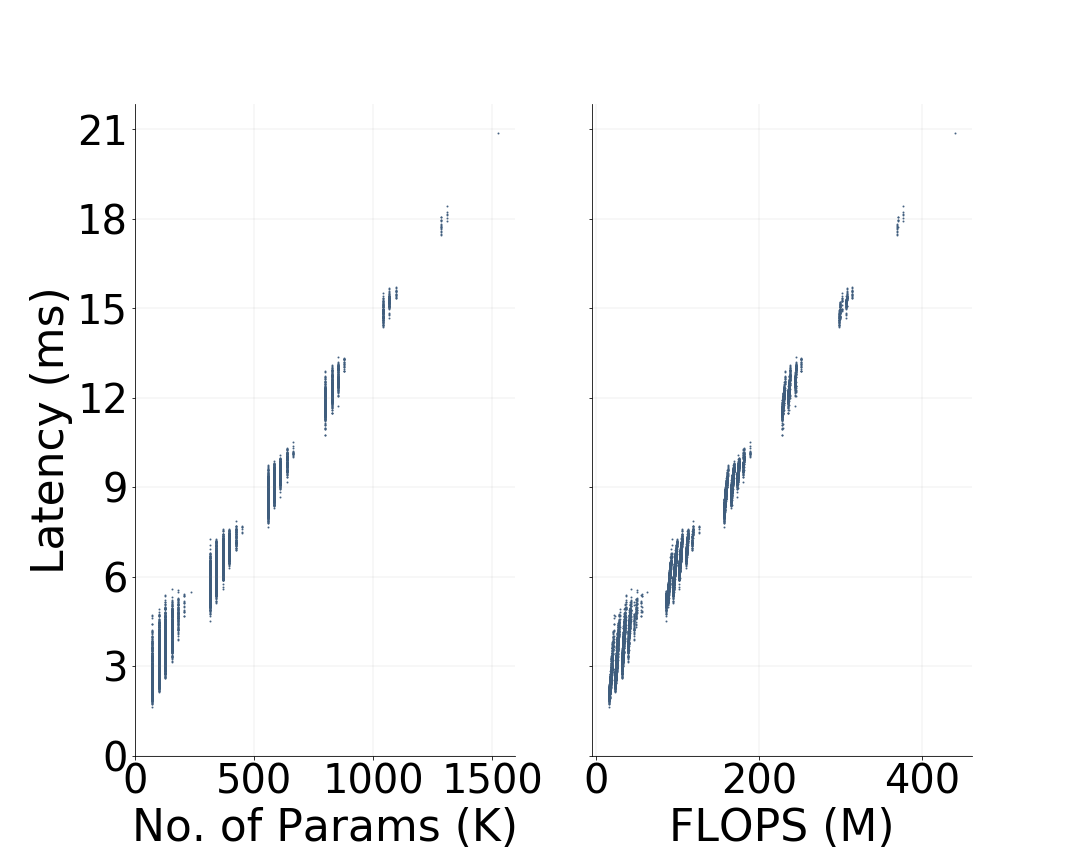}
        \caption{Mobile GPU}
        \end{subfigure}
        \begin{subfigure}[b]{.32\textwidth}
        \includegraphics[width=\linewidth]{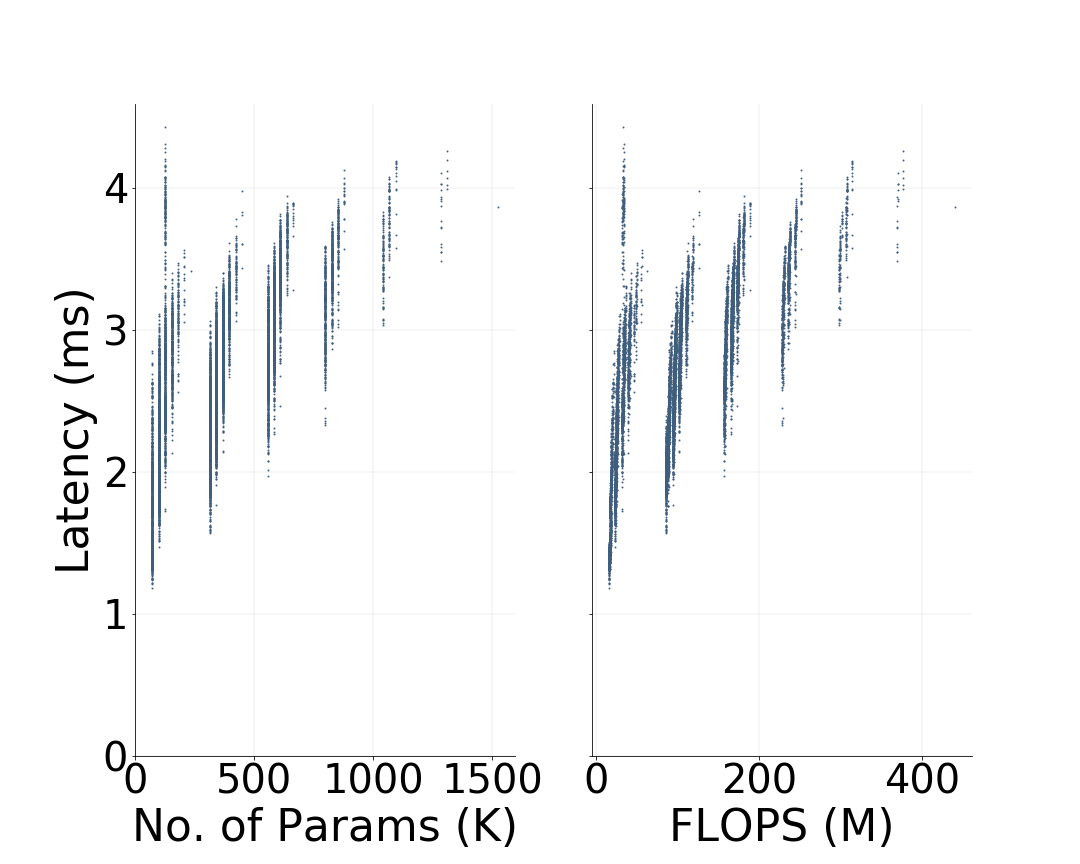}
        \caption{Mobile DSP}
        \end{subfigure}
    \end{tabularx}
    \caption{In most cases, FLOPS and the number of parameters are not a good approximation towards run-time latency on-device.}
    \label{fig:latency_vs_params}
\end{figure}

In Figure~\ref{fig:latency_vs_params}, we show the scatter plots of models taken from NAS-Bench-201 dataset that illustrates the connection between the latency of various devices and number of parameters/FLOPS. Each point in the plots represents the average latency of running a model on the stated device. We can see that latency is not strongly correlated with FLOPS or number of parameters. These metrics are unreliable proxies to predict latency.

\begin{figure}[!ht]
    \centering
    \begin{tabularx}{\linewidth}{CCC}
        \begin{subfigure}[b]{.32\textwidth}
        \includegraphics[width=\linewidth]{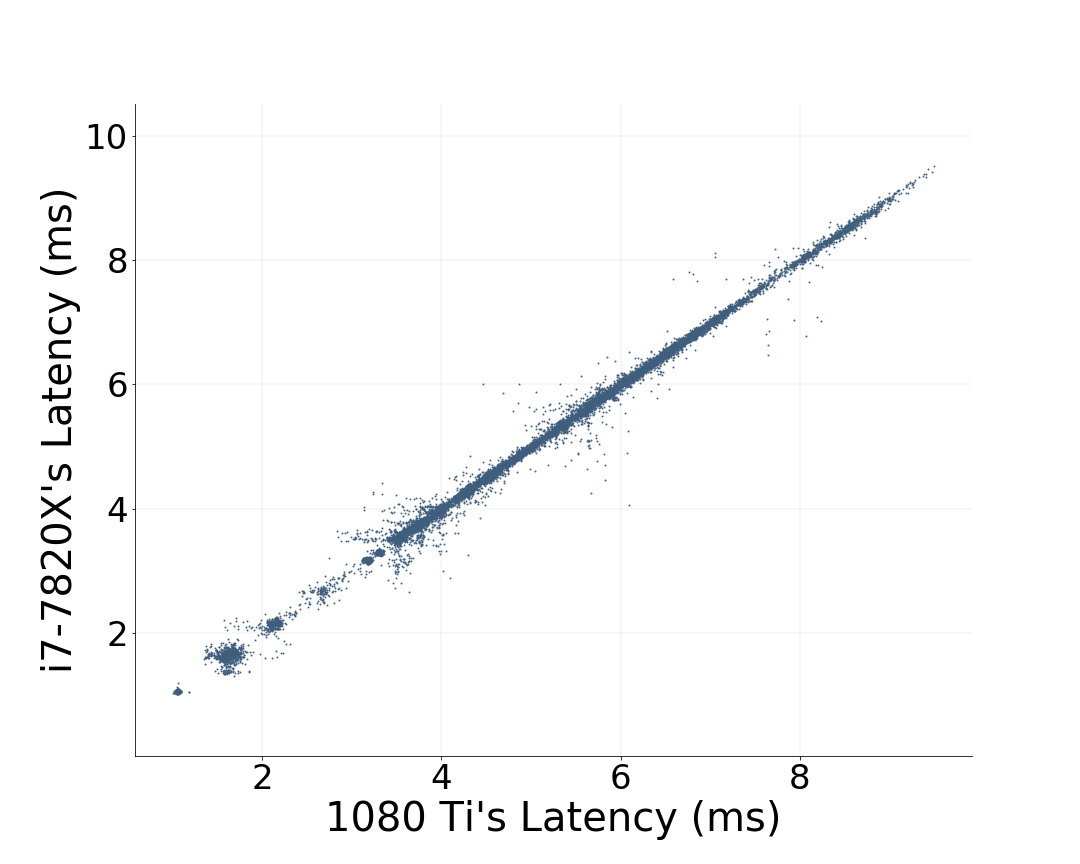}
        \caption{Desktop CPU vs Desktop GPU}
        \end{subfigure}
        \begin{subfigure}[b]{.32\textwidth}
        \includegraphics[width=\linewidth]{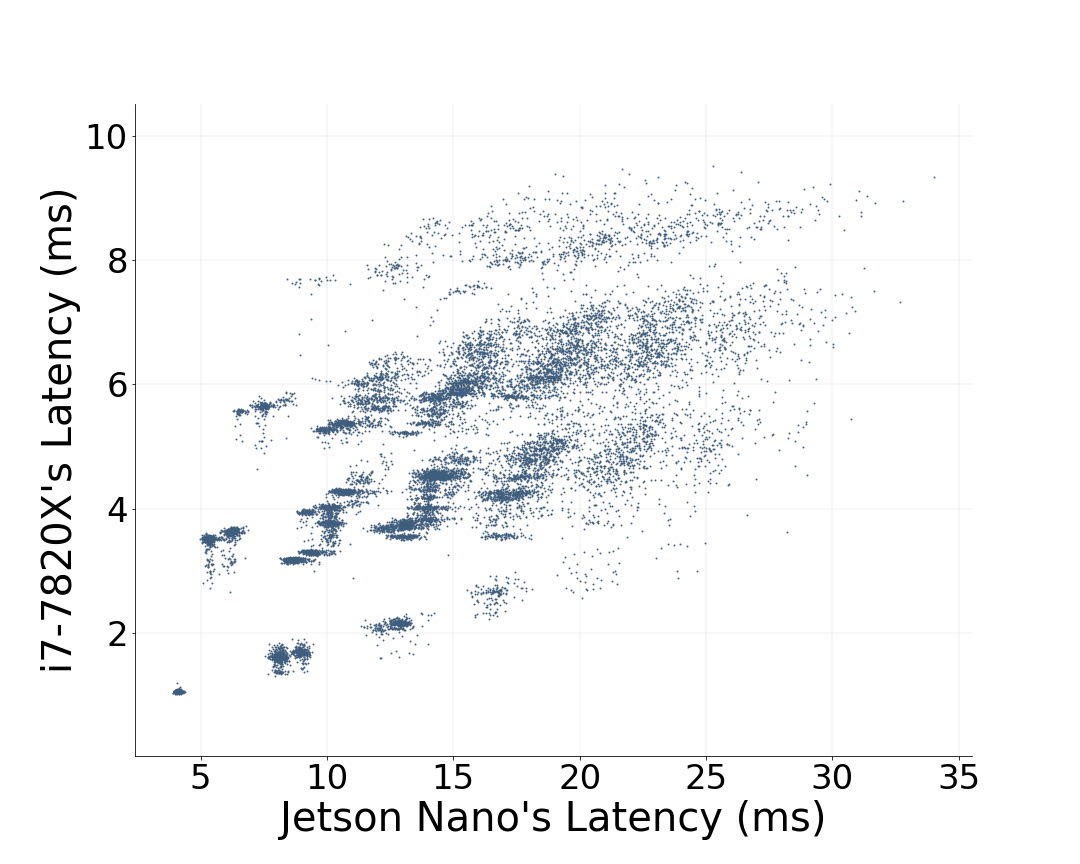}
        \caption{Desktop CPU vs Embed. GPU}
        \end{subfigure}
        \begin{subfigure}[b]{.32\textwidth}
        \includegraphics[width=\linewidth]{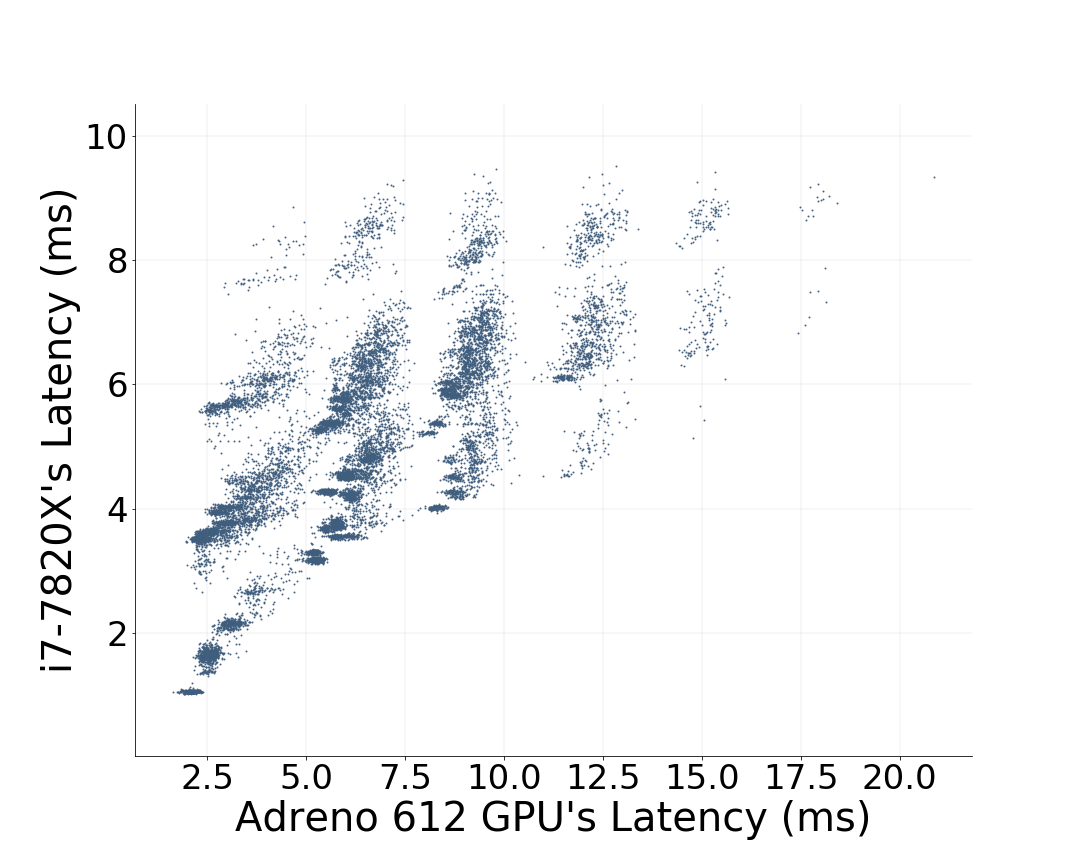}
        \caption{Desktop CPU vs Mobile GPU}
        \end{subfigure}
        \begin{subfigure}[b]{.32\textwidth}
        \includegraphics[width=\linewidth]{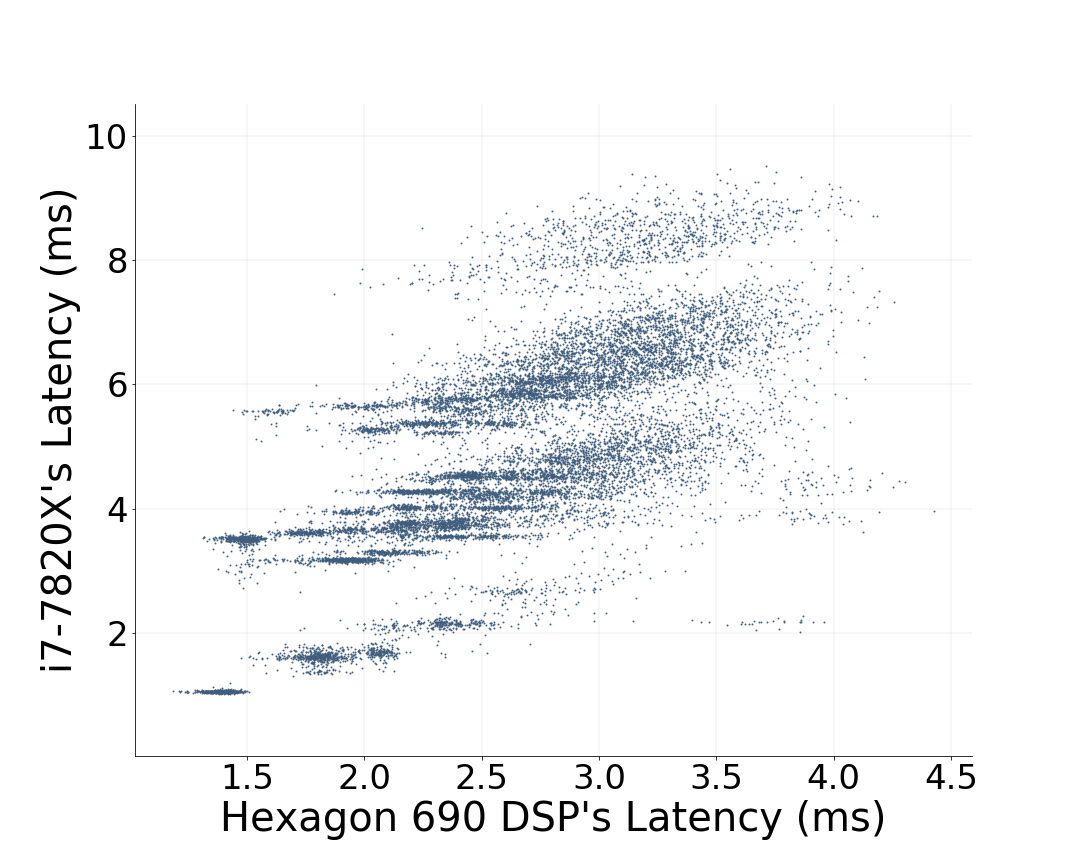}
        \caption{Desktop CPU vs Mobile DSP}
        \end{subfigure}
        \begin{subfigure}[b]{.32\textwidth}
        \includegraphics[width=\linewidth]{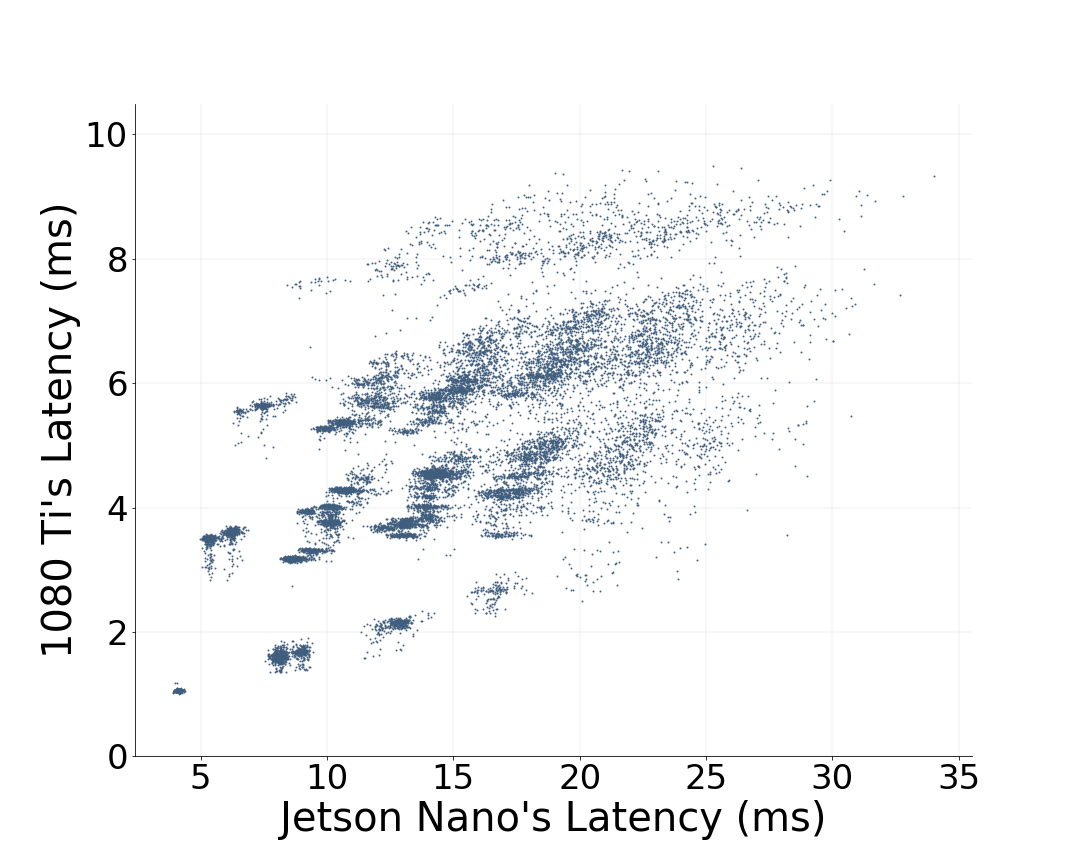}
        \caption{Desktop GPU vs Embed. GPU}
        \end{subfigure}
        &
        \begin{subfigure}[b]{.32\textwidth}
        \includegraphics[width=\linewidth]{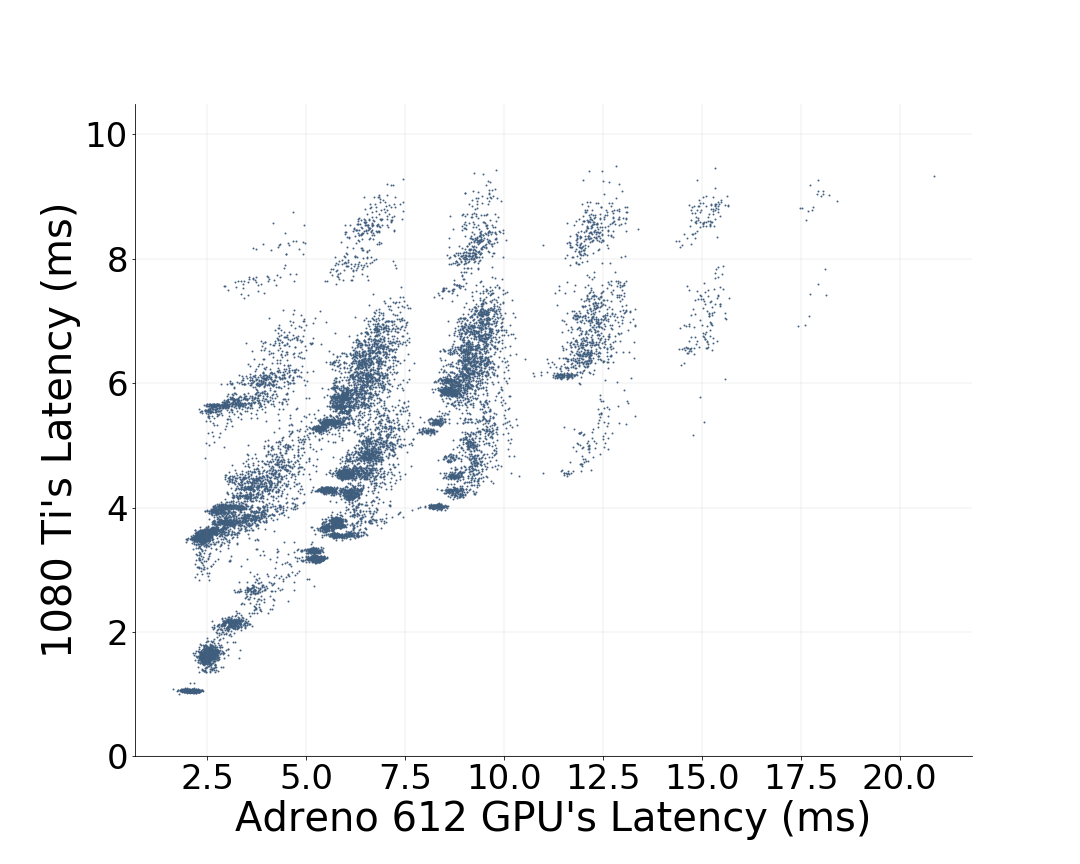}
        \caption{Desktop GPU vs Mobile GPU}
        \end{subfigure}
        \begin{subfigure}[b]{.32\textwidth}
        \includegraphics[width=\linewidth]{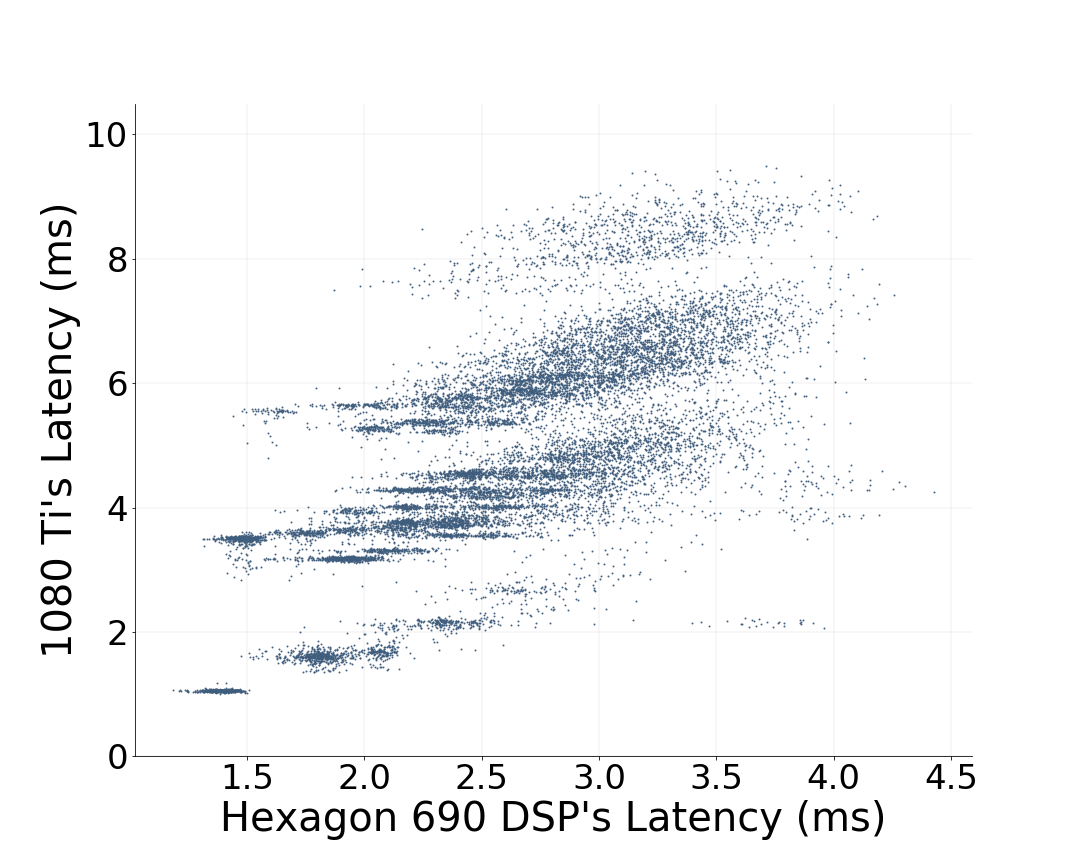}
        \caption{Desktop GPU vs Mobile DSP}
        \end{subfigure}
        \begin{subfigure}[b]{.32\textwidth}
        \includegraphics[width=\linewidth]{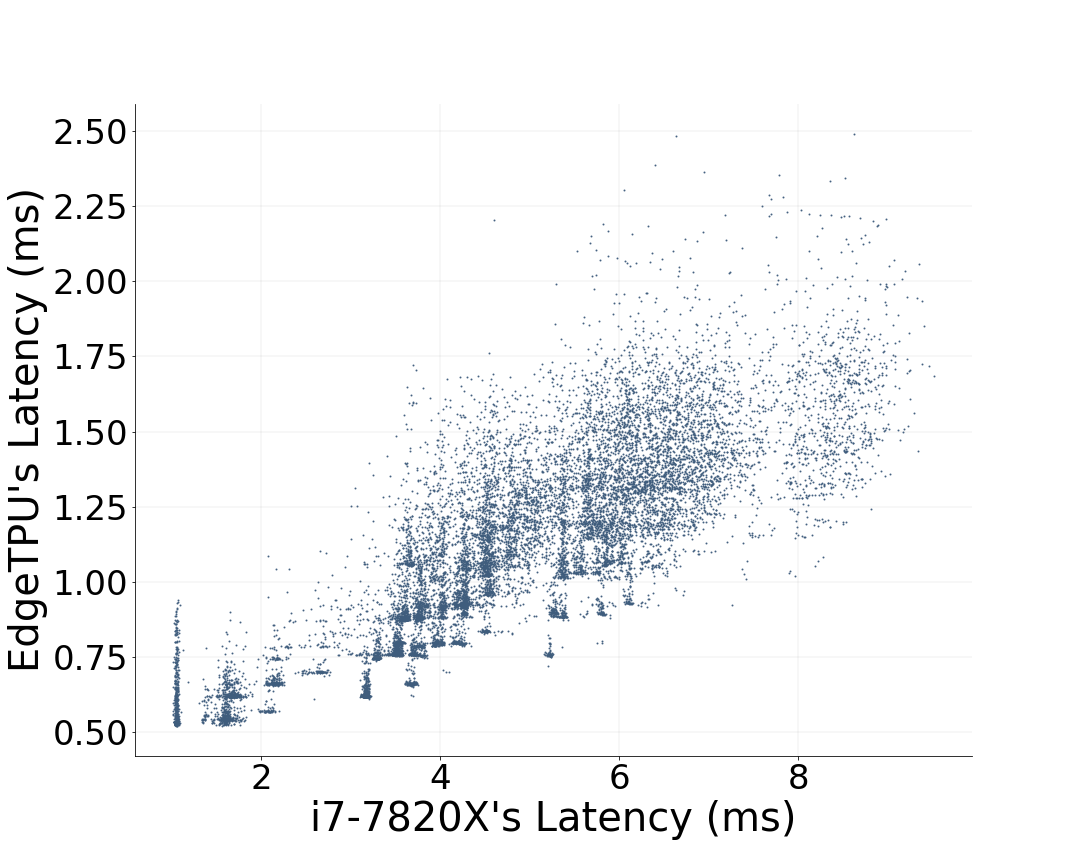}
        \caption{Embed. TPU vs Desktop CPU}
        \end{subfigure}
        \begin{subfigure}[b]{.32\textwidth}
        \includegraphics[width=\linewidth]{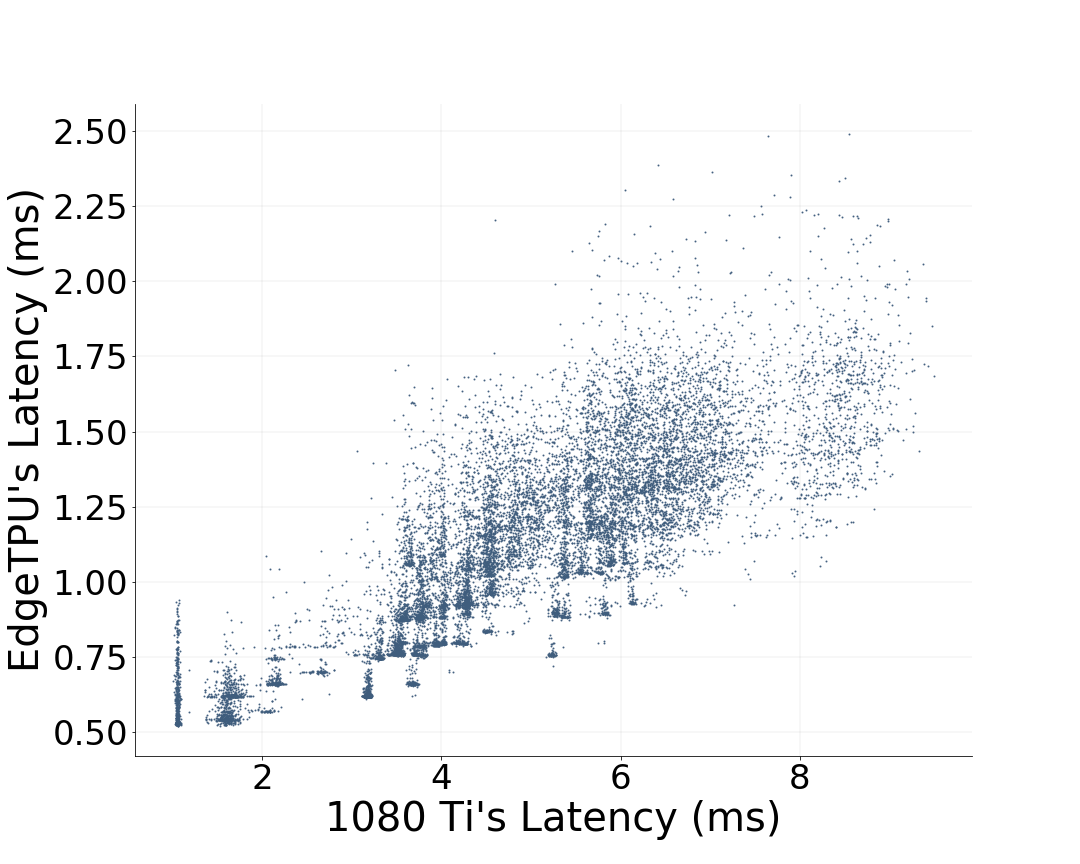}
        \caption{Embed. TPU vs Desktop GPU }
        \end{subfigure}
        \begin{subfigure}[b]{.32\textwidth}
        \includegraphics[width=\linewidth]{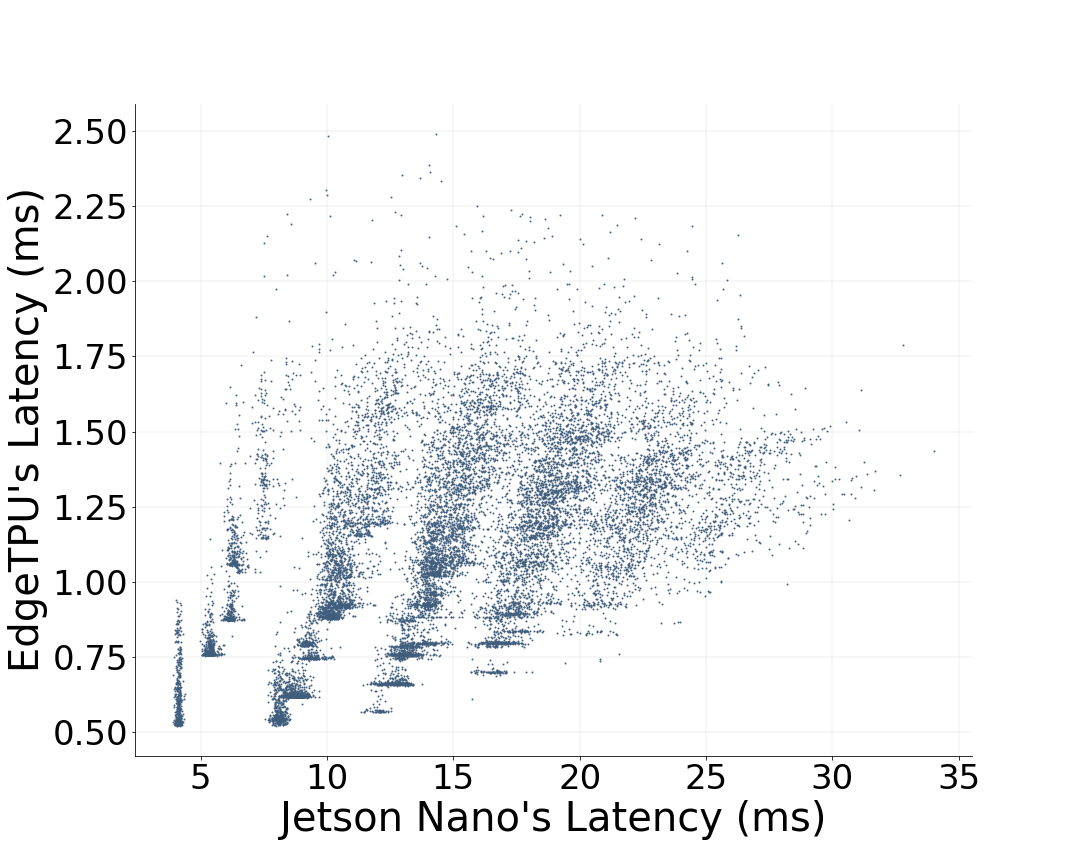}
        \caption{Embed. TPU vs Embed. GPU}
        \end{subfigure}
        &
        \begin{subfigure}[b]{.32\textwidth}
        \includegraphics[width=\linewidth]{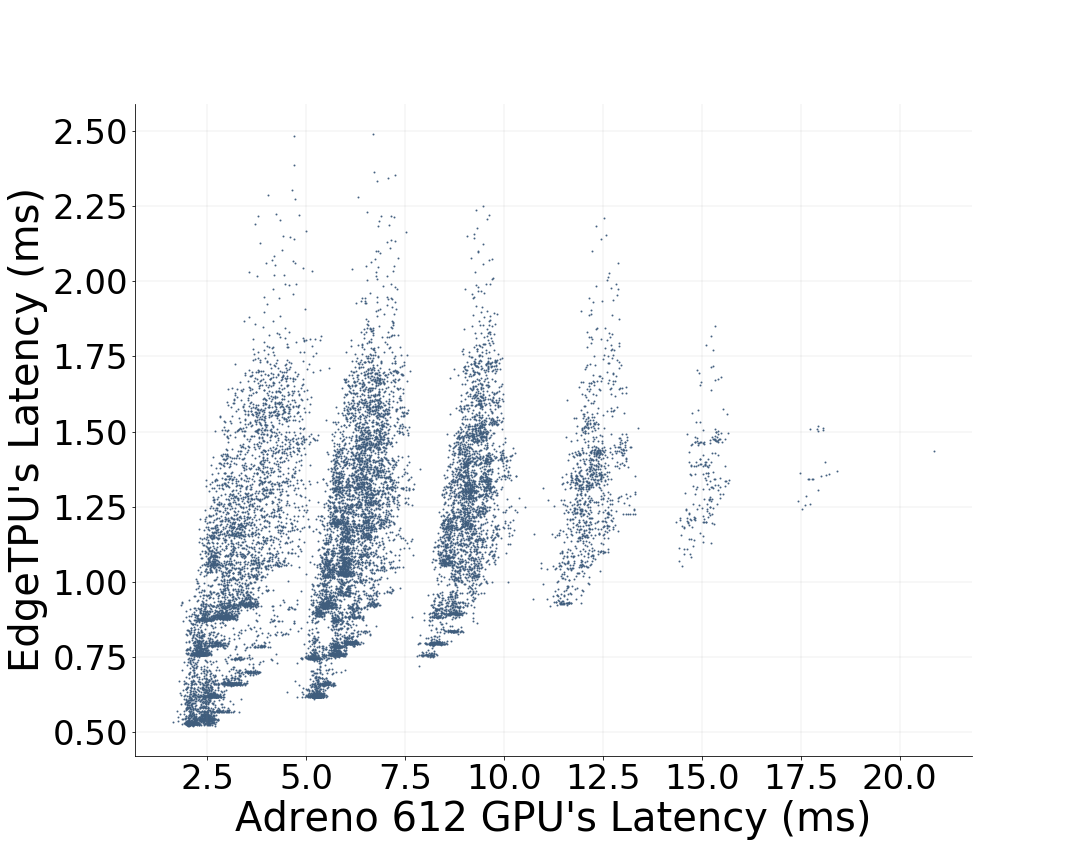}
        \caption{Embed. TPU vs Mobile GPU}
        \end{subfigure}
        \begin{subfigure}[b]{.32\textwidth}
        \includegraphics[width=\linewidth]{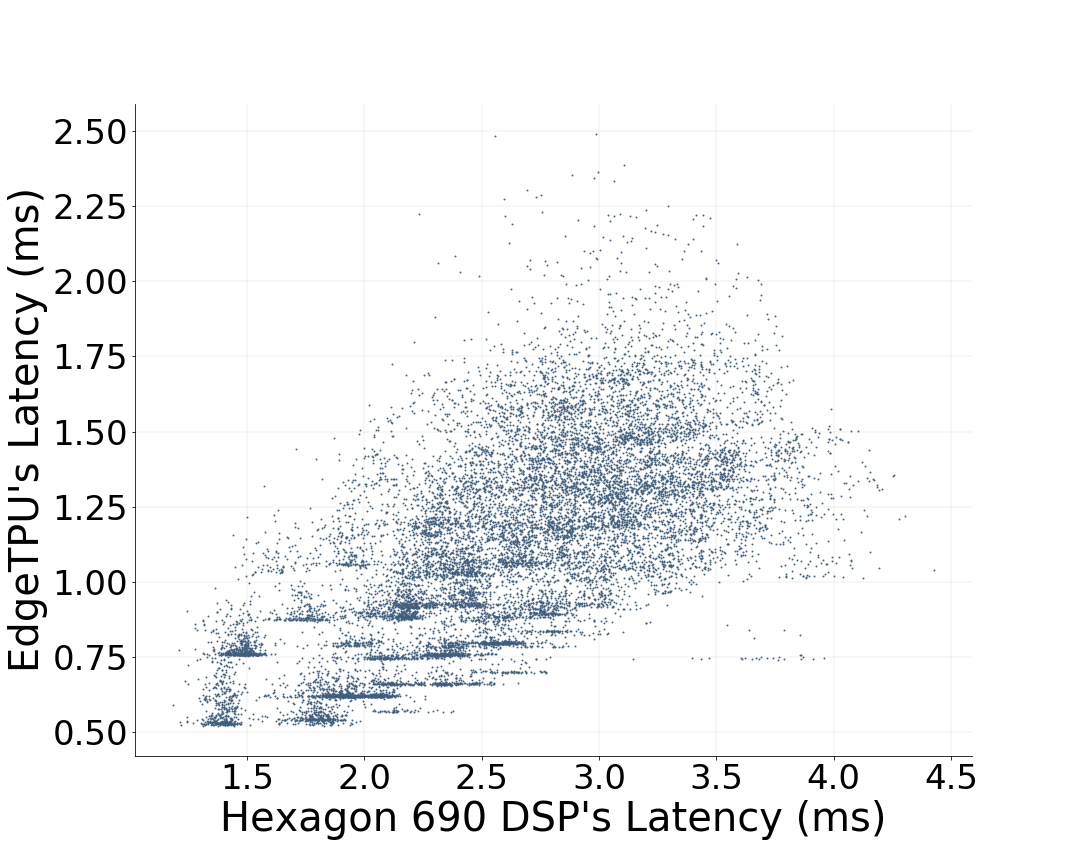}
        \caption{Embed. TPU vs Mobile DSP}
        \end{subfigure}
        \begin{subfigure}[b]{.32\textwidth}
        \includegraphics[width=\linewidth]{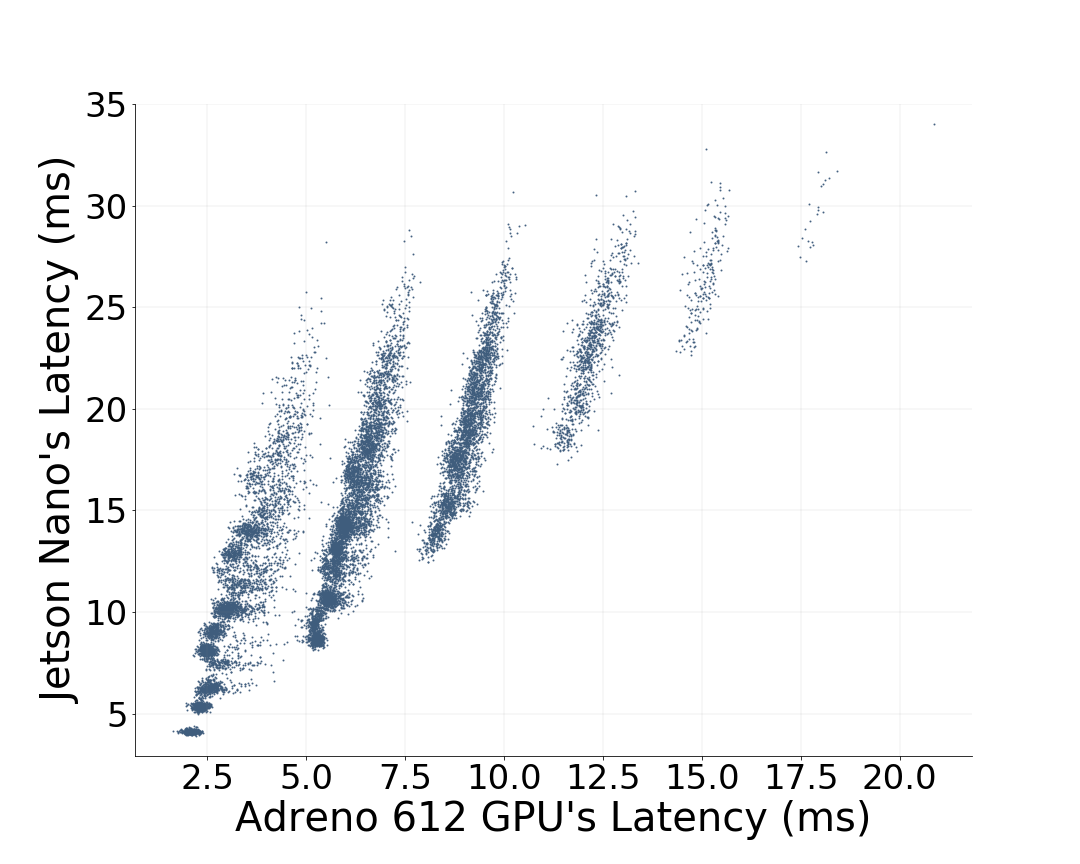}
        \caption{Embed. GPU vs Mobile GPU}
        \end{subfigure}
        \begin{subfigure}[b]{.32\textwidth}
        \includegraphics[width=\linewidth]{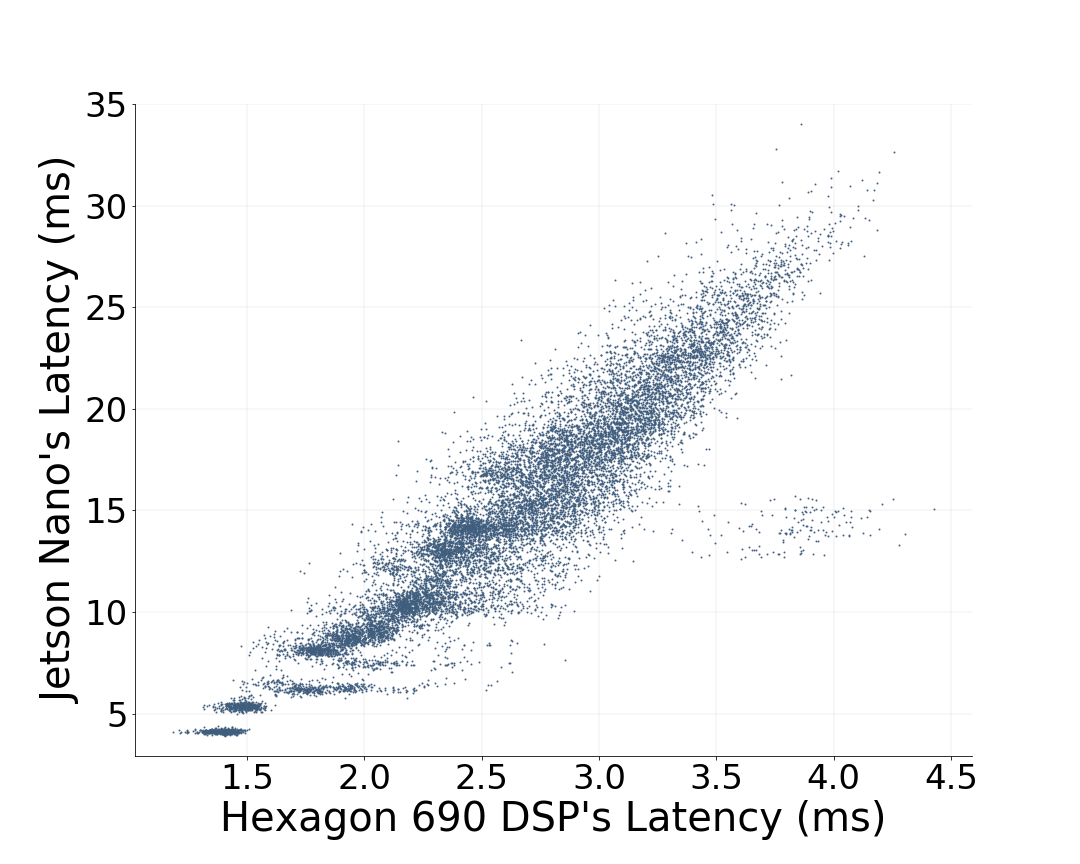}
        \caption{Embed. GPU vs Mobile DSP}
        \end{subfigure}
        \begin{subfigure}[b]{.32\textwidth}
        \includegraphics[width=\linewidth]{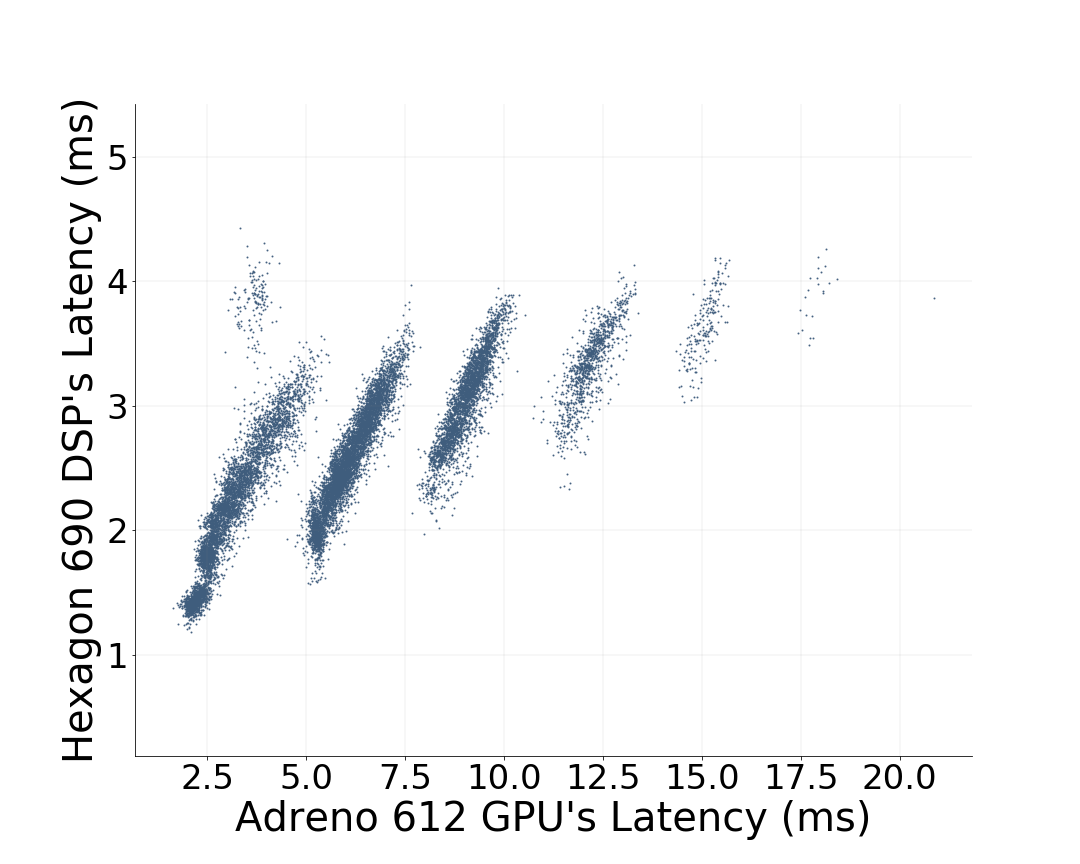}
        \caption{Mobile DSP vs Mobile GPU}
        \end{subfigure}
    \end{tabularx}
    \caption{Latency differs for each class of devices.}
    \label{fig:latency_correlation}
\end{figure}

Figure~\ref{fig:latency_correlation} and Table~\ref{tab:latency_correlation} illustrate the latency correlation between devices. Most of the metrics are not strongly correlated which indicates that having a dedicated latency predictor trained for each class of devices is necessary to provide good latency estimation. This motivates us to provide LatBench as a latency dataset.

\begin{table}[!ht]
\small
\centering
\caption{Latency correlation between various devices.}
\label{tab:latency_correlation}
\begin{tabular}{@{}lccccccc@{}}
 \toprule
        & D. CPU & D. GPU & E. GPU & E. TPU & M. GPU & M. DSP \\
 \midrule
D. CPU  & 1.000  & 0.997  & 0.700  & 0.844  & 0.751  & 0.727  \\
D. GPU  & 0.997  & 1.000  & 0.702  & 0.844  & 0.752  & 0.728  \\
E. GPU  & 0.700  & 0.702  & 1.000  & 0.574  & 0.866  & 0.821  \\
E. TPU  & 0.844  & 0.844  & 0.574  & 1.000  & 0.548  & 0.690  \\
M. GPU  & 0.751  & 0.752  & 0.866  & 0.548  & 1.000  & 0.821  \\
M. DSP  & 0.727  & 0.728  & 0.821  & 0.690  & 0.821  & 1.000  \\
 \bottomrule
\end{tabular}
\end{table}